\documentclass[journal]{IEEEtran}
\ifCLASSINFOpdf
   \usepackage[pdftex]{graphicx}
   \graphicspath{{../Figures/}}
   \DeclareGraphicsExtensions{.pdf,.jpeg,.png}
\else
\fi
\usepackage[table]{xcolor}
\usepackage[utf8x]{inputenc}
\usepackage{amsmath}
\usepackage{array}
\ifCLASSOPTIONcompsoc
 \usepackage[caption=false,font=normalsize,labelfont=sf,textfont=sf]{subfig}
\else
  \usepackage[caption=false,font=footnotesize]{subfig}
\fi
\usepackage{stfloats}
\usepackage{tabularx,booktabs}
\usepackage{multirow}
\usepackage{amssymb}
\usepackage{array}
\usepackage{xcolor}
\usepackage{cite}
\begin{document}
\title{\texttt{SR-GNN}: Spatial Relation-aware Graph Neural Network for Fine-Grained Image Categorization}
\author{Asish~Bera$^*$, ~\IEEEmembership{~Member,~IEEE},
        Zachary~Wharton$^*$, 
        Yonghuai~Liu,~\IEEEmembership{~Senior Member,~IEEE}, \\
        Nik~Bessis,~\IEEEmembership{~ Senior Member,~IEEE},
        and
        ~Ardhendu Behera$^\dagger$,~\IEEEmembership{~Member,~IEEE}
\thanks{A. Bera, A. Behera, Z. Wharton, Y. Liu, and N. Bessis   are with the Department of Computer Science, Edge Hill University, UK.}
\thanks{$*$ Equal contribution, $\dagger$ Corresponding author, beheraa@edgehill.ac.uk}%
\thanks{Manuscript received Month 08, 2021; revised Month 02 and 04, 2022.}}
\markboth{Journal of \LaTeX\ Class Files,~Vol.~00, No.~0, September~2022}%
{Shell \MakeLowercase{\textit{et al.}}: Bare Demo of IEEEtran.cls for IEEE Journals}

\maketitle
\begin{abstract}
Over the past few years, a significant progress has been made in deep convolutional neural networks (CNNs)-based image recognition. This is mainly due to the strong ability of such networks in mining discriminative object pose and parts information from texture and shape. This is often inappropriate for fine-grained visual classification (FGVC) since it exhibits high intra-class and low inter-class variances due to occlusions, deformation, illuminations, etc. Thus, an expressive feature representation describing global structural information is a key to characterize an object/ scene. To this end, we propose a method that effectively captures subtle changes by aggregating context-aware features from most relevant image-regions and their importance in discriminating fine-grained categories avoiding the bounding-box and/or distinguishable part annotations. Our approach is inspired by the recent advancement in self-attention and graph neural networks (GNNs) approaches to include a simple yet effective relation-aware feature transformation and its refinement using a context-aware attention mechanism to boost the discriminability of the transformed feature in an end-to-end learning process. Our model is evaluated on eight benchmark datasets consisting of fine-grained objects and human-object interactions. It outperforms the state-of-the-art approaches by a significant margin in recognition accuracy.
\end{abstract}

\begin{IEEEkeywords}
Attention mechanism, Convolutional Neural Networks, Graph Neural Networks, Human action, Fine-grained visual recognition, Relation-aware feature transformation.
\end{IEEEkeywords}

\IEEEpeerreviewmaketitle
\vspace{-0.2 cm}
\section{Introduction}
\IEEEPARstart{T}{he} advent of deep convolutional neural networks (CNN) has significantly enhanced image recognition performance in the past decade. It is achieved mainly due to their abilities to provide a high-level description (\textit{e.g.}, global shape and appearance) of image content by capturing discriminative object-pose and -parts information from texture and shape. This high-level description is more apposite for the large-scale visual classification (LSVC) tasks consisting of distinctive categories (\textit{e.g.}, ImageNet and COCO datasets). However, their performance in solving fine-grained visual classification (FGVC) problems is not at the same level as in LSVC. This is mainly due to the subtle changes between hard-to-distinguish object classes in FGVC, but most often visually measurable by humans. Common datasets in FGVC include  different types of birds \cite{wah2011caltech}, flowers \cite{nilsback2008automated}, dogs \cite{khosla2011novel}, aircraft \cite{maji2013fine}, car models \cite{krause20133d}, etc. A typical observation in FGVC is that objects from different classes share visually similar structures (large inter-class similarities), and objects in the same class often exhibit significant variations due to different structures, lighting, clutter and viewpoints (large intra-class variations). As a result, it is a challenging task to learn a unified and discriminative representation for each class.
\begin{figure}[t]
    \centering
    \includegraphics[width=0.39\textwidth, keepaspectratio ]{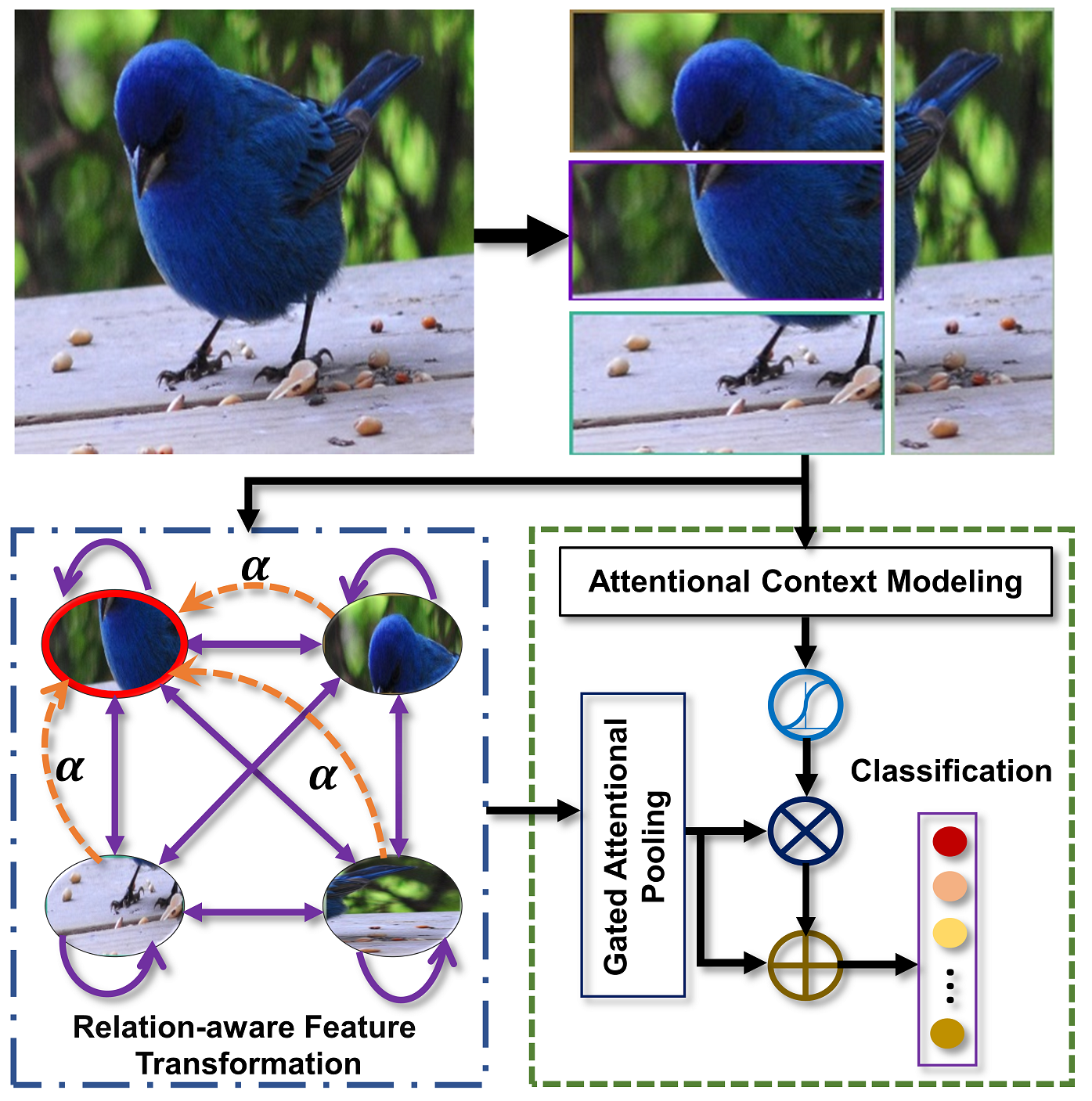}
    \caption{Our \texttt{SR-GNN} consists of a GNN-based relation-aware feature transformation by propagating information between image regions and an attentional context modeling to refine these transformed features. They jointly tackle the challenge of describing and discriminating subtle variations in FGVC by exploring the visual-spatial relationships among regions and aggregating the context-aware features. For clarity, 4 different regions are shown here. }
    \label{fig:high_level}
    \vspace{-.5cm}
\end{figure}
A key step to address this challenge is to extract discriminating features from vital object-parts and combine them for the representation of a consistent distinctive global structure of a given class. The current state-of-the-art (SotA) approaches are mainly craftily designed to extract such discriminative features and structures by exploring 1) part annotations from humans, and 2) automatically finding these discriminative parts from the whole image. We refer the interested readers to \cite{wei2021fine} for a detailed survey. Most of the earlier works belong to the first category in which the locations of discriminative object-parts are given (\textit{e.g.}, bounding box or mask). Some methods learn part-based detectors, and some leverage semantic segmentation to localize distinct parts. The parts annotation is a cumbersome and expensive human labeling task that is often prone to human errors and requires expert knowledge. Moreover, part-based methods limit both scalability and practicality of real-world FGVC applications. Thus, many recent methods have used image-level labels to guide their models in identifying the key object parts to discriminate the sub-categories  
by exploring attention mechanisms in the image space or feature space \cite{behera2021context, behera2020regional, liu2019bidirectional,  bera2021attend} to automatically mine discriminative features. 

\noindent\textbf{Motivation: }
In this work, we propose a simple yet effective connection between the image space and feature space to discriminate subtle variances in FGVC. Our approach is motivated by the recent success of Graph Neural Networks (GNN) \cite{kipf2016semi} and attention mechanisms \cite{vaswani2017attention,BahdanauCB14} in deep learning. Many SotA methods for FGVC use a  pre-trained object/part detector (or proposal from mask R-CNN) in a weakly supervised manner, resulting in the absence of detailed description, which is indispensable to capture better object-part and part-part relationships to model the subtle changes. These parts can be affected by occlusions, noisy backgrounds, pose variations and ambiguous repetitive patterns. Thus, multiple partial descriptors for a  part are potentially useful in disambiguating and discriminating subtle changes since the model learns meaningful complementary information to provide a rich representation of an object. Our method neither considers object-part proposal/bounding-box nor tries to localize them. Instead, it automatically learns a richer representation of an object by exploring the attention-driven visual-spatial relationships among a pool of geometrically-constrained regions. These regions are generated using the region proposal in the Regional Attention Network (RAN) \cite{behera2020regional}, which uses cells and blocks in computing histograms of oriented gradients (HOG). 

To describe a richer representation with the discrimination power for subtle variations, we design a spatial relation-aware GNN 
(\texttt{SR-GNN}) to model visual-spatial relations between regions. These relationships are captured using a novel relation-aware feature transformation and its refinement via attentional context modeling, as conceptually shown in Fig. \ref{fig:high_level}. Firstly, a backbone CNN is used to extract high-level visual features. These are upsampled for feature pooling using geometrically-constrained regions of various sizes and positions  (Fig. \ref{fig:full_model}(a)). Secondly, it transforms these pooled features using relation-aware feature transformation leveraging GNN that captures the visual-spatial relationships via propagating information between regions represented as the nodes of a connected graph (Fig. \ref{fig:full_model}(b)) to enhance the discriminative power of features. To address the limitations of over-smoothing and a large number of learnable parameters in GNN \cite{kipf2016semi}, it adapts the topic sensitive PageRank \cite{klicpera2018predict} using the approximate personalized propagation of neural predictions (APPNP) message-passing algorithm via power iteration, achieving linear computational complexity. Then it applies a novel gated attentional pooling to the learned graph nodes for final feature representation. Finally, it employs an attentional context modeling (Fig. \ref{fig:full_model}(c)) that explores self-attention \cite{vaswani2017attention} and weighted attention \cite{BahdanauCB14} in an innovative way to learn a weight vector, which is multiplied with the final relation-aware transformed feature extracted in the previous step as a refined feature before classification. 

Performance-wise, CAP \cite{behera2021context} is currently top of the  leaderboard for many FGVC datasets. It uses attention to accumulate features from integral regions in a context-aware fashion. Then it uses an LSTM to learn the spatial arrangements (context encoding) of these integral regions for subtle discrimination to tackle FGVC tasks. Finally, it aggregates information by grouping similar responses of the LSTM's hidden states to generate locally aggregated descriptors using NetVLAD in the classification step. Our proposed \texttt{SR-GNN} is significantly different from CAP in the following aspects: (a) introduction of a relation-aware spatial graph with the APPNP message-passing algorithm to extract more expressive features by capturing visual-spatial relationships via propagating information between regions, (b) a graph-based gated attentional pooling to aggregate features from graph nodes, and (c) an attentional context modeling that consists of self-attention and weighted attention to compute a weight vector for the refinement of the final relation-aware features for classification. The only similarity in both approaches is the feature extraction using a CNN backbone. Although the context-aware attention in CAP and self-attention in \texttt{SR-GNN} (Section \ref{sec:cotext}) are inspired by the same self-attention mechanism in natural language processing \cite{vaswani2017attention}, they are explored differently to solve the specific problem in hand. In CAP, it accumulates features from various regions and an LSTM is then applied for context encoding by considering sequential information. Whereas in \texttt{SR-GNN}, it is investigated in a novel way to compute a weight vector (Fig. \ref{fig:full_model}(c)) by exploring contextual information via adapting self-attention \cite{vaswani2017attention} and weighted attention \cite{BahdanauCB14}. Unlike in CAP, the weighted attention does not consider sequential information, but learns the weight vector from multiple regions by joint learning. To the best of our knowledge, we are the first to investigate the efficiency of the PageRank algorithm leveraging APPNP to advance the FGVC accuracy. These key concepts are also novel in comparison to other SotA methods, including more recent vision Transformers (ViT) \cite{dosovitskiy2020image,liu2021swin, he2021transfg, miao2021complemental}. 

The main contributions of this paper are: 
1) A novel relation-aware visual representation and its refinement via attentional spatial context for enriching region-level description to capture the subtle changes and eventually enhance the FGVC performance; 2) An easy-to-use end-to-end FGVC deep network that does not require object/parts bounding boxes annotation or proposal and thus has an advantage of easy implementation; 
3) A proposal of a gated attentional pooling for the automatic aggregation of the relation-aware features; and
4) Ablation studies and visual analysis of the performance of
\texttt{SR-GNN}. 

The rest of this paper is organized as follows: Section \ref{rel_work} summarizes related works on FGVC. Section \ref{proposed} describes the proposed framework. The experimental results are discussed in Section \ref{experiments}, and an in-depth ablation study is presented in Section \ref{Ablation}, followed by a conclusion in Section \ref{conclusion}.

\section{Related Work} \label{rel_work}
Our work is closely related to weakly-supervised object-parts, attentional and GNN methods for FGVC, including human actions. We present a concise survey of these approaches.
\vspace{ - 0.5 cm}
\subsection {Object-Parts Based Methods} 
Informative object-parts are crucial and are explored \cite{ zhang2016weakly, yao2017autobd, ge2019weakly} for robust subtle discrimination. Distinct object-parts are selected at multiple scales from object proposals in \cite{zhang2016weakly} to distinguish subtle variations. In \cite{yao2017autobd}, the objectness map is generated using deep features for part-level and  object-level descriptions and their fusion for visual discrimination. Object detection and instance segmentation pipeline are iterated in \cite{ge2019weakly} for complementary part localization, and then LSTM is used to encode contextual details. Similarly, local details are learned from distinct patches which are generated by shuffling the whole image into smaller patches \cite{chen2019destruction, zhong2020random,wang2019weakly}. 
The global image structure is randomly disrupted by a region confusion technique in \cite{chen2019destruction} to learn finer  details and semantic correlation within sub-regions. Whereas random erasing in \cite{zhong2020random} introduces additional noise by object-part occlusion to select informative patches. A region grouping sub-network in \cite{wang2019weakly} learns correlation weight coefficients between regions to select and refine discriminatory patches. Similarly, vision and language modalities are combined in \cite{he2017fine2}. The vision localizes objects using saliency and co-segmentation, while the language applies cross-modal analysis to correlate natural language descriptions and discriminative object parts. A multi-scale and multi-granularity deep reinforcement learning in \cite{he2019and} finds hierarchical discriminative regions in multiple granularities and automatically determines the number of such regions to boost the accuracy. Likewise, a hierarchical representation of image-regions enhances  action classification accuracy in \cite{li2018reassessing}. Most of these approaches focus on locating the informative object-parts and then extract expressive feature descriptors. In sharp contrast, our method learns distinct features by mining visual-spatial correlations using contextual cues from a set of regions, relying on cells and blocks used in HOG \cite{behera2020regional}. 
\begin{figure*}[t]
    \centering
    \includegraphics[width=0.9\textwidth]{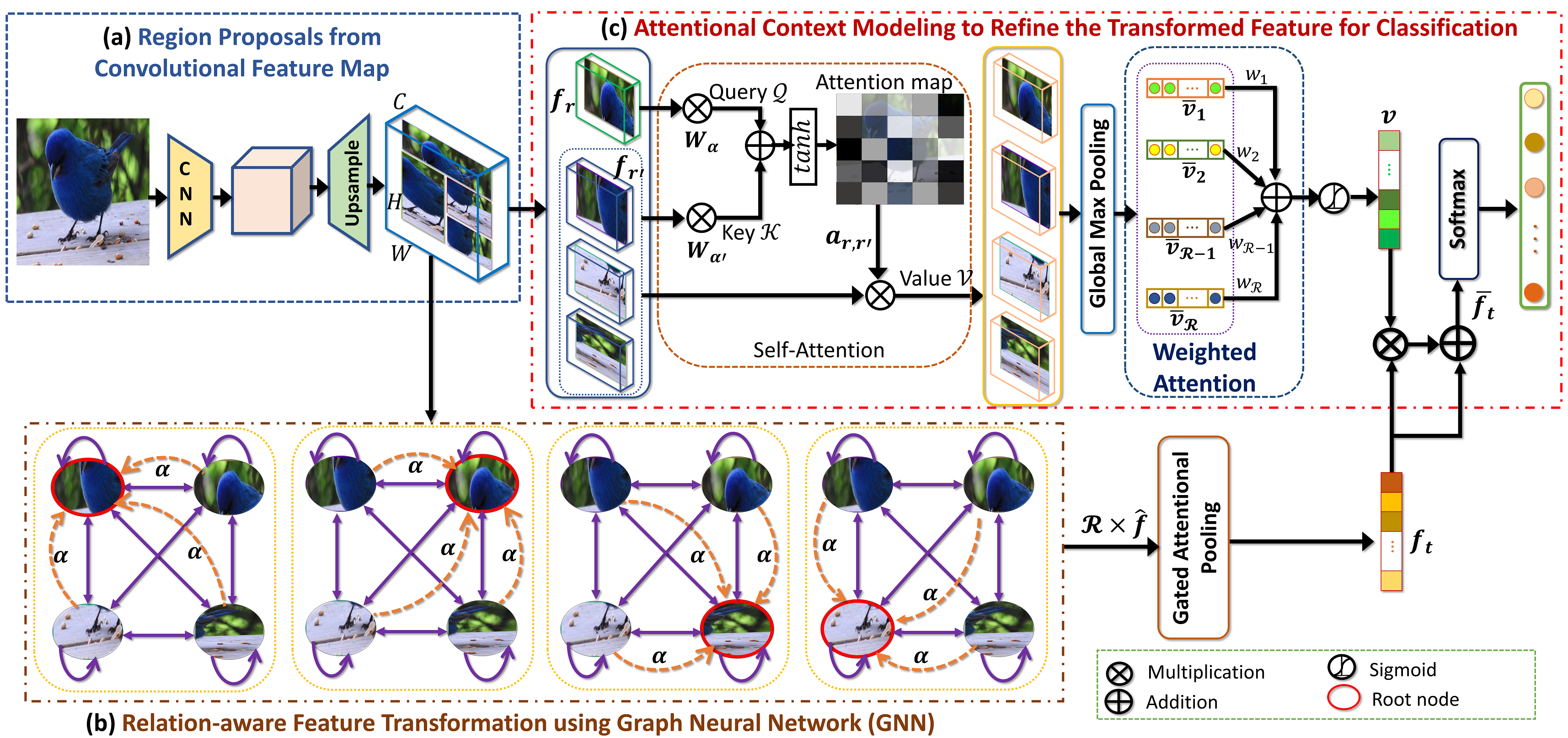}
    \caption{The architecture of our \texttt{SR-GNN}. (a) Extract features using a set of regions from the upsampled CNN features of a given input image.  (b) \textbf{Relation-aware Feature Transformation:} it updates each region's visual-spatial relationships by propagating information between them using message propagation in GNN. Then, the transformed features from all regions are used by the gated attentional pooling to produce the final transformed features $f_t$. (c) \textbf{Attentional Context Refinement:} firstly, it computes an attention-focused context vector $\mathbf{v}_r$ from region-pooled features using self-attention. Next, a context refinement weight-vector $\mathbf{v}$ is computed over the weighted summation of all $\mathbf{v}_r$. Finally, the context vector $\mathbf{v}$ is used to refine the transformed feature $f_t$ and then feeds it to the \texttt{Softmax} layer for classification.}
    \label{fig:full_model}
    \vspace{-0.4cm}
\end{figure*}
\subsection {Attention-based Approaches} 
The attention mechanism is proliferated to identify salient regions and/or subtle discriminatory features to attain superior performance \cite{behera2021context, behera2020regional, liu2019bidirectional,  lopez2020pay, zheng2019looking}. A trilinear attention sampling in \cite{zheng2019looking} learns features from hundreds of part proposals and then applies knowledge distillation to integrate them. Top-down and bottom-up attentions are combined in an attentional pyramid CNN \cite{ding2020weakly} to aggregate high-level semantic and low-level finer features. In \cite{liu2019bidirectional}, a feedback path is connected from a recognition agent to an attention agent to optimize region proposals. 
Regional attention network (RAN) \cite{behera2020regional} presents a hybrid attention method that focuses on semantic informativeness from multiple regional contexts for fine-grained gesture/action recognition. Attend and guide network (AG-Net) \cite{bera2021attend} applies to scale-invariant feature transform (SIFT) keypoints and Gaussian mixture model to propose regions that are guided by the attention mechanism for fine-grained visual categorization of objects and human actions. Modular attention in \cite{lopez2020pay} applies multiple attention modules to focus on region-based predictions refined by attention gates. Attention on feature channels is explored in \cite{chang2020devil} to focus on discriminatory regions. A sparse attentional framework in \cite{ding2019selective} follows a selective sampling technique to estimate finer details. A counterfactual attention learning is proposed in \cite{rao2021counterfactual} to measure the visual attention quality that guides the learning process via counterfactual intervention to learn more useful attention for enhanced FGVC accuracy. Similarly, object extent learning and spatial context learning are integrated in look-into-object \cite{zhou2020look} to understand the object structure by automatically modeling the context information among regions. In \cite{zhuang2020learning}, attentive pairwise interaction network discovers contrastive cues from a pair of images, and discriminates them with pairwise attentional interaction in an end-to-end manner. More recently, a sequence of image patches with positional embedding and multi-head self-attention are integrated in vision Transformers \cite{dosovitskiy2020image, liu2021swin, he2021transfg, miao2021complemental} to enhance FGVC. Swin Transformer \cite{liu2021swin} exploits a hierarchical shifting window-based self-attention with linear computational complexity. A part selection module is adapted to improve over pure ViT in \cite{he2021transfg} by integrating raw attention weights of the Transformer into an attention map to guide the ViT. A complemental attention module and multi-feature fusion module are combined in \cite{miao2021complemental} using Swin Transformer. Inspired by these, we propose a simple yet effective attention mechanism to refine the GNN-driven features at multi-scale and their aggregation for further performance improvement. 
\vspace{ -0.2cm}
\subsection {Graph Neural Networks (GNN)} 
Following the CNN concept, GNN is proposed to explore problems consisting of non-Euclidean data. It is powerful for smoother messages passing between neighboring nodes to enhance performance \cite{kipf2016semi}. Recently, it has been explored in zero-shot recognition, multi-label image recognition, image captioning, visual question answering, and the others \cite{wu2020comprehensive}. However, its efficacy in FGVC is yet to be fully explored. In \cite{wang2020category}, GNN is used to learn latent attributes by modeling semantic correspondence between discriminative regions within the same sub-category. In  \cite{wang2020graph}, region correlation is explored to discover informative regions using the criss-cross graph propagation sub-network and correlation feature via a unified framework. However, both methods are limited to a few regions per image (\textit{e.g.}, 4) which may be sub-optimal for building and propagating information within sub-networks for effective context modeling to address FGVC. A graph-based relation discovery (GaRD) method \cite {zhao2021graph} learns the positional and semantic feature relationships and adopts a feature grouping strategy to tackle FGVC. We propose a GNN-based spatial relation-aware feature aggregation by considering multiple partial descriptors to propagate and capture finer complementary information between neighboring regions by approximating topic-sensitive PageRank \cite{klicpera2018predict}.  
\section{Proposed Approach} \label{proposed}
The proposed \texttt{SR-GNN} architecture is shown in Fig. \ref{fig:full_model}. It takes as input an input image, extracts a high-level convolutional feature map, applies region-based visual-spatial feature selection and refines the transformed features using an attentional spatial context modeling to advance FGVC. 
\vspace{-0.2 cm}
\subsection{Problem Formulation}
To train an image classifier, a set of $N$ images $I=\{I_n|n=1, 2, \dots, N\}$ and their respective class labels are given. The classifier learns a mapping function $\mathcal{F}$ that predicts $\hat{y}_n=\mathcal{F}\left(I_n\right)$, which matches the true label $y_n$. During training, it learns $\mathcal{F}$ by minimizing a loss $\mathcal{L}\left(y_n,\hat{y}_n\right)$ between the true and the predicted labels. In this work, $\mathcal{F}$ is an end-to-end deep network in which we introduce a simple yet effective network modification to advance the SotA in FGVC. The mechanism focuses on two main components to capture the fine-grained changes in images: 1) relation-aware feature selection and transformation, and 2) an attentional context modeling to refine these transformed features. Therefore, the mapping function $\mathcal{F}$ consists of:\vspace{-.15cm}
\begin{equation}
    \mathcal{F} = \text{Softmax}\left(\underbrace{\mathcal{F}_1(I_n;\theta_t)}_\textrm{Feature Transform}\bigotimes \overbrace{\sigma(\mathcal{F}_2(I_n;\theta_c))}^\textrm{Attentional Context}\right),
    \label{eq_1}
\end{equation} 
where $\theta_t$ and $\theta_c$ are the learnable parameter sets for the feature transformation from the given image $I_n$ to a high-level descriptor and the attentional context refinement, respectively. $\sigma(.)$ is an element-wise sigmoid function to regulate how much of the transformed feature should be considered in decision making. 
\vspace{-0.5 cm}
\subsection{CNN Feature Map and Region Proposals}\label{sec:roi}
We use the lightweight Xception \cite{chollet2017xception} backbone for extracting CNN features like CAP \cite{behera2021context} 
and upsample it for region proposals as in \cite{behera2020regional} by exploring cells and blocks in the HOG computation. The region proposal generates $\mathcal{R}$ possible regions of different aspect ratios and areas. For clarity, 4 regions are shown in Fig. \ref{fig:regions}. Each region is then represented with a feature vector $f$ of dimension $w (\text{width})\times h(\text{height})\times C(\text{channels})$ via bilinear interpolation to implement differentiable image transformations (conceptualized in Fig. \ref{fig:full_model}(a)).

\begin{figure}[t]
\vspace{-0.5 cm}
\renewcommand*\thesubfigure{\arabic{subfigure}}
    \centering
    \subfloat[]{\includegraphics[width=0.12\textwidth] {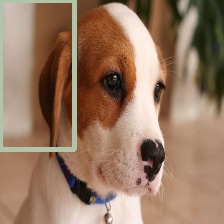}}\hfill
        \subfloat[]{\includegraphics[width=0.12\textwidth] {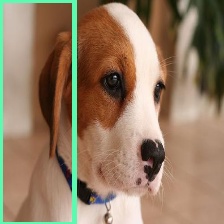}}\hfill
            \subfloat[]{\includegraphics[width=0.12\textwidth] {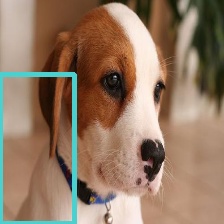}}\hfill
                \subfloat[]{\includegraphics[width=0.12\textwidth] {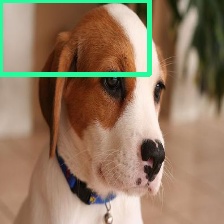}}
    \caption{Various regions (4 shown for clarity) from the region proposal. These regions are used for bilinear pooling from the upsampled CNN features. }
    \label{fig:regions}
\vspace{-0.3 cm}    
\end{figure}
\vspace{-0.2 cm}
\subsection{Relation-Aware Feature Transformation}\label{sec:feat}
We represent an image $I_n$ using  $\mathcal{R}$ regions. The essential aim is to update each region's visual representation $f$ by propagating information between regions to characterize their visual-spatial relationships, which capture the subtle variations between them. Thus, the first step to representing these relationships is to build a graph $G=(\mathcal{R},E)$ with nodes $\mathcal{R}$ and the connections between them via edges $E$. As a result, GNN can then be used to learn and reason visual-spatial relationships by propagating messages from one region to its connected neighbors in the graph. The nodes are described by a set $\mathbf{X}=\{f\}$ consisting of a number $\mathcal{R}$  of input features $f$, and the respective output $\mathbf{Y} = \{\hat{f}\}$ with transformed feature $\hat{f}$ per node. The graph $G$ is described by the adjacency matrix $\mathbf{A} \in \mathbb{R}^{\mathcal{R}\times \mathcal{R}}$. The adjacency matrix $\tilde{\mathbf{A}}=\mathbf{A}+\mathbf{I}_{\mathcal{R}}$ denotes $\mathbf{A}$ with added self-loops and $\mathbf{I}_{\mathcal{R}}$ is the identity matrix (Fig. \ref{fig:full_model}(b)). A well-known message passing algorithm is the GNN \cite{kipf2016semi} in which a simple layer-wise propagation rule is used: $\mathbf{H}^{(l+1)} = \sigma\left(\hat{\tilde{\mathbf{A}}}\mathbf{H}^{(l)}\mathbf{W}^{(l)}\right)$, with $\mathbf{H}^{(0)}=\mathbf{X}$ and $\mathbf{H}^{(L)}=\mathbf{Y}$, $l=0, 1, \dots, L-1$ being the number of layers, $\mathbf{W}^{(l)}$ is a weight matrix for the $l$-th layer and $\sigma(.)$ is a non-linear activation function (\textit{e.g.}, ReLU). $\hat{\tilde{\mathbf{A}}}=\tilde{\mathbf{D}}^{−1/2}\tilde{\mathbf{A}}\tilde{\mathbf{D}}^{−1/2}$ is the symmetrically normalized adjacency matrix, and $\tilde{\mathbf{D}}$ is the diagonal node degree matrix of $\tilde{\mathbf{A}}$. The GNN message passing algorithm is limited to a smaller neighborhood mainly due to 1) aggregation by averaging causes over-smoothing if too many layers are used and thus, loses its focus on the local neighborhood; and 2) a larger neighborhood significantly increases the depth and number of learnable parameters since the common aggregation schemes use learnable weight matrices in each layer. To address these, we adapt the approximate personalized propagation of neural predictions (APPNP) message-passing algorithm \cite{klicpera2018predict}. It achieves the linear computational complexity by approximating topic-sensitive PageRank via power iteration, relating to a random walk with restarts. Each power iteration step is calculated as: \vspace{-.1cm}
\begin{equation}
\begin{split}
\begin{aligned}
    \mathbf{H} &= \text{MLP}\left(\text{GAP}(\mathbf{X});\theta\right), \\
    \mathbf{Y}^{(0)} &= \mathbf{H}, \\
    \mathbf{Y}^{(k+1)} &= (1-\alpha)\hat{\tilde{\mathbf{A}}}\mathbf{Y}^{(k)}+\alpha \mathbf{H},\\
    \mathbf{Y}^{(K)}&= \text{Sigmoid} \left( (1-\alpha)\hat{\tilde{\mathbf{A}}}\mathbf{Y}^{(K-1)}+\alpha \mathbf{H}\right),
    \end{aligned}
\end{split}
\label{eq_2}
\end{equation} 
where \textit{global average pooling} (GAP) at each node reduces the dimension of $f$: $w\times h\times C \rightarrow 1\times 1\times C$; $\alpha \in(0,1]$ is the teleport (or restart) probability influencing the size of the neighborhood for each node; and $K$ is the number of power iteration steps ($k \in [0,K-2]$). MLP is a multi-layer perceptron with a parameter $\theta$ for predicting $\mathbf{H}$ that allows preserving the node’s local neighborhood, and acts as both the starting vector and the teleport set. For example, every column of $\mathbf{H}$ defines a distribution over regions that acts as a teleport set. Note that MLP operates on each node’s feature $f$ independently, allowing for parallelization. 

Each node transforms a region into a feature vector $\hat{f}$. Now the aim is to aggregate all nodes' features (\textit{i.e.}, $\mathcal{R}\times \hat{f}$) into a single image-level descriptor $f_t$. We achieve this by adapting the gated attentional pooling \cite{li2016gated} that is computed as:\vspace{-.12cm}
\begin{equation}
    f_t = \sum_{i=1}^{\mathcal{R}} \sigma(\hat{f}_i \mathbf{W}_1+\mathbf{b}_1)\odot (\hat{f}_i \mathbf{W}_2+\mathbf{b}_2),
    \label{eq_3}
\end{equation}
where weight matrices $\mathbf{W}_1$ and $\mathbf{W}_2$, and biases $\mathbf{b}_1$ and $\mathbf{b}_2$ are learnable parameters. $\hat{f}_i\in\mathbf{Y}$ represents the $i^{th}$ node output feature of the graph $G$ in (\ref{eq_2}). $\sigma(.)$ is an element-wise sigmoid and acts as a soft attention mechanism that decides which regions are more relevant to the current graph-level task, and $\odot$ is the Hadamard product. The learnable feature transformation parameter in (\ref{eq_1}) is thus $\theta_t=\{\theta, \mathbf{W}_1, \mathbf{W}_2, \mathbf{b}_1,\mathbf{b}_2\}$. 
\subsection{Transformed Feature Refinement}\label{sec:cotext}
It is inspired by the self-attention mechanism in natural language processing \cite{vaswani2017attention}. The self-attention handles a long path-length contextual modeling by a lightweight gating mechanism in which the attention matrix is generated using a simple dot-product. In self-attention, \textit{query} $\mathcal{Q}$, \textit{key} $\mathcal{K}$ and \textit{value} $\mathcal{V}$ are learned from the same input, and are different from the traditional attention-based sequence-to-sequence models. Often, $\mathcal{Q}$, $\mathcal{K}$ and $\mathcal{V}$ are learned by three independent transformation layers. The dot product of $\mathcal{Q}$ and $\mathcal{K}$ results in the attention weight matrix, which is multiplied with $\mathcal{V}$ to produce the desired transformed feature representation. We adapt this principle to compute attention within a given region $r$ (self-loop) as well as between other regions $r$ and $r'$ ($r,r' \in \mathcal{R}$ and $r'\ne r$). The aim is to generate an attention-focused context vector (\textit{i.e.}, \textit{value} $\mathcal{V}$) that enables our model to selectively focus on more relevant regions to generate holistic context information. Thus, in our self-attention although $\mathcal{Q}$, $\mathcal{K}$ and $\mathcal{V}$ vectors are learned from the same input image but focus on different regions \textit{i.e.}, $\mathcal{Q}$ is learned from $r$ whereas, $\mathcal{K}$ and $\mathcal{V}$ are learned from $r'$. Let $f_r$ and $f_{r'}$ be the high-level convolutional features representing the regions $r$ and $r'$, respectively. The attention-focused context vector $\mathbf{v}_r \in \mathcal{V}$ for region $r$ is computed as (Fig. \ref{fig:full_model}(c)):\vspace{-.07cm}
\begin{equation}
\begin{split}
    \mathbf{v}_r = \sum_{r'=1}^{\mathcal{R}}&a_{r,r'}f_{r'},\text{ } a_{r,r'}=\text{Softmax}(\mathbf{W}_{a}\alpha_{r,r'}+\mathbf{b}_a)\\
    \alpha_{r,r'} &= \text{tanh}(\overbrace{\mathbf{W}_\alpha f_r}^\textrm{\textit{query} $\mathcal{Q}$} + \overbrace{\mathbf{W}_{\alpha'}f_{r'}}^\textrm{\textit{key}  $\mathcal{K}$}+\mathbf{b}_{\alpha}),
\end{split}
\end{equation}
where $\mathbf{W}_{\alpha}$ and $\mathbf{W}_{\alpha'}$ are weight matrices for computing $\mathcal{Q}$ and $\mathcal{K}$ from the respective regions $r$ and $r'$ ;  $\mathbf{W}_{a}$ is their nonlinear combination; $\mathbf{b}_a$ and $\mathbf{b}_\alpha$ are the biases. The attention-driven context vector $\mathbf{v}_r$ infers the \textit{strength} of $f_r$ \textit{conditioned on itself and its neighborhood} (Fig. \ref{fig:full_model}(c)). The final context refinement weight vector $\mathbf{v}$ representing all the regions $\mathcal{R}$ is computed using an element-wise sigmoid activation function $\sigma(.)$ over a weighted summation of all $\mathbf{v}_r\in \mathcal{V}$ using an attention importance weight $w_r$. 
\begin{equation}
\begin{split}
    \mathbf{v}&=\sigma\left(\sum_{r=1}^{\mathcal{R}}\bar{\mathbf{v}}_r w_r\right), \text{ where } \bar{\mathbf{v}}_r = \text{GMP}(\mathbf{v}_r) \text{ and }\\
    w_r&=\text{Softmax}(\mathbf{W}_\beta \bar{\mathbf{v}}_r + \mathbf{b}_\beta),
\end{split}
\end{equation}
where GMP is the \textit{global max-pooling}, weight matrix $\mathbf{W}_\beta$ and bias $\mathbf{b}_\beta$ are learnable parameters. It is similar to the approach in \cite{BahdanauCB14} for solving machine translation problems where the model searches for parts of a source sentence relevant to predicting a target word. However, our model does not consider the sequential information but learns to emphasize latent representations of multiple regions by joint learning. To improve the gradient flow, this refinement of weight vector $\mathbf{v}$ is used to enhance the relation-aware image-level feature $f_t$ in (\ref{eq_3}) via a skip connection \textit{i.e.}, $\bar{f}_t = f_t + f_t \otimes \mathbf{v}$  before passing it to a \texttt{Softmax} layer for estimating the target class probability $\hat{y}_n$ for the image $I_n$. The learnable attentional context refinement parameter in (\ref{eq_1}) is thus $\theta_c=\{\mathbf{W}_{\alpha}, \mathbf{W}_{\alpha'}, \mathbf{W}_{a}, \mathbf{b}_a,\mathbf{b}_\alpha, \mathbf{W}_\beta, \mathbf{b}_\beta\}$.
\section{Experimental Results and Discussions} \label{experiments}
We first present the datasets, experimental details followed by comparison to the SotA. Then, we analyze our model's complexity followed by qualitative analysis to get an insight into the  decision-making process. Finally, we conduct ablation studies to evaluate its key components and parameters. 

\begin{table}
\begin{center}
 \caption{Statistics of the Datasets Used in Our Experiments. Accuracy (\%) of the Best SotA and Our \texttt{SR-GNN}.  }
 \label{table:overall_accuracy}
\begin{tabular}{|l| c c c c|}
 \hline
Dataset &  \#Train / \#Test & \#Class & SotA & \texttt{SR-GNN} \\
    \hline
Aircraft & 6,667 / 3,333 &100 &94.9 \cite{behera2021context}  & \textbf{95.4} \\ 
CUB-200 & 5,994 / 5,794 &200 &91.8 \cite{behera2021context} &\textbf{91.9} \\ 
Cars & 8,144 / 8,041 &196 & 95.7 \cite{behera2021context} &\textbf{96.1} \\ 
 Dogs & 12,000 / 8,580 &120 &97.1\cite{ge2019weakly} &\textbf{97.3} \\ Flowers & 2,040 / 6,149 &102  &97.7 \cite{chang2020devil} &\textbf{97.9} \\ 
NABirds &23,929 / 24,633 &555 &91.0 \cite{behera2021context} &\textbf{91.2} \\
\hline
Stanford-40 & 4,000 / 5,532 &40 &97.8 \cite{bera2021attend} &\textbf{98.8} \\
PPMI-24 & 2,110 / 2,099 &24 &98.6 \cite{behera2020regional} &\textbf{98.9} \\

\hline
\end{tabular}
\end{center}
 \vspace{-0.5cm}
\end{table}
\subsection {Datasets}
Our model avoids object/part bounding box labels for evaluation on eight benchmark datasets (fine-grained objects/ human-actions, detailed in Table \ref{table:overall_accuracy}): Aircraft \cite{maji2013fine}, Caltech-UCSD Birds (CUB-200) \cite{wah2011caltech}, Stanford Cars \cite{krause20133d}, Stanford Dogs \cite{khosla2011novel}, Oxford Flowers  \cite{nilsback2008automated},  NABirds \cite{van2015building}, Stanford-40 actions \cite{yao2011human}, and People Playing Musical Instruments (PPMI-24) \cite{yao2010grouplet}. The top-1 accuracy (\%) is used for the evaluation. 

\begin{table*}[t]
 \centering
 \caption{Accuracy (\%) Comparison with the Most Recent Top-10 SotA Methods. {\color{magenta}*} Involves Transfer/Joint Learning Strategy for Objects/Patches/Regions Involving More Than One Dataset (Target and Secondary). {\color{orange}$\mathsection$} applies Vision Transformer. {\color{green}$^\dagger$} Uses Additional Text Description. The Last Three Rows Show the Accuracy of Base CNN (Xception \cite{chollet2017xception}), \texttt{SR-GNN} without the Attentional Refinement (``W/o Refine'') Module (Fig. \ref{fig:full_model}(c)), and Full \texttt{SR-GNN} Model. The Following Abbreviations are Used to Denote Various CNN Backbones: RN34/RN50/RN101/RN152 for ResNet-34/50/101/152; In-v3 for Inception-v3; BCNN for Bilinear CNN; Xcep for Xception, DN161/DN201 for DenseNet-161/201; ViT-B for Vision Transformer-B-16; Swin for Swin Transformer with Swin-Base-224; GN for GoogleNet; WRN for Wide Residual Networks; SE for Squeeze-and-Excitation Networks. Coding implies Encoding/Codebook; Param as Parametric, and Fusion for Multiple CNNs.}
  \label{table:sota_comp}  
 \begin{small}
  \begin{tabular}{|p{1.7 cm} p{8 mm} p{5 mm}|p{1.7 cm} p{8 mm} p{5 mm}|p{1.8 cm} p{8 mm} p{5 mm}|p{1.9 cm} p{13 mm} p{5 mm}|}
    \hline
      \multicolumn{3}{|c|}{\textbf{Aircraft}}&
      \multicolumn{3}{c|}{\textbf{CUB-200}}&
      \multicolumn{3}{c|}{\textbf{Cars}}&       
      \multicolumn{3}{c|}{\textbf{Dogs}} 
      \\
{Method} & {CNN} &  {Acc} & {Method} & {CNN} &  {Acc} & {Method} & {CNN} &  {Acc} & {Method} & {CNN} &  {Acc}   \\ 
      \hline 
DCL \cite{chen2019destruction} & RN50 & 93.0 & 
CSC \cite{wang2020category} & RN50 & 89.2   & DCL \cite{chen2019destruction} & RN50 & 94.5 &
Cross-X \cite{luo2019cross}  & RN50 & 88.9    \\
GCL \cite{wang2020graph} & RN50 & 93.2 & 
DAN \cite{hu2019see} & In-v3 & 89.4 & S3Ns \cite{ding2019selective} & RN50 & 94.7 & MRDMN \cite{xu2021multiresolution}  & RN50 & 89.1   \\
CAMF{\color{orange}$^\mathsection$} \cite{miao2021complemental} & Swin & 93.3 & BARM \cite{liu2019bidirectional}& DN161 & 89.5 &
TrnFG{\color{orange}$^\mathsection$}\cite{he2021transfg} & ViT-B & 94.8 & APIN \cite{zhuang2020learning} & RN101  &90.3  \\
PMG \cite{chang2021your} & RN50 & 93.6 &  
GaRD\cite{zhao2021graph} & RN50 & 89.6 &  
CSC \cite{wang2020category} & RN50& 94.9 &  
ViT{\color{orange}$^\mathsection$} \cite{dosovitskiy2020image} & ViT-B &	91.7 \\ 
SCAP \cite{liu2021learning} & RN50 & 93.6 &  
PMG \cite {chang2021your} & RN50 & 89.9 & 
GaRD\cite{zhao2021graph} & RN50 & 95.1 &
DAN \cite {hu2019see} & In-v3 &92.2    \\
CSC\cite{wang2020category} & RN50 & 93.8 & 
APIN \cite{zhuang2020learning} &DN161 & 90.0  & PMG \cite{chang2021your} &RN50 & 95.1  & 
TrnFG{\color{orange}$^\mathsection$}\cite{he2021transfg} & ViT-B &92.3 \\ 
APIN \cite{zhuang2020learning}  &DN161 & 93.9  &
CPM$^{\color{magenta}*}$ \cite{ge2019weakly} & GN  & 90.4 & 
APIN\cite{zhuang2020learning} & DN161 & 95.3 & 
CAMF{\color{orange}$^\mathsection$} \cite{miao2021complemental} & Swin & 92.8  \\ 
APCN \cite{ding2021ap} & RN50 & 94.1  & CAMF{\color{orange}$^\mathsection$} \cite{miao2021complemental} & Swin &  91.2  & CAMF{\color{orange}$^\mathsection$} \cite{miao2021complemental} & Swin & 95.3 & WARN \cite{lopez2020pay} & WRN50 &92.9   \\
GaRD\cite{zhao2021graph} & RN50 & 94.3 & TrnFG{\color{orange}$^\mathsection$}\cite{he2021transfg} & ViT-B & 91.7 & APCN \cite{ding2021ap}  &RN50 & 95.4   & 
CAP \cite{behera2021context} & Xcep & 96.1 \\
 CAP \cite{behera2021context} & RN50 & 94.9 & CAP \cite{behera2021context} & Xcep & 91.8  &  CAP \cite{behera2021context} & Xcep & 95.7  &  CPM$^{\color{magenta}*}$ \cite{ge2019weakly} & GN & 97.1  \\
    \hline
    Base CNN & Xcep & 79.5   & Base CNN & Xcep &75.6  & Base CNN & Xcep &84.8 & Base CNN  & Xcep  & 82.7  \\
    W/o Refine &  & 93.5   &W/o Refine &   &90.2 &W/o Refine & &93.7  &W/o Refine &  &96.5 \\
    \textbf{\texttt{SR-GNN}} &  & \textbf{95.4} &\textbf{\texttt{SR-GNN}} &  & \textbf{91.9} &\textbf{\texttt{SR-GNN}} &  & \textbf{96.1} & \textbf{\texttt{SR-GNN}} &  & \textbf{97.3}   \\
    \hline
    \hline
 \multicolumn{3}{|c|}{\textbf{Flowers}} &
  \multicolumn{3}{|c|}{\textbf{NABirds} \cite{van2015building}} & \multicolumn{3}{|c|}{\textbf{Stanford-40} \cite{yao2011human}} & \multicolumn{3}{|c|}{\textbf{PPMI-24} \cite{yao2010grouplet}}\\
     {Method} & {CNN} &  {Acc} &  {Method} & {CNN} & {Acc} & {Method} & {CNN} & {Acc} & {Method} & {CNN} & {Acc} \\ 
      \hline 
 MGE\cite {zhang2019learning} & RN50  & 95.9 &
 Cross-X \cite {luo2019cross} & SE  & 86.4 & 
  CAM  \cite{zhou2016learning} & GN &72.6 & 
  LLC \cite{wang2010locality} & Coding & 39.7 \\
  PBC \cite{huang2016pbc} & GN & 96.1 & 
  SPA \cite{ali2019parametric} & Param  &87.6  &
  ProCRC \cite{cai2016probabilistic} & VGG19 & 80.9 & 
  ScSPM \cite{yang2009linear} & Coding  & 41.5\\
IntAct \cite {xie2016interactive} & VGG19   & 96.4 & DSTL$^{\color{magenta}*}$  \cite{cui2018large} & In-v3 &87.9 & 
Introsp \cite{rosenfeld2016visual} &VGG16 & 81.7 & 
CSDL \cite{gao2013learning} & Coding  & 48.8  \\
SJFT$^{\color{magenta}*}$  \cite{ge2017borrowing} & RN152  & 97.0 & GaRD\cite{zhao2021graph} & RN50 & 88.0  & 
PKPCR \cite{lan2018prior} & VGG19 & 82.4 & 
Exemplr \cite{Hu2013recognising} &  Dictionary  &49.3 \\
OPAM$^{\color{magenta}*}$  \cite{peng2017object} & VGG  & 97.1 & 
  APIN \cite{zhuang2020learning}  &DN161  &88.1  & 
Concepts  \cite{RosenfeldU16b} & VGG16 & 83.1 & 
VLAD \cite{zhang2016towards} & Coding  &50.7  \\
Cos.Ls$^{\color{magenta}*}$\cite{barz2020deep} & RN50 &	97.2 & CSPE \cite {korsch2019classification} & In-v3   & 88.5 & 
Color  \cite{lavinia2020new} & Fusion &84.2 &  
Color \cite{lavinia2020new} & Fusion &65.9 \\ 
PMA$^{\color{green}\dagger}$  \cite{song2020bi} & VGG16 &97.4  & MGE \cite{zhang2019learning} & RN101  &88.6 & 
$\alpha$-pool \cite{simon2017generalized} & VGGM &86.0 & 
DSFNet \cite{li2020deep} & RN34  &72.3 \\
DSTL$^{\color{magenta}*}$  \cite{cui2018large} & In-v3  & 97.6  & 
ViT{\color{orange}$^\mathsection$} \cite{dosovitskiy2020image} & ViT-B &	89.9 & Implicit \cite{simon2018whole} & RN50 & 87.7 & 
Coding \cite{li2018reassessing} &  NASNet & 82.3  \\
MCL$^{\color{magenta}*}$ \cite{chang2020devil} & BCNN  &97.7 &  
TrnFG{\color{orange}$^\mathsection$}\cite{he2021transfg} & ViT-B  & 90.8 &  RAN \cite{behera2020regional} & RN50 &97.4 & 
 AG-Net \cite{bera2021attend} & RN50 & 98.2 \\
CAP \cite{behera2021context} &Xcep  &97.7 & CAP \cite{behera2021context} &Xcep &91.0 & AG-Net \cite{bera2021attend} & RN50 &97.8 & RAN \cite{behera2020regional} & DN201 & 98.6\\
         \hline
 Base CNN & Xcep & 91.9 &   Base CNN & Xcep& 68.1 & Base CNN &Xcep  &  80.0 & Base CNN  & Xcep & 79.3\\
 W/o Refine &  &97.1 & W/o Refine &  &89.9 & W/o Refine &  & 97.9 & W/o Refine &  & 97.9 \\
\textbf{\texttt{SR-GNN}} &  & \textbf{97.9}  & \textbf{\texttt{SR-GNN}} &  & \textbf{91.2} & \textbf{\texttt{SR-GNN}} &  & \textbf{98.8} &\textbf{\texttt{SR-GNN}} & & \textbf{98.9} \\
    \hline  
  \end{tabular}
  \end{small}
\vspace{-0.5cm}
\end{table*}
\vspace{-0.3 cm}
\subsection {Implementation Details} TensorFlow 2.0 is used for implementation. Like CAP \cite{behera2021context}, we use Xception \cite{chollet2017xception} as a backbone CNN. The output dimension 7$\times$7$\times$2048 is upsampled to 42$\times$42$\times$2048
for region pooling (Fig. \ref{fig:full_model}(a)). Region proposal in \cite{behera2020regional} is used with a HOG cell-size of 14$\times$14 to generate 27 optimal region proposals ($\mathcal{R}$), consisting of a minimum region of 2 cells to a maximum of the full image. The region-pooling size of $w$$=$$h$$=$$7$ is used. The feature transformation module (Fig. \ref{fig:full_model}(b)) consists of two GNN layers with an optimal output size of 1024. Each layer contains a single-layer MLP with the teleport probability $\alpha$=$0.3$. The number of channels  is  kept the same ($C$=$2048$) as  Xception output. Source codes of \texttt{SR-GNN} will be available via the GitHub repository at \textit{https://github.com/ArdhenduBehera/SR-GNN}.

\vspace{ -0.2  cm}
\subsection {Experimental Settings}
Pre-trained ImageNet weights are used to initialize base CNN for faster convergence with the size of images being 256$\times$256. We apply the data augmentation of random rotation ($\pm$15 degrees), random scaling (1$\pm$0.15), and then random cropping to select the image size of 224$\times$224. The Stochastic Gradient Descent (SGD) is used to optimize the categorical cross-entropy loss function with an  initial learning rate of $10^{-3}$ and multiplied by 0.1 after every 50 epochs. The model is trained for 150 epochs with a mini-batch size of 8 using NVIDIA Titan V GPU (12GB). 

\vspace{ -0.3 cm}
\subsection{Performance Comparisons with State-of-the-Art Methods} 
The accuracy (\%) of \texttt{SR-GNN} over eight datasets and its comparisons to the previous best results (according to the best of our knowledge) are given in Table \ref{table:overall_accuracy}. 
The accuracy of \texttt{SR-GNN} over each dataset is compared with those of the top-10 SotA methods in the literature in Table \ref{table:sota_comp}. These SotA methods are based on attention mechanism \cite{behera2021context, behera2020regional, chang2020devil, zhang2019learning, lopez2020pay, ding2019selective, liu2019bidirectional}, discriminative object-part localization \cite {ge2019weakly, wang2019weakly, chen2019destruction}, 
mutual reinforcement learning \cite{liu2019bidirectional}, GNN \cite{wang2020graph, wang2020category, zhao2021graph}, vision transformers \cite{dosovitskiy2020image, liu2021swin, he2021transfg, miao2021complemental}, etc.  \texttt{SR-GNN} clearly outperforms all previous methods over eight datasets and their accuracy gain over each dataset is given in parenthesis: Aircraft (0.5\%), CUB-200 (0.1\%), Cars (0.4\%), Dogs (0.2\%), Flowers (0.2\%), NABirds (0.2\%), Stanford-40 (1.0\%), and PPMI-24 (0.3\%). These margins of improvements are very significant since FGVC is a challenging task to discriminate various subcategories. This is evident from the top-10 SotA accuracies over each dataset (Table \ref{table:sota_comp}) that achieve the successive marginal gain between 0.1-0.3\% over the past 2-3 years. For example, a cumulative gain of 1.2\% is achieved by the top-10 SotA methods over the Cars dataset within the past 3 years with an average of 0.13\% (9 successive differences). Our gain of 0.4\% over CAP is thus significantly higher. Similarly, a cumulative gain of 1.9\% (DCL to CAP) is achieved over the Aircraft dataset (average: 0.21\%). Our \texttt{SR-GNN} gains 0.5\% in comparison to CAP and is thus also significant. Moreover, some methods attain similar accuracy, \textit{e.g.}, GaRD \cite{zhao2021graph} and PMG \cite {chang2021your} on Cars: 95.1\%. \texttt{SR-GNN} outperforms over eight datasets with a gain of between 0.1\% to 1.0\%.  Moreover, our accuracy gain is 0.1\% - 1.2\% over six FGVC datasets over CAP currently at the top of the leaderboard. 
Many SotA methods are weakly-supervised such as localization of objects/parts using pre-trained object/part detector and/or proposals using semantic segmentation (\textit{e.g.}, mask R-CNN 
or Grad-CAM). The process often includes at least two steps: firstly, detect the weakly-supervised regions and then apply the fine-grained recognition. Moreover, additional secondary datasets (\textit{e.g.}, COCO in \cite{ge2019weakly} for Dogs: 97.1\%, and ImageNet in \cite{chang2020devil} for Flowers: 97.7\%) are used for further training to achieve SotA accuracy \cite{ge2019weakly}. In sharp contrast, our \texttt{SR-GNN} is a single-step process that is trained end-to-end using only the target datasets and is thus computationally efficient and easy to implement.  

\begin{table}
\begin{center}
 \caption{Accuracy (\%) of our \texttt{SR-GNN} using ResNet-50 base CNN with different sizes of images for FGVC.} 
 \label{table:RN50}
\begin{tabular}{|p{0.5cm}|p{0.68 cm}p{0.68 cm}p{0.68 cm}|p{1.03 cm}p{1.03 cm}p{1.2 cm}|}
 \hline
\multicolumn{1}{|c|}{}&
\multicolumn{3}{c|}{224$\times$224, CAP \cite{behera2021context}}&   \multicolumn{3}{c|}{448$\times$448 size} \\
 & Airc. &CUB &NAB&Cars&Flowers&Dogs  \\
    \hline
SotA &\textbf{94.9} &90.9 & \textbf{88.8}&95.4\cite{ding2021ap}  &96.8\cite{chang2020devil} & {88.9} \cite{luo2019cross}\\   \hline
Ours & {94.8} &\textbf{91.0} & \textbf{88.8}  &\textbf{95.8} &\textbf{98.0} &\textbf{97.1} \\
\hline
\end{tabular} 
 \end{center}
  \vspace{-0.6cm}
\end{table}

We have explicitly compared the performance of our method with the SotA ones implemented with ResNet-50 backbone using image sizes of 224$\times$224 and 448$\times$448. The results are given in Table \ref{table:RN50}. It is evident that CAP performs the best among the existing methods on Aircraft (94.9\%), CUB (90.9\%), and NABirds (88.8\%) with an image size of 224$\times$224 and in  this case, our method achieves a very competitive results ($\pm$ 0.1\%).  Alternatively, AP-CNN (Cars: 95.4\%) \cite{ding2021ap}, MCL (Flowers: 96.8\%) \cite{chang2020devil}, and Cross-X (Dogs: 88.9\%) \cite{luo2019cross} use an image size of 448$\times$448 with ResNet-50 instead. With such image size, our \texttt{SR-GNN} achieves 95.8\% on Cars, and 98.0\% on Flowers; and gains a margin of 8.2\% over Cross-X on Dogs (\texttt{SR-GNN}: 97.1\%). Though, we attain an accuracy of 97.1\%  over Dogs as CPM \cite{ge2019weakly}, the latter applies a complex training process using GoogleNet backbone. Clearly, our \texttt{SR-GNN} outperforms many SotA methods with an image size of 224$\times$224 over all the datasets using Xception or ResNet-50 backbone. 

\begin{table}[h]
\vspace{ -0.3cm}
\begin{center}
 \caption{ Comparison of our \texttt{SR-GNN} with Vision Transformers. 
 Model Complexity is Given in Parameters (Million) and GFLOPs (Billion), for Input-size 384$\times$384, as Provided in \cite{liu2021swin}. Top: input-size: 224$\times$224, Mid: 448$\times$448, 
 and bottom: 224$\times$224. }
  \label{table:ViT}
 \vspace{ -0.3cm}
\begin{tabular}{|p{14 mm}|p{11.0 mm} p{2.5 mm} p{2.8 mm} p{2.8 mm} p{3.2 mm} p{3.3 mm}|c|}
 \hline
Method & Transformer & CUB & Car  & Dog & NAB  & Air & Parm (GFlop)\\
    \hline
Swin \cite{liu2021swin} & Swin-224 & 89.7 &94.2  &91.8 &- &91.0 & 88 (15.4)\\
CAMF \cite{miao2021complemental} & Swin-224 & 90.9 &94.8  &92.6 &-  &92.9 &-\\
\hline
ViT \cite{dosovitskiy2020image} & ViT-B-16 & 90.3 &93.7  &91.7 &89.9 &-   &86 (55.4)\\
TrnFG \cite{he2021transfg} & ViT-B-16 & 91.7 &94.8  &92.3 & 90.8 &- &-\\
Swin \cite{liu2021swin} & Swin-224 & 90.7 &94.8  &92.5 &-  &93.0 & 88 (47.0)\\
CAMF \cite{miao2021complemental} & Swin-224 & 91.2 &95.3  &92.8 &-  &93.3 &-\\
\hline
\textbf{\texttt{SR-GNN}} &- &\textbf{91.9} & \textbf{96.1} & \textbf{97.3} & \textbf{91.2}  & \textbf{95.4} & \textbf{30.9 (9.8)}\\
\hline
\end{tabular}
 \end{center}
  \vspace{- 0.3 cm}
\end{table}
More recently, vision Transformer such as ViT \cite{dosovitskiy2020image} uses fixed-size patches, and Swin Transformer \cite{liu2021swin} applies a shifted window scheme to construct a hierarchical representation of patches. In contrast, we use multi-scale regions leveraging  GNN for subtle discrimination. ViT often requires large-scale training datasets (\textit{e.g.}, JFT-300M, ImageNet-22K, etc.) and then fine-tuning on a target dataset to perform well for FGVC. Unlike CNNs, a Transformer is built with a relatively complex, larger, and heavier architecture. For example, ViT base model consists of 86M parameters and 55.4B GFLOPs (Table \ref{table:ViT}). Recently, CAMF \cite{miao2021complemental} has demonstrated that a Swin Transformer can achieve better performance than pure ViT with an image size of 448$\times$448. Whereas our method (224$\times$224) outperforms vision Transformers with both sizes of 448$\times$448 and 224$\times$224 with a clear margin on five FGVC datasets (Table \ref{table:ViT}). Our gain (in parenthesis) on each dataset is: Dogs (4.5\%), Aircraft (2.1\%), Cars (0.8\%), CUB (0.7\%), and NABirds (0.4\%). Moreover, our method incurs significantly less computational overload than the Transformers. For example, \texttt{SR-GNN} (224$\times$224) consists of 30.9M parameters and 9.8B GFLOPs. This is 57.1M parameters and 37.2B GFLOPs lesser than the Swin Transformer. Furthermore, \texttt{SR-GNN} expeditiously outperforms these SotA models with a notable margin with an end-to-end training and simple evaluation protocol avoiding additional secondary data and resource constraints, justifying its wider adaptability. 
\begin{table} [t]
\begin{center}
 \caption{Accuracy (\%) of \texttt{SR-GNN} with Other SotA Base CNNs Instead of Xception Using the Same Test Setup (224$\times$224 Img-Size, $\alpha=0.3$ \& $\mathcal{R}=27$). CAP is Used as Backbone by Replacing the CNN Feature Map and Region Proposals (Fig. \ref{fig:full_model}(a)), and  Relation-aware Feature Transformation by GNN (Fig. \ref{fig:full_model}(b)).}
 \label{table:baseCNN}
\begin{tabular}{|l| c c c |c|}
 \hline
Dataset & ResNet-50 & Inception-V3 & NASNetMobile & CAP\\
\hline
  Aircraft &94.8 & 94.4 & 94.4 & 95.1 \\
  CUB &91.0 & 90.7 & 90.8 & 91.9 \\
  Cars & 93.1 & 94.1 & 95.6 & 95.8\\
  Dogs & 93.2 & 95.3 &95.9 & 96.6\\ 
  Flowers &97.4 & 97.4 & 97.3 & 97.8\\
  NABirds &88.8 & 89.2 & 88.6 & 90.7\\
\hline
\end{tabular}
 \end{center}
 \vspace{-0.4 cm}
\end{table}
\begin{table*}[h]
\begin{center}
\caption{\texttt{SR-GNN}'s Capacity and Computational Overhead for Different Regions Using an NVIDIA Titan V GPU (12GB). }  
\label{table:time}
\begin{tabular}{|l |c |c |c |c |c |c |c|}
 \hline
\#No. of &\#Trainable params &GFLOPs &Per-image inference time  &\multicolumn{4}{|c|}{Training time (batch size 8) in $\sim$hours}\\
    \cline{5-8}
    Regions &in millions ($\sim$M) &in billions ($\sim$B) &in millisecond ($\sim$ms) &Aircraft &Cars &Dogs &Flowers \\
    \toprule
    11 &30.9 &9.4 &3.9 &3.5 &8.4 &13.4 &1.9\\
    19 &30.9 &9.6 &5.0 &4.1 &10.2 &15.2 &2.6\\
    27 &30.9 &9.8 &5.0 &4.5 &11.2 &17.0 &2.8\\
    36 &30.9 &10.1 &6.0 &5.0 &12.6 &18.3 &3.1\\
   \bottomrule
\end{tabular}
 \end{center}
 \vspace{- 0.5 cm}
\end{table*}

\noindent\textbf{Performance using other SotA base CNNs:} \texttt{SR-GNN} uses the lightweight Xception\cite{chollet2017xception} as a backbone to extract CNN features for further processing. It can easily be integrated into other CNN backbones with a little computational overhead. In order to verify this, we have evaluated our \texttt{SR-GNN} using three different SotA CNN backbones: ResNet-50 \cite{he2016deep}, Inception-V3 \cite{szegedy2016rethinking}, and NASNetMobile \cite{zoph2018learning}, with an image resolution of 224$\times$224 over the six FGVC datasets. The results are provided in Table \ref{table:baseCNN}. Our method using these backbones is very similar to the one using Xception and consistently outperforms the SotA approaches in Table \ref{table:sota_comp} with the same backbones. However, \texttt{SR-GNN}'s accuracy using Xception is slightly higher than the similar backbones such as Inception-V3 and ResNet-50, and is thus our optimal choice. The main reason could be the architectural design of Xception in which depth-wise separable convolutions are used within the Inception module. It is built with a linear stack of depth-wise separable convolutional layers with residual connections. The design leads to a better representation of high-level CNN features in comparison to the ResNet-50 and Inception-V3 architectures. NASNetMobile \cite{zoph2018learning} is a lightweight model that is designed for mobile and embedded vision systems. It involves significantly less computational cost. From the performance (Table \ref{table:baseCNN}), the accuracy using this mobile architecture is as competitive as the standard CNNs. Generally, many approaches consider ResNet-50 and our method significantly outperforms those using the same ResNet-50 backbone, as evident from Table \ref{table:sota_comp}-\ref{table:RN50}.  A similar trend can be observed for Inception-V3.

We have also evaluated our model by replacing the region proposals (Fig. \ref{fig:full_model}(a)) and feature transformation using GNN (Fig. \ref{fig:full_model}(b)) with the CAP \cite{behera2021context} model. The results are given in Table \ref{table:baseCNN}. The performance is aligned with the original CAP \textit{i.e.}, the accuracy (Aircraft: 95.1\%, CUB: 91.9\%, Cars: 95.8\%, Dogs: 96.6\%, Flowers: 97.8\% and NABirds: 90.7\%) is superior to SotA methods including CAP, except NABirds on which CAP’s accuracy is 91.0\% \cite{behera2021context}. Nevertheless, the accuracy on NABirds (90.7\%) is still superior to the other approaches in Table \ref{table:sota_comp}. Moreover, \texttt{SR-GNN} surpasses these results over Aircraft (0.3\%), Cars (0.3\%), Dogs (0.7\%), Flowers (0.1\%), and NABirds (0.5\%) with a clear margin and achieves the same accuracy of 91.9\% over CUB. This justifies the benefits of our novel GNN-driven relation-aware feature transformation (Fig. \ref{fig:full_model}(b)) and attentional context refinement (Fig. \ref{fig:full_model}(c)) modules, and their significance in enhancing FGVC accuracy. Also, \texttt{SR-GNN} is lighter than CAP requiring 3.3M and 0.4B fewer (Table \ref{table:Complx2}) parameters and GFLOPs, respectively, implying its computational efficiency. 

\noindent\textbf{Performance on Human-Object Interactions:} To demonstrate our method under general data diversity, we tested our \texttt{SR-GNN} on the Stanford-40 actions \cite{yao2011human} and People Playing Musical Instruments (PPMI-24) \cite{yao2010grouplet} datasets, representing fine-grained human-object interactions. Its accuracy is 98.8\% on Stanford-40 and 98.9\% on PPMI-24. It outperforms the best results attained by AG-Net (Stanford-40: 97.8\%) \cite{bera2021attend} and RAN (PPMI-24: 98.6\%) \cite{behera2020regional}. Our model also \textit{learns the importance} (weight) of a region via a novel attentional context to refine the transformed features. Whereas, CAP uses LSTM to learn spatial arrangement between regions, and an LSTM-driven feature encoding to aggregate the information from its hidden states. AG-Net and RAN learn features from each region independently without feature interaction and use a Squeeze-and-Excitation block to extract features followed by an attention module.

A generalized conventional average and bilinear pooling, namely $\alpha$-pooling \cite{simon2017generalized}, achieves an accuracy of 86.0\%  over the Stanford-40 actions. The $\alpha$-pooling enhances the performance in implicit pose normalization (87.7\%) \cite{simon2018whole} over this dataset, and achieves SotA accuracy compared to other prior works. However, our method attains an impressive margin (11.1\%) over this work. Even without feature refinement (W/o Refine), our model achieves the best result over this dataset. 

Some prior methods have extracted traditional/hand-crafted feature descriptors (\textit{e.g.}, SIFT) and applied bag-of-feature encoding techniques \cite{zhang2016towards,gao2013learning} over which deep features attain better performance. A reinforcement learning method, DSFNet \cite{li2020deep} captures the global discriminative information and fine-grained representations on PPMI-24. Hierarchical learning based on the spatial pyramid is presented in \cite{li2018reassessing}. Their pre-trained networks achieve better performance than the other existing approaches on this dataset. However, our method gains a high margin of 16.6\% over their approach.      

\noindent \textit{Comparison using mAP evaluation metric}: Many works consider the mAP (mean average precision) as an evaluation metric on the above-mentioned two datasets. For a fair comparison, we have  evaluated the performance of \texttt{SR-GNN} using mAP. Our approach achieves 96.6\% mAP on Stanford-40 which is 0.4\% improved over AG-Net (96.2\%) \cite{bera2021attend}. Similarly, we have attained higher mAP in comparison to the human mask loss (94.1\%) \cite{liu2018loss}, part-action network (91.2\%) \cite{zhao2017single}, and many recent works on Stanford-40. We have achieved improved mAP (95.3\%) on PPMI-24 over existing works such as VLAD spatial pyramid (81.3\%) \cite{yan2017action}, 10-model color fusion (65.9\%) \cite{lavinia2020new}, and the others. While, the mAP of \texttt{SR-GNN} (95.3\%) on PPMI-24 is 1.4\% lower than RAN (96.7\%), it attains 0.3\% gain in accuracy. 

The accuracy of our model is compared without the attentional context refinement module (Fig. \ref{fig:full_model}(c)) and is given in the 2nd-last row of Table \ref{table:sota_comp}. A notable observation is that even without context refinement  (``W/o Refine''), \texttt{SR-GNN} outperforms many methods tested on Dogs (96.5\%, 2nd-best), NABirds (89.9\%, 3rd-best, same as ViT), Stanford-40 (97.9\%, best), and PPMI-24 (97.9\%, 3rd-best). Also, the accuracies on Aircraft (93.5\%), CUB (90.2\%), and Flowers (97.1\%)  are competitively retained within the accuracies of the top-10 SotA methods. Our attentional context refinement module enhances the overall accuracy on diverse datasets, while avoiding additional parts-level annotations, vision Transformers, secondary datasets and/or pre-trained subnetworks to enhance the accuracy. 
%
\vspace{ -0.4 cm}
\begin{table}[h]
\begin{center}
 \caption{Computational Complexity Comparison of the proposed \texttt{SR-GNN} with State-of-the-Arts.}
  \label{table:Complx2}
\begin{tabular}{|lccc|}
 \hline
Method & Param (M) & GFLOPs (B) & Inf. Time/img (ms) \\
    \hline
AG-Net \cite{bera2021attend} & 54.8 & 10.4 &5.2 \\
MRDMN-L \cite{xu2021multiresolution} & 51.2 &14.0 & \underline{4.9}\\
TASN \cite{zheng2019looking} & 37.3  &21.9 &7.5 \\
CAP \cite{behera2021context} & 34.2 &10.2 &\textbf{4.2}\\ 
\hline
Base CNN & 20.9 & 9.2 & 2.7\\
W/o Refine &  24.4 & 9.3 & \underline{4.9} \\
\textbf{\texttt{SR-GNN}}  &\textbf{30.9} & \textbf{9.8} & 5.0\\
\hline
\end{tabular}
 \end{center}
 \vspace{- 0.7 cm}
\end{table}

\begin{figure*}
\vspace{ -0.5 cm}
    \centering
    \subfloat[Base CNN within \texttt{SR-GNN}]{\includegraphics[width=0.22\textwidth] {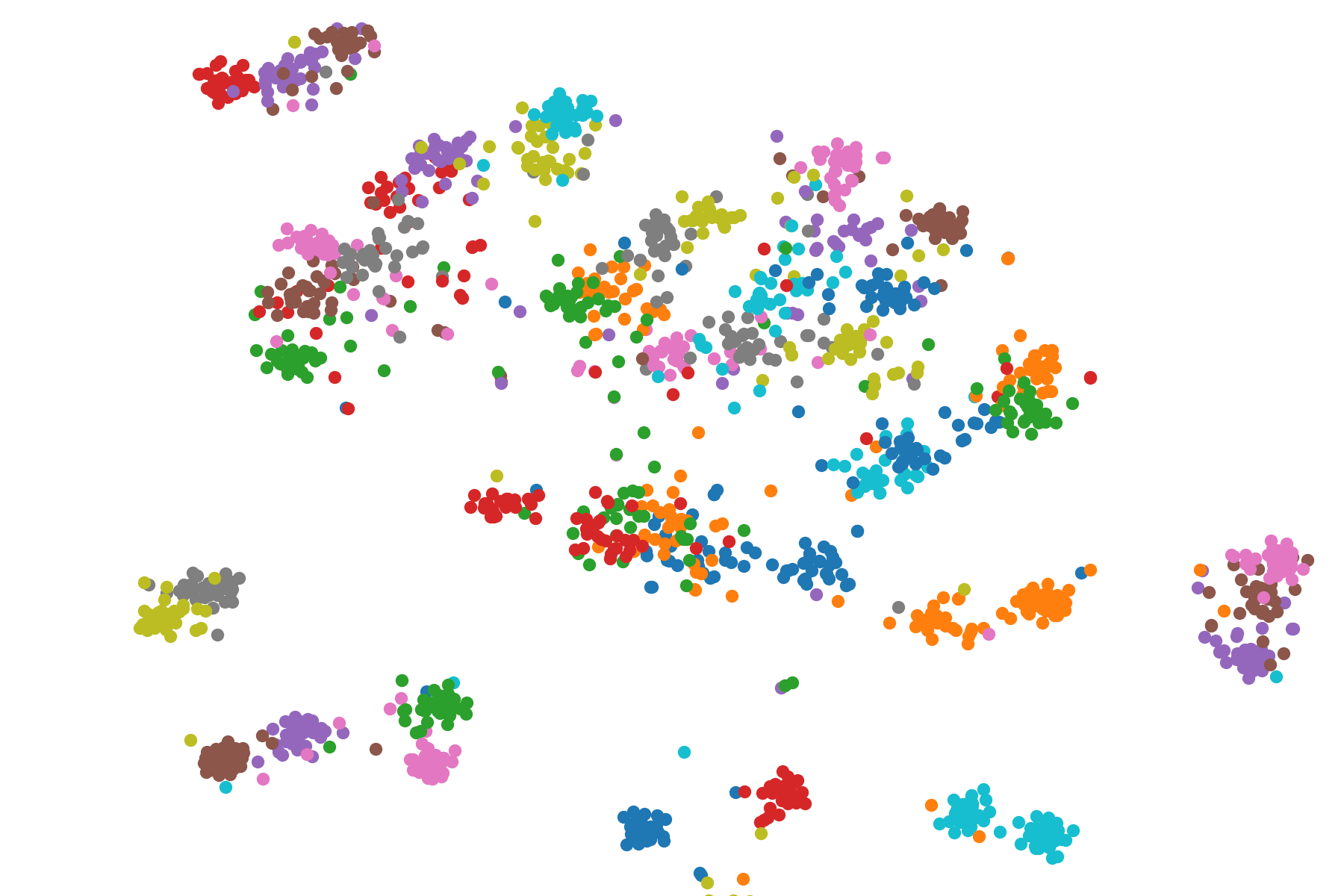}}\hfill
    \subfloat[Transformed Feature $f_t$]{\includegraphics[width=0.22\textwidth] {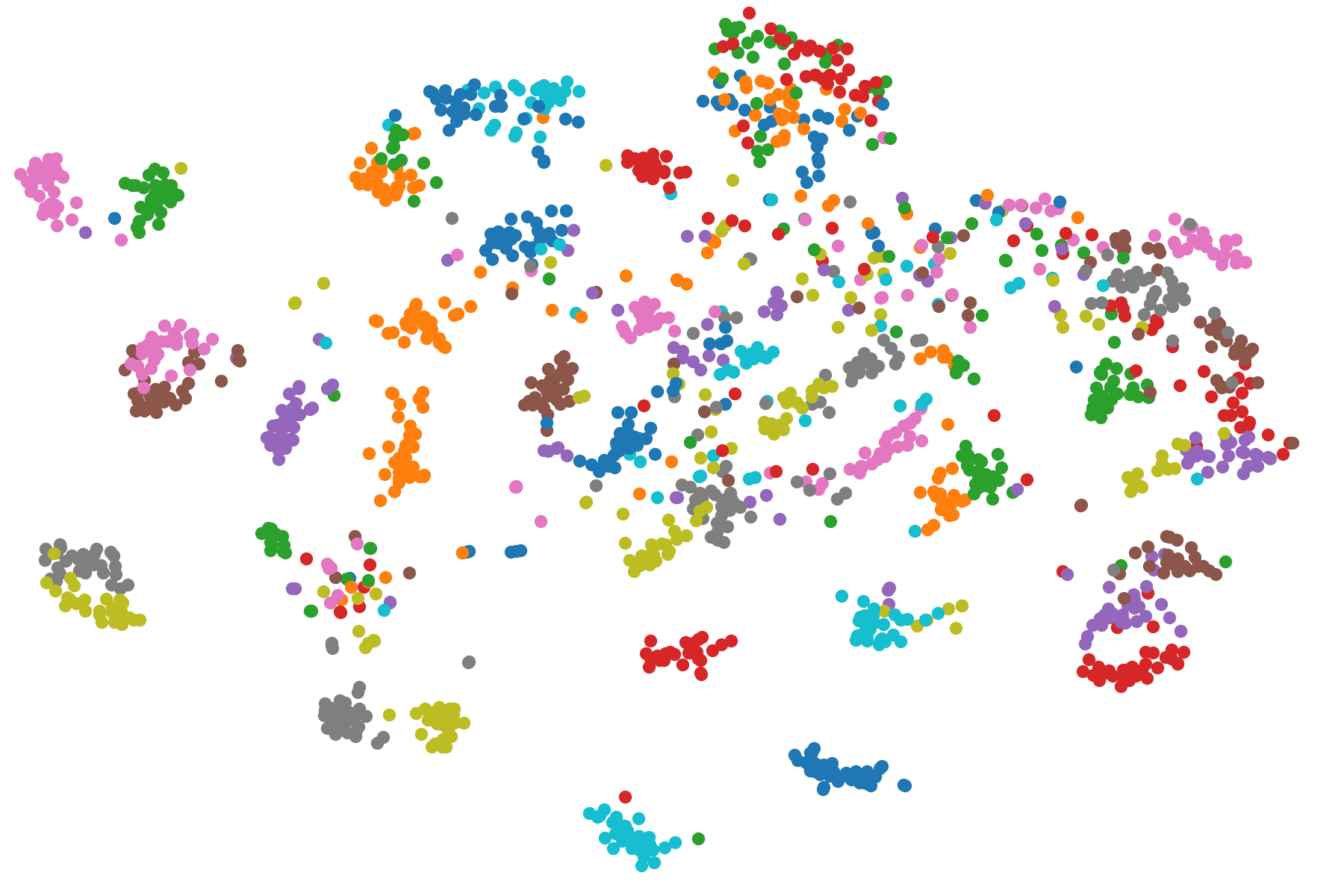}}\hfill
    \subfloat[Context Refinement ($\mathbf{v}$)]{\includegraphics[width=0.22\textwidth] {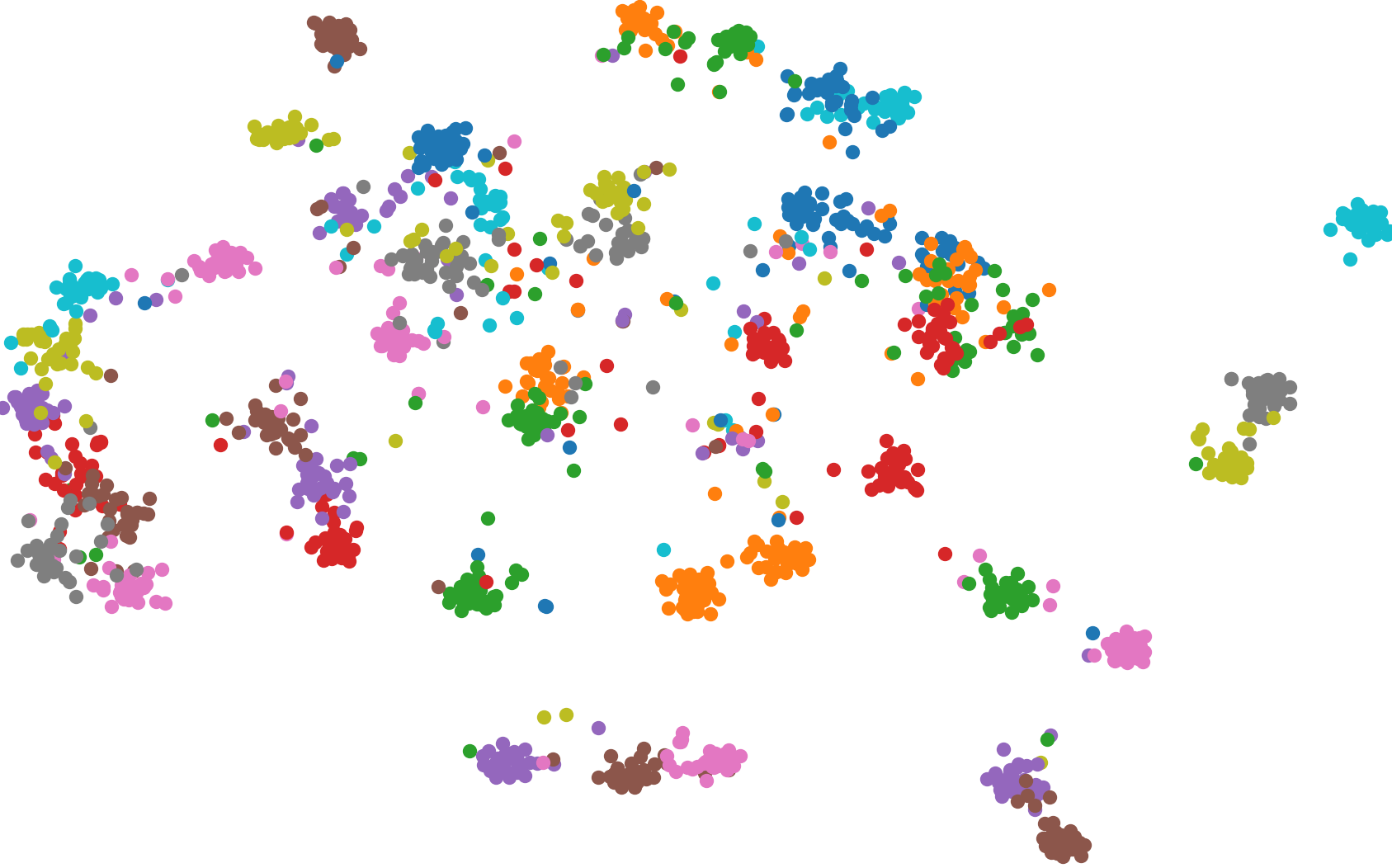}}\hfill \\
    \subfloat[Final feature  $\bar{f}_t = f_t + f_t \otimes \mathbf{v}$]{\includegraphics[width=0.23\textwidth] {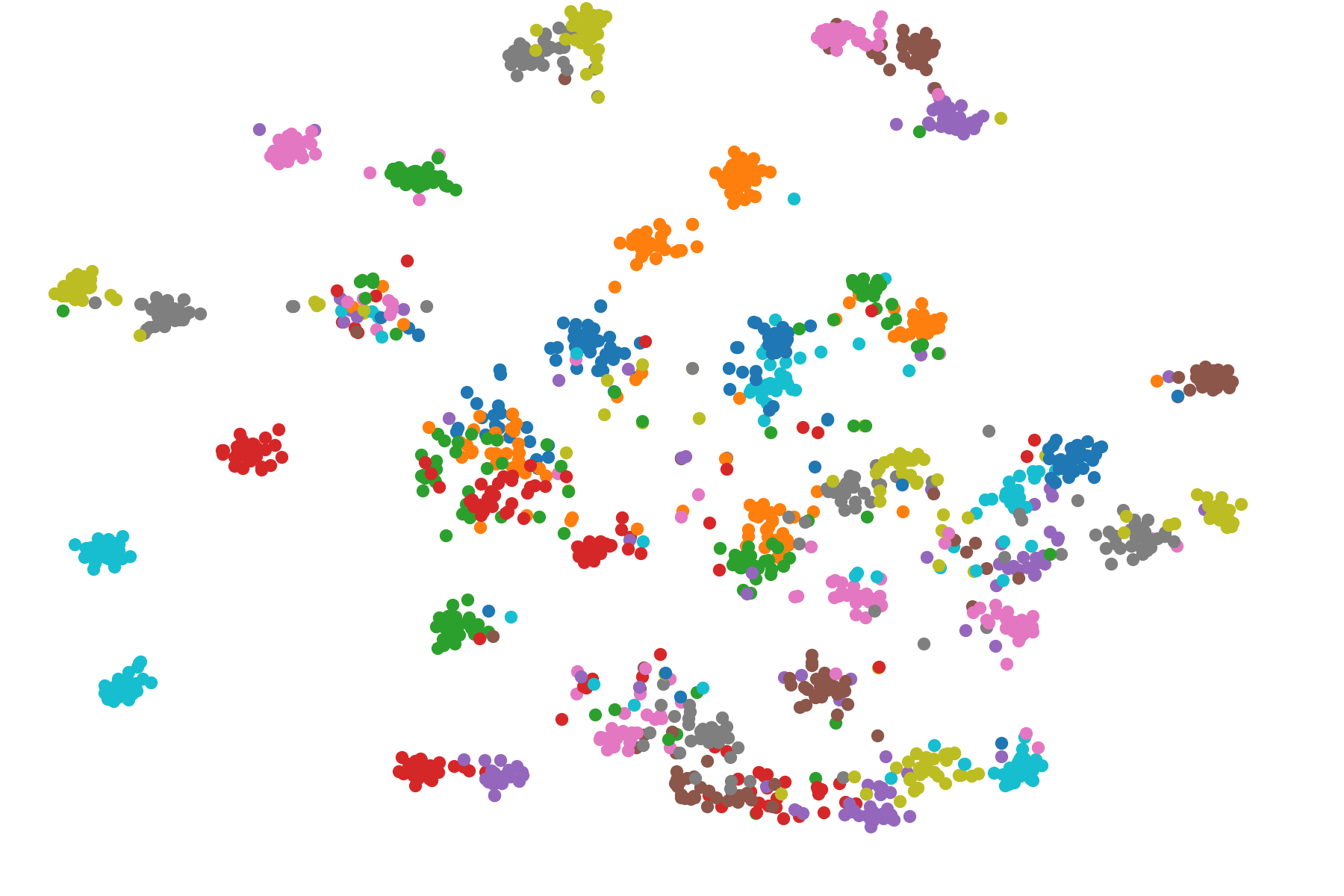}} \hfill
    \subfloat[Final feature $\bar{f}_t = f_t$ \textbf{without} the context refinement (weight $\mathbf{v}$) module]{\includegraphics[width=0.25\textwidth] {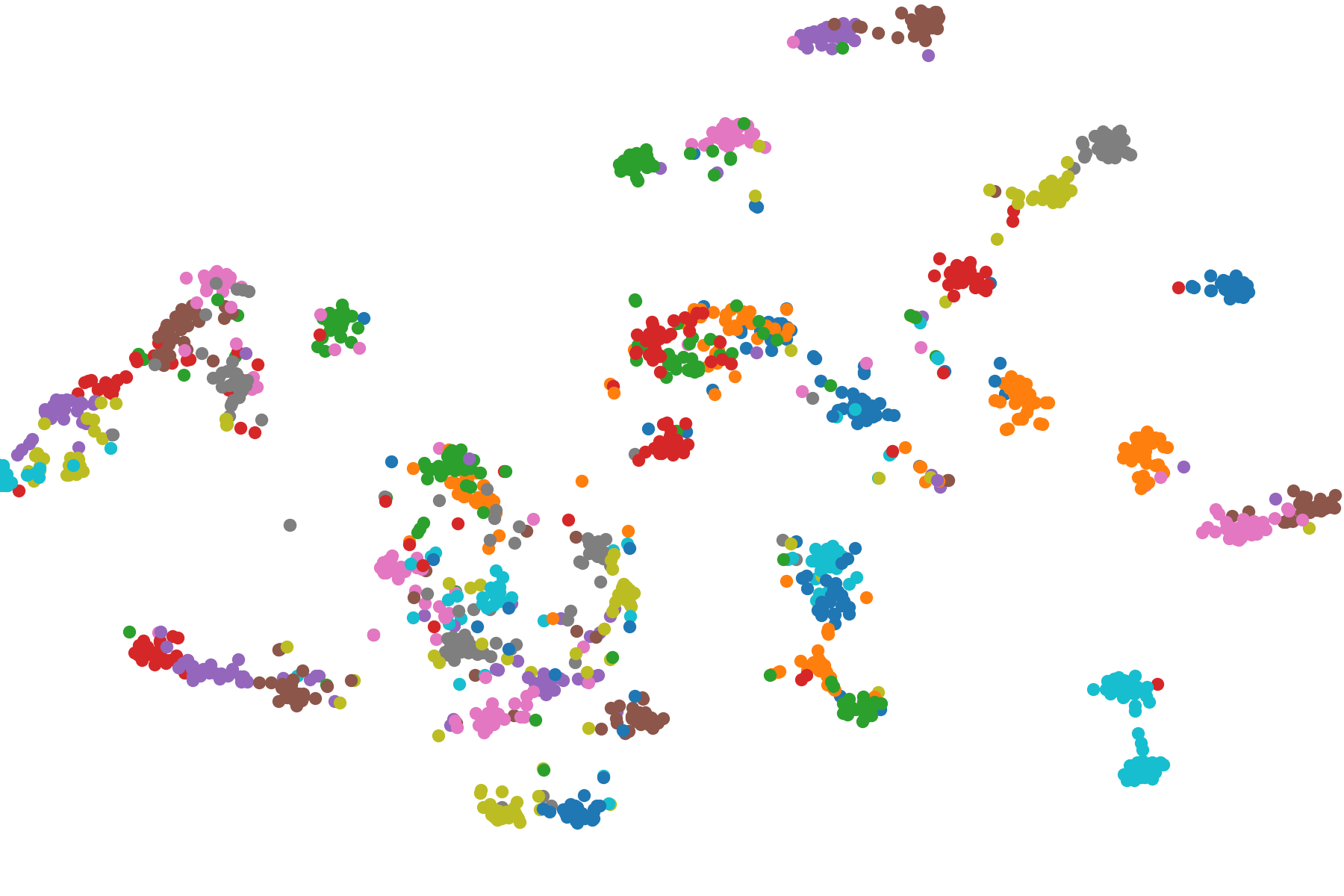}} \hfill
     \subfloat[Standalone Base CNN]{\includegraphics[width=0.23\textwidth] {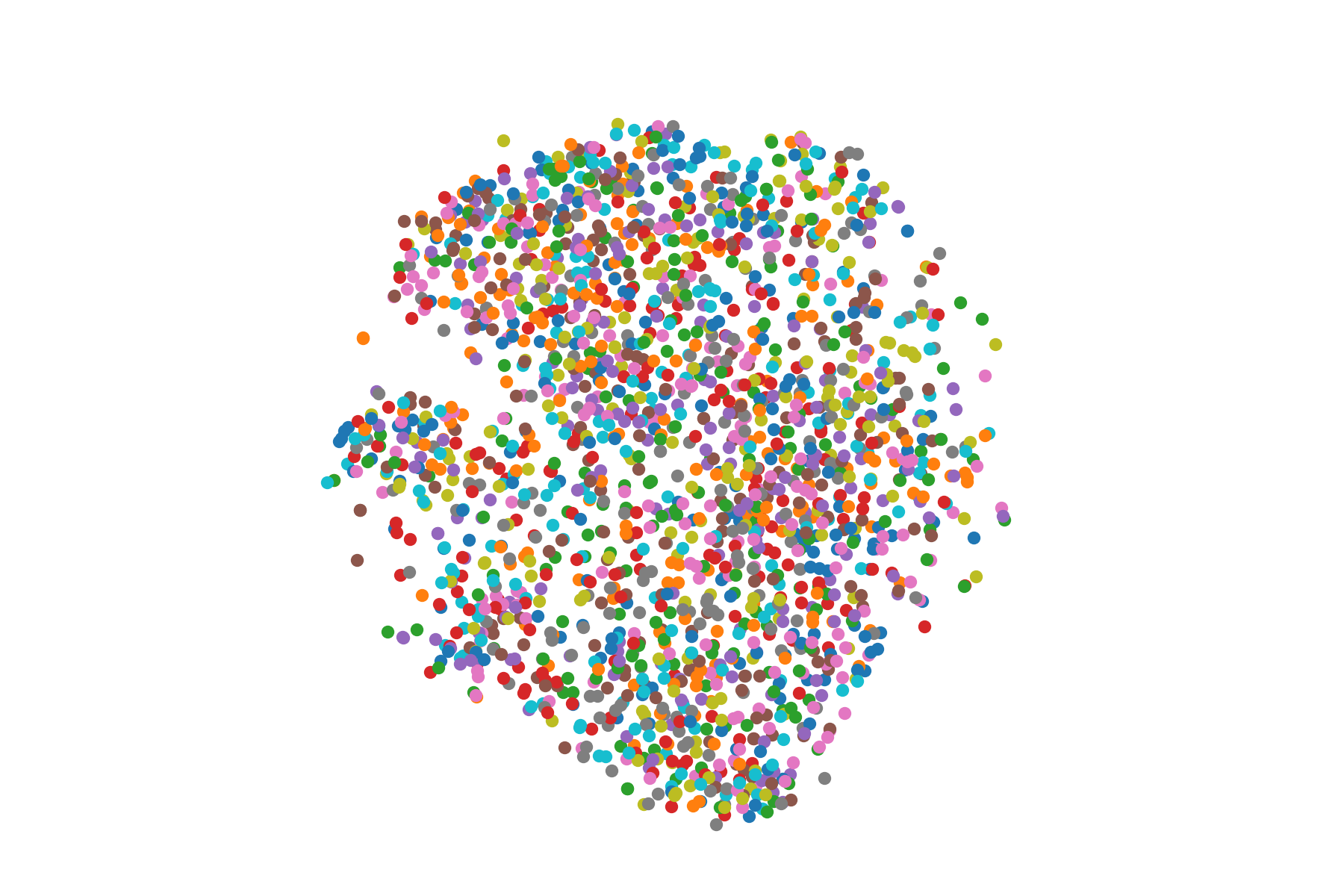}}\hspace{1cm}
 
 \caption{\texttt{SR-GNN}'s discriminability of the Aircraft test-set using t-SNE \cite{van2014accelerating} to visualize class separability and compactness using features from a) base CNN (Xception, Fig. \ref{fig:full_model}(a)) within our model, b) relation-aware transformed feature using GNN (Fig. \ref{fig:full_model}(b)), c) attentional context refinement weight-vector $\mathbf{v}$ (Fig. \ref{fig:full_model}(c)), and d) the final image-level feature map $\bar{f}_t$ for classification (Fig. \ref{fig:full_model}(c)). Each color represents a particular class. There are 50 classes chosen randomly from the \textbf{Aircraft's} test set. e) \texttt{SR-GNN} \textbf{without} the context refinement module, and f) Standalone Xception base CNN without our modules (re-trained on the Aircraft dataset).} 
    \label{fig:TSNE_Aircraft}
\end{figure*}
\vspace{ -0.2 cm}
\begin{table*}
\begin{center}
\vspace{ -0.2 cm}
\caption{Davies-Bouldin \cite{davies1979cluster} Index (Lower is Better) to Quantify Cluster Similarities Using the t-SNE \cite{van2014accelerating} Outputs over All Test Images from Aircraft 
Dataset. For the Training and Validation of the Backbone CNN (Xception), We Use the Standard Transfer Learning by Fine-Tuning it on the Target Dataset Using the Same Data Augmentation and Hyper-Parameters (Sec. \ref{experiments}). The Clusters Generated by \texttt{SR-GNN} are More Compact and Separated Than the Baseline Xception (Last Row). The Final Feature Description of \texttt{SR-GNN} is Better Than the Individual Relation-Aware Feature Transform  and Context Refinement Modules. }
\begin{tabular}{|l| c c c|}
 \hline
Feature Extraction Point & Aircraft & Cars & Flowers\\
    \toprule
    \multicolumn{4}{|c|}{\texttt{SR-GNN}'s different extraction points}\\
    \midrule
   Base CNN (Fig. \ref{fig:full_model}(a)) &  4.00 (Fig. \ref{fig:TSNE_Aircraft}(a))  & 9.89 (Fig. 
   8(a)) & 1.52 (Fig. 
   9(a))\\
   Transformed feature $f_t$ (Fig. \ref{fig:full_model}(b)) &  3.34 (Fig. \ref{fig:TSNE_Aircraft}(b))  & 2.49 (Fig. 8(b))& 1.35 (Fig. 9(b))\\
   Attentional Context Refinement Weight $\mathbf{v}$ (Fig. \ref{fig:full_model}(c)) &  2.65  (Fig. \ref{fig:TSNE_Aircraft}(c)) & 2.25 (Fig. 8(c))  & 1.12 (Fig. 9(c))\\
   Final feature  $\bar{f}_t = f_t + f_t \otimes \mathbf{v}$ (Fig. \ref{fig:full_model}(c)) &  \textbf{2.07} (Fig. \ref{fig:TSNE_Aircraft}(d)) & \textbf{2.25} (Fig. 8(d)) & \textbf{1.02} (Fig. 9(d))\\
   \midrule
   \multicolumn{4}{|c|}{\texttt{SR-GNN}'s  \textbf{without} the context refinement (weight $\mathbf{v}$) module}\\
    \midrule
   Final feature w/o refinement $\bar{f}_t = f_t$ &  3.42 (Fig. \ref{fig:TSNE_Aircraft}(e)) & 3.49 (Fig. 8(e)) & 1.19 (Fig. 9(e))\\
   \midrule
   \multicolumn{4}{|c|}{Base CNN (Xception) trained on target dataset (Transfer Learning)}\\
    \midrule
   Xception (baseline)  & 77.22 (Fig. \ref{fig:TSNE_Aircraft}(f)) & 95.85 (Fig. 8(f)) & 38.67 (Fig. 9(f))  \\
   \bottomrule
\end{tabular}
 \label{table:davies_bouldin_score}
 \end{center}
 \vspace{-0.5 cm}
\end{table*}
\begin{figure*} [!t]
\begin{center}
\includegraphics[width=0.15\linewidth, height=0.075\linewidth]{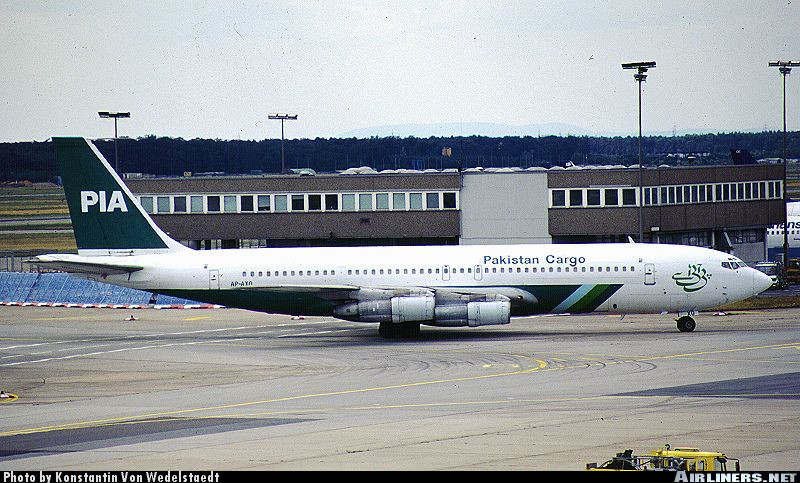}
\includegraphics[width=0.15\linewidth, height=0.075\linewidth]{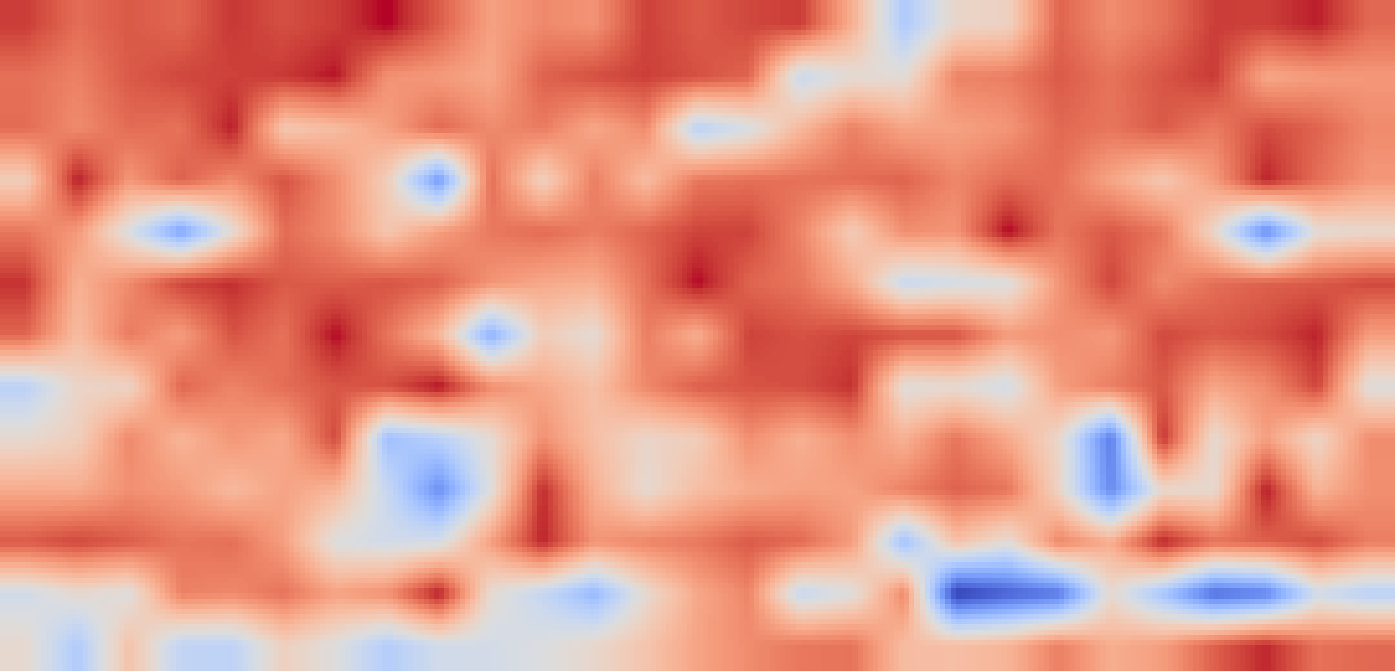}\hfill
\includegraphics[width=0.15\linewidth, height=0.075\linewidth]{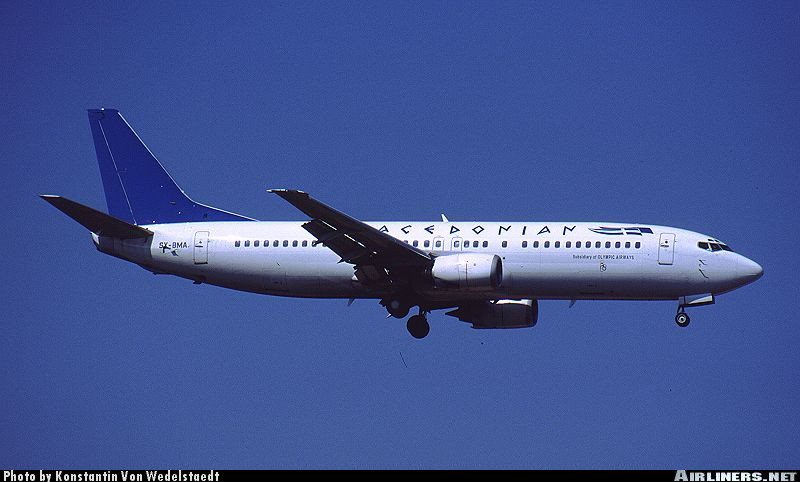}
\includegraphics[width=0.15\linewidth, height=0.075\linewidth]{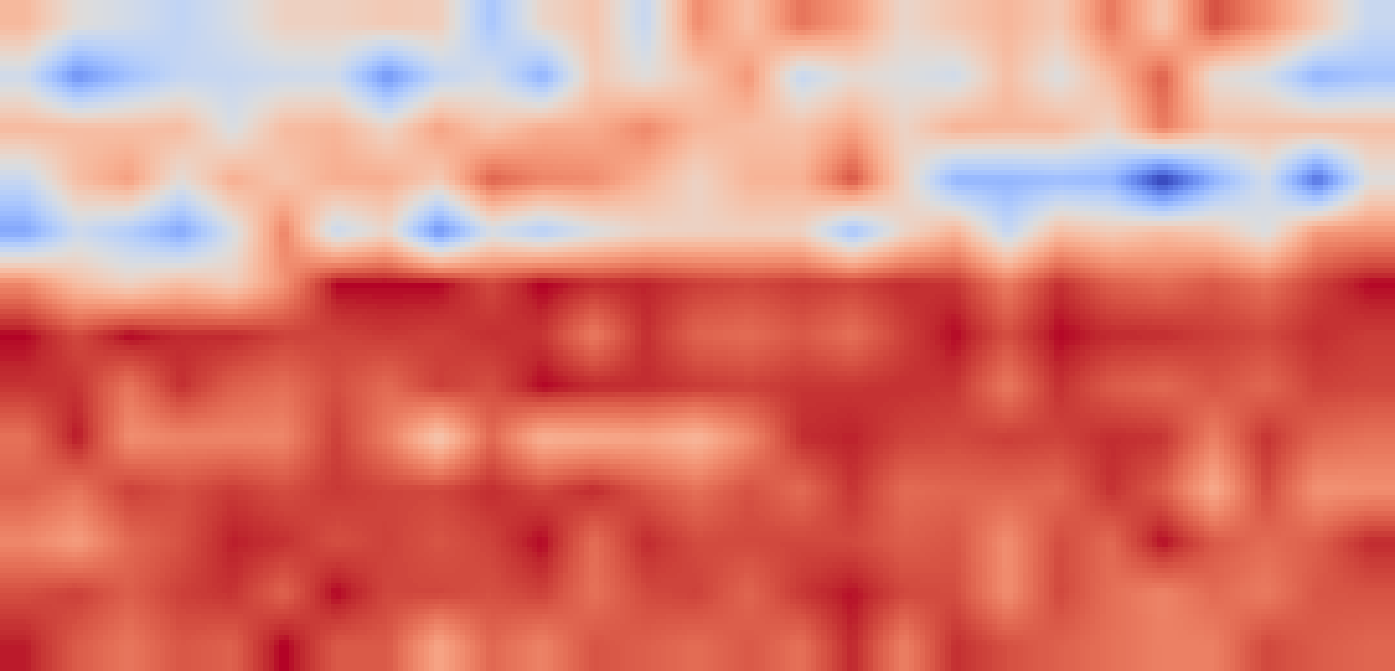}\hfill
\includegraphics[width=0.15\linewidth, height=0.075\linewidth]{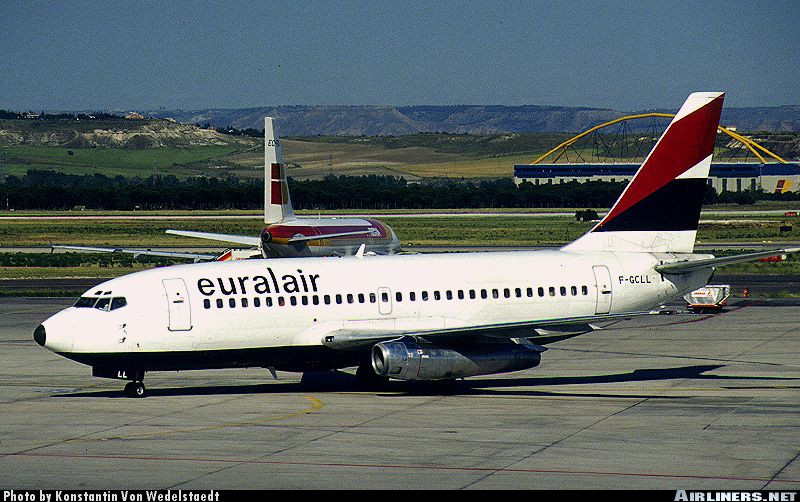}
\includegraphics[width=0.15\linewidth, height=0.075\linewidth]{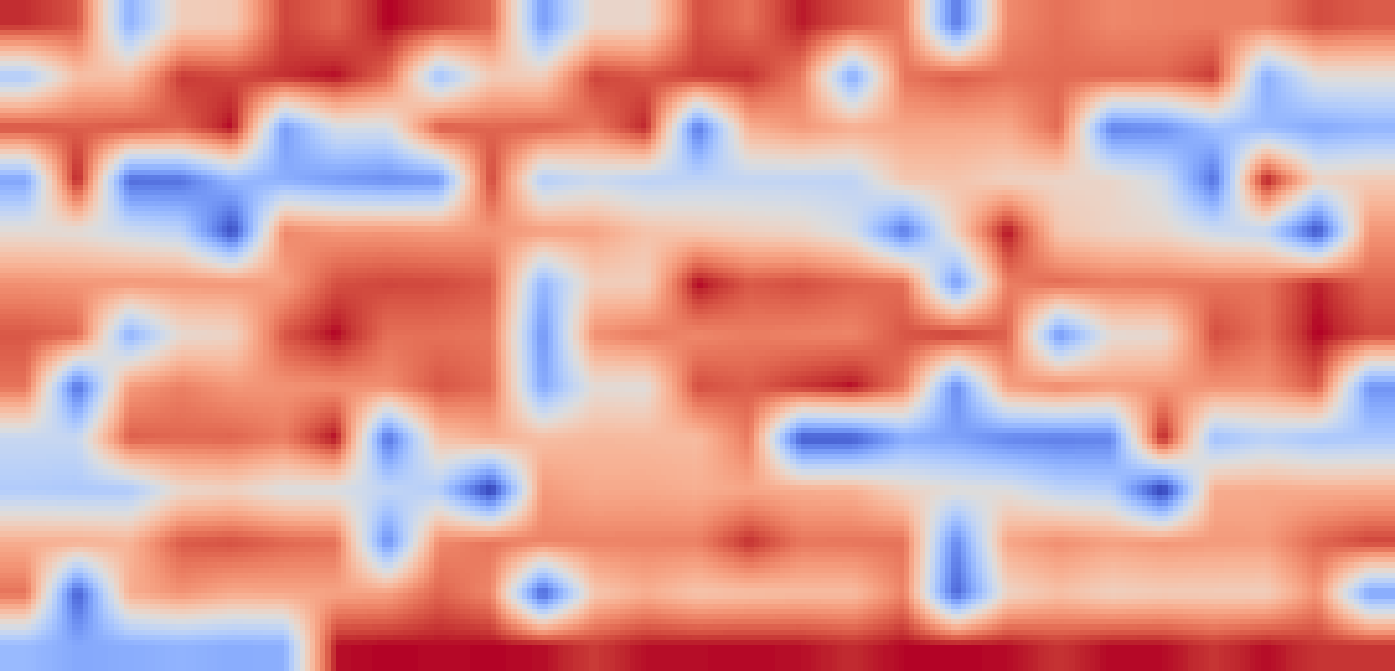} \\ \vspace{1mm}
\includegraphics[width=0.15\linewidth, height=0.075\linewidth]{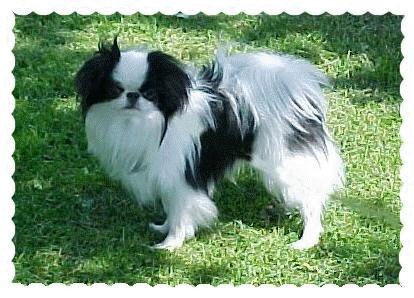}
\includegraphics[width=0.15\linewidth, height=0.075\linewidth]{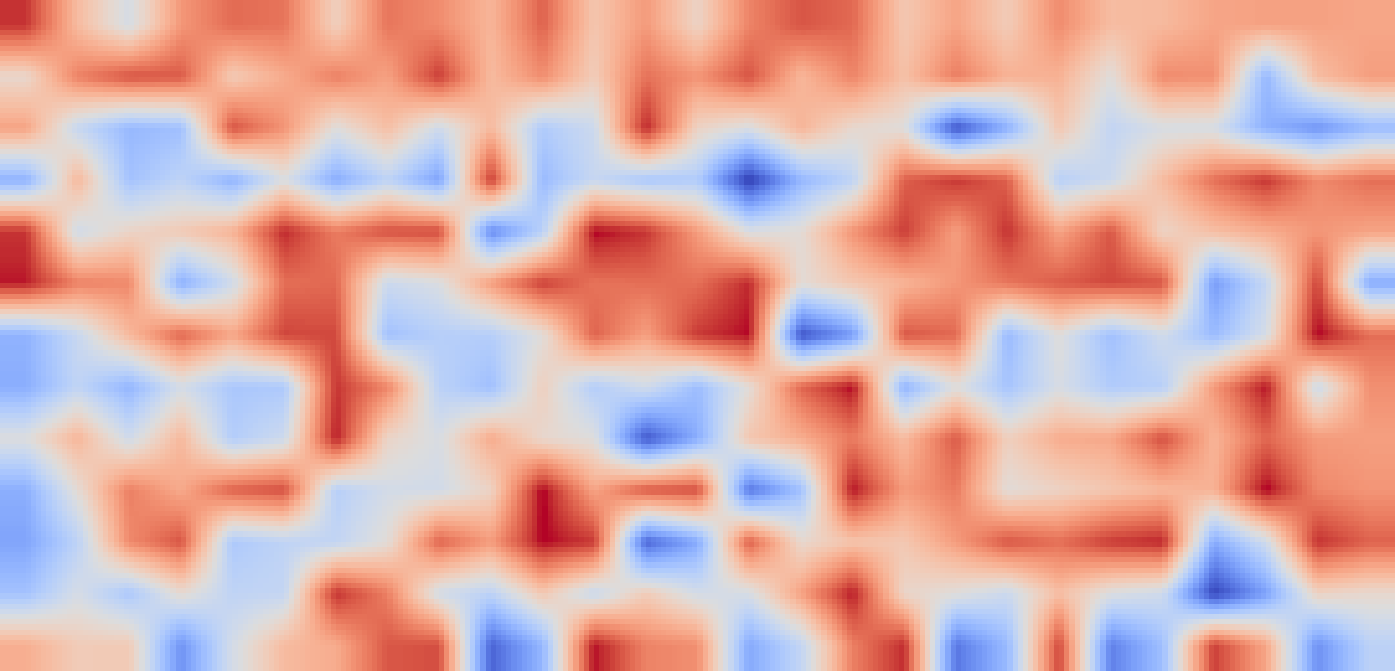}\hfill
\includegraphics[width=0.15\linewidth, height=0.075\linewidth]{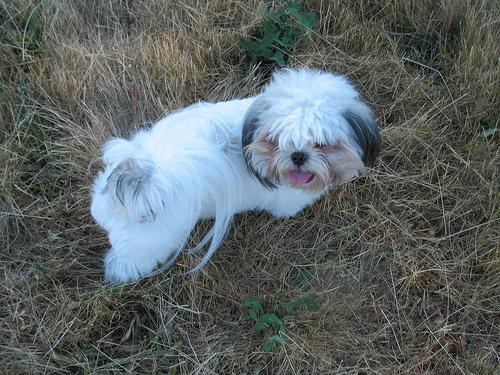}
\includegraphics[width=0.15\linewidth, height=0.075\linewidth]{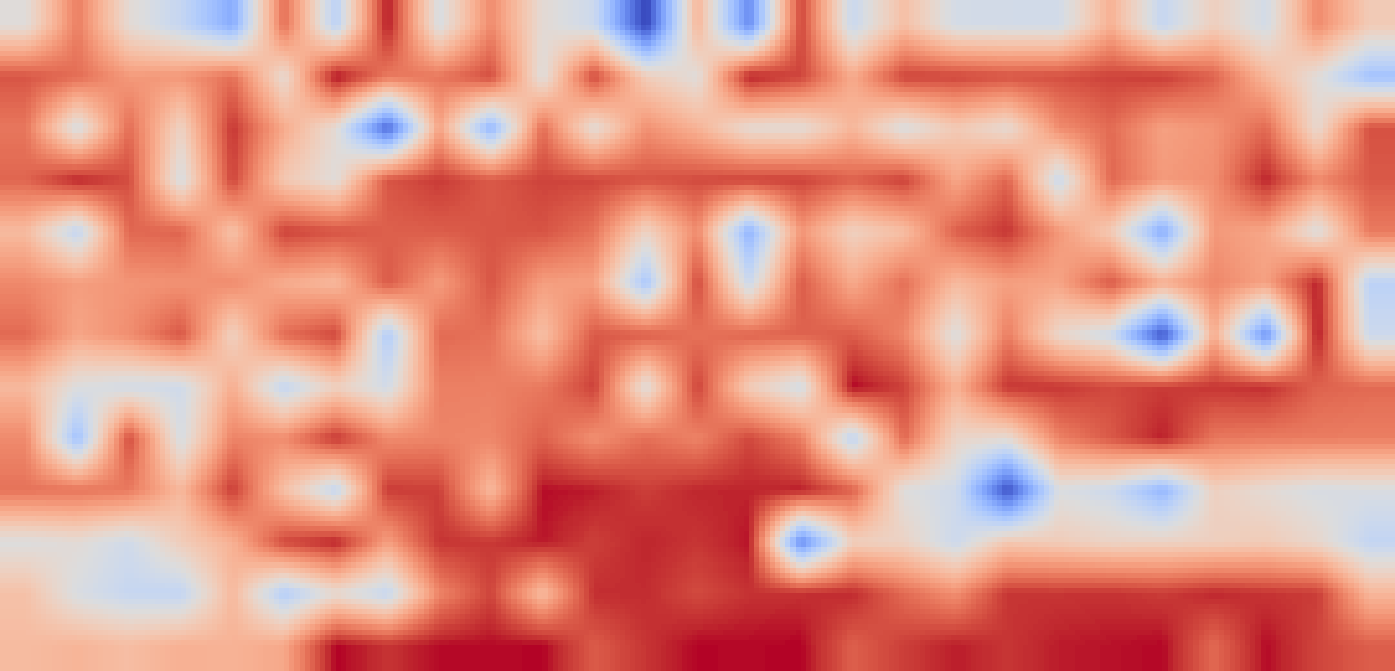}\hfill
\includegraphics[width=0.15\linewidth, height=0.075\linewidth]{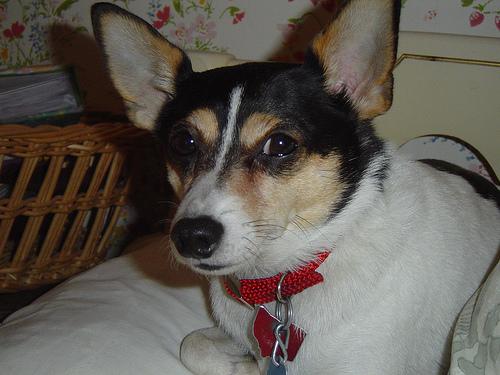}
\includegraphics[width=0.15\linewidth, height=0.075\linewidth]{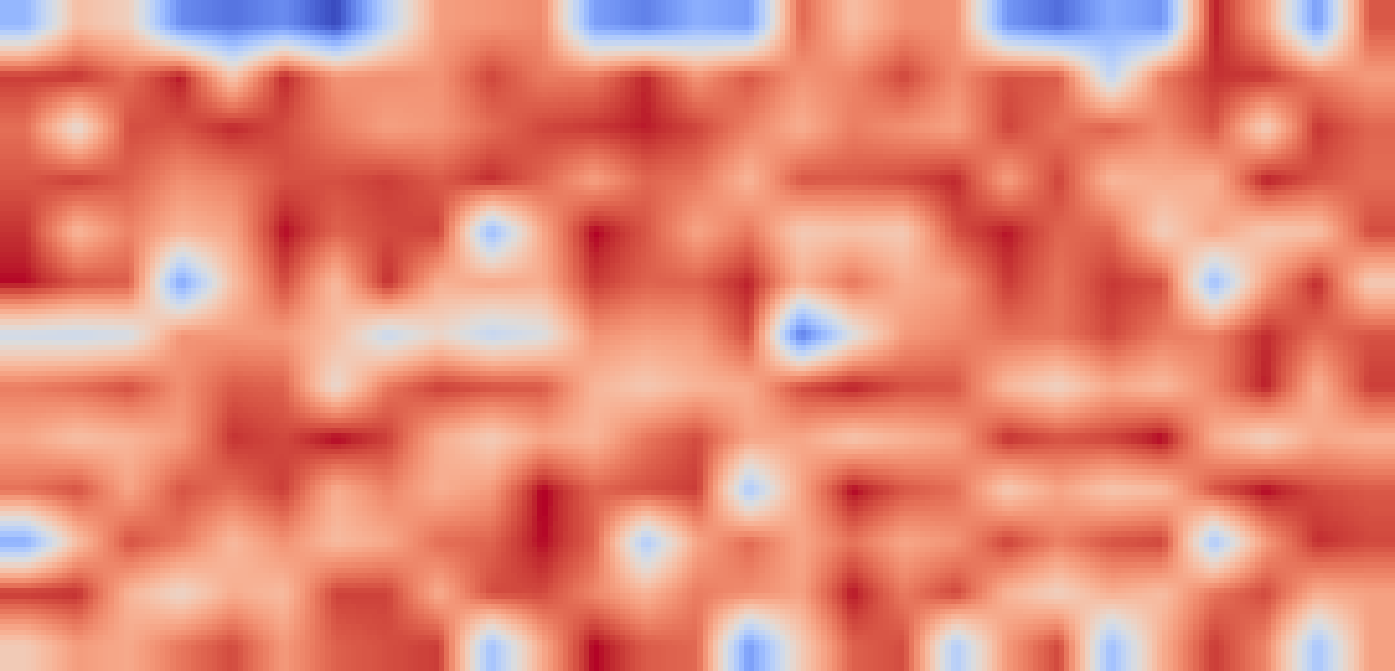}

\end{center}
\vspace{-0.3cm}
   \caption{ Visualization of the relation-aware transformation using \textbf{cosine similarity} to measure pairwise relationships ({\color{cyan}cool} to {\color{red}warm} $\Rightarrow$ weak to strong) between nodes in the graph. Top (Aircraft): 707-320, 737-400 and 737-200 (left to right). Bottom (Dogs): Japanese Spaniel, Shih Tzu and Toy Terrier.
    }
\label{fig:fig_tsne}
\vspace{-1.5em}
\end{figure*}
\begin{figure*}[t]
\centering
    \subfloat[Joint attentions map]{\includegraphics[width=0.15\textwidth] {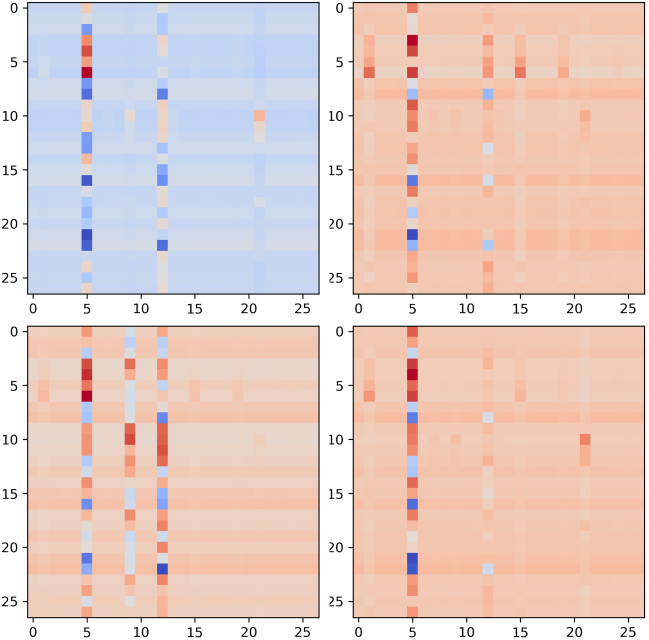}}\hfill
     \subfloat[737-600: top-2 $\mathcal{R}$ (5 \& 12) and their top-3 joint attentions]{\includegraphics[width=0.42\textwidth] {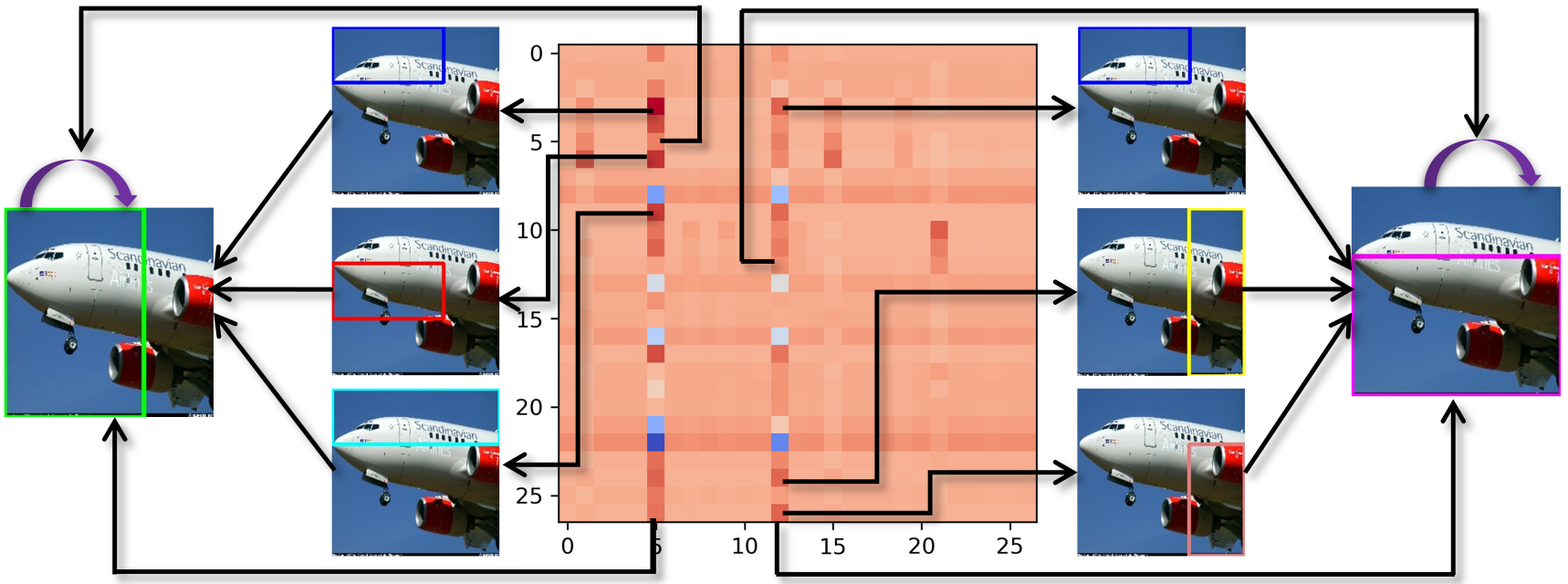}}\hfill
    \subfloat[F-16A\_B: top-2 $\mathcal{R}$ (15 \& 19) and their top-3 joint attentions]{\includegraphics[width=0.42\textwidth] {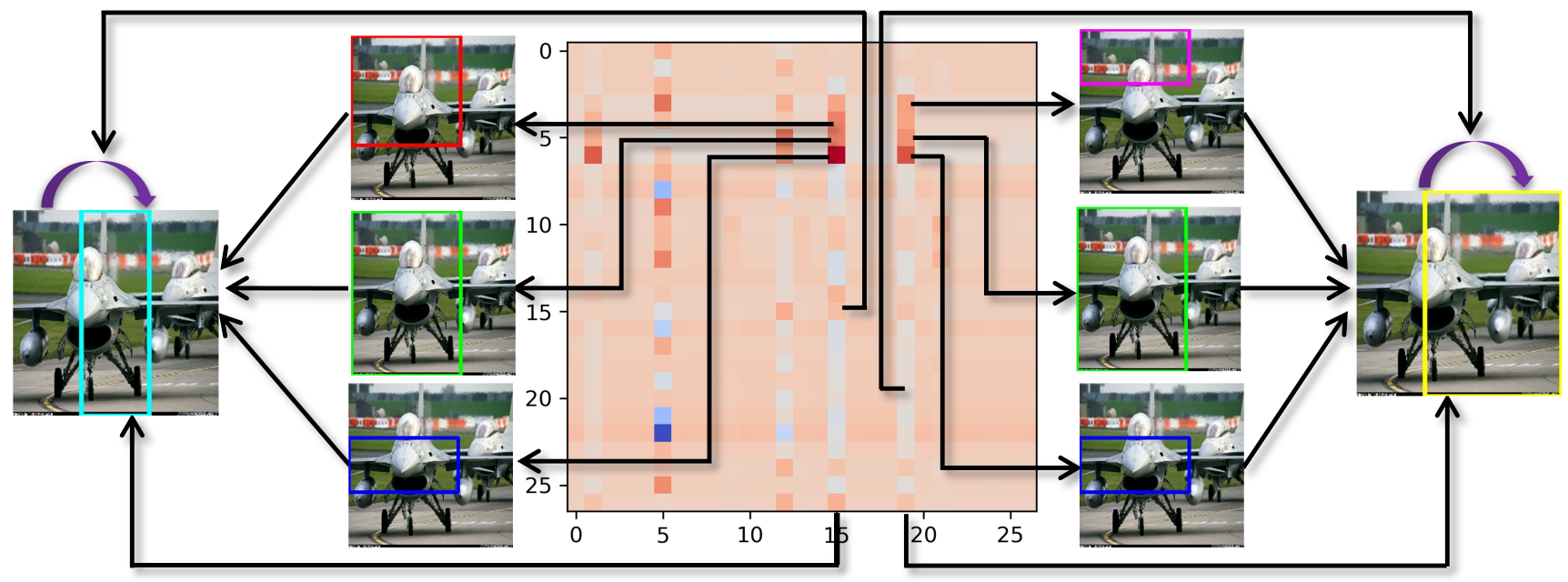}}
    \caption{Visualization of attentional context refinement (Fig. \ref{fig:full_model}(c)) in our \texttt{SR-GNN} over the sub-types in the Aircraft dataset: a) joint attentional maps for `A340-200', `ATR-72', `DC-10' and `ERJ 135' aircraft sub-types. b) Top-2 regions (cols 5 \& 12) contributing towards sub-type `Boeing 737-600' conditioned on the respective other top-3 regions (rows) in joint decision-making. The self-attention (self-loop) is also shown in the top-2 regions. c) Similarly, top-2 regions (cols 15 \& 19) contributing towards sub-type `F-16A\_B' conditioned on the respective other top-3 regions (rows). Regions are shown in the respective original images.}
    \label{fig:fig_attn}
    \vspace{-0.2cm}
\end{figure*}

\vspace{ 0.3 cm}
\subsection {Model Complexity} 
\texttt{SR-GNN}'s capacity and computational complexity is assessed using  GFLOPs (Giga floating point operations), model size as a number of trainable parameters in millions (M) and per-image inference time in milliseconds (ms) \cite{tan2019efficientnet}. These values over four datasets is given in Table \ref{table:time}. Its comparison to the SotA methods is also provided in Table \ref{table:Complx2}. For $\mathcal{R}$=27, its complexity in terms of trainable parameters (base CNN: 20.9M, W/o Refine: 24.4M and full model: 30.9M), per-image inference time (base CNN: 2.7ms, W/o Refine: 4.9ms and \texttt{SR-GNN}: 5.0ms) and GFLOPs (base CNN: 9.2B, W/o Refine: 9.3B and full model: 9.8B) are given in Table \ref{table:time}. The key modules only add a little overhead to the base CNN in terms of trainable parameters and GFLOPs: 1) relation-aware feature transformation (Fig. \ref{fig:full_model}(b)): 3.4M and 0.18B; and 2) attentional context modeling (Fig. \ref{fig:full_model}(c)):  6.4M and 0.51B. GFLOPs and model’s trainable parameters are widely used by the community to compare the computational efficiency of various deep models \cite{tan2019efficientnet}. By considering this, our \texttt{SR-GNN} (param: 30.9M, GFLOPs: 9.8B) is computationally a lighter model than CAP (param: 34.2M, GFLOPs: 10.2B) using Xception that is more lightweight than the other SotA models in Table \ref{table:Complx2}. Likewise, using ResNet-50 as a base CNN,  \texttt{SR-GNN} (param: 33.4M, GFLOPs: 8.4B) is lighter than RAN (param: 49.0M, GFLOPs: 8.5B) and AG-Net (param: 54.8M, GFLOPs: 10.4B, and per-image inference time: 5.2 ms). 

 The inference time is dependent on types of GPU, hardware and software environments used. For example, both our \texttt{SR-GNN} and CAP \cite{behera2021context} use the same Titan V GPU (12 GB) to run the model, but CAP is implemented using TensorFlow 1.x whereas, \texttt{SR-GNN} runs using TensorFlow 2.x. It is well-known that the TensorFlow 2.x is significantly slower\footnote{https://github.com/tensorflow/tensorflow/issues/33487} than TensorFlow 1.x, resulting in increase of the inference time for \texttt{SR-GNN} (5.0ms) in comparison to CAP (4.2ms) even though the former is more lightweight (param: 30.9M, GFLOPs: 9.8B) than the latter (param: 34.2M, GFLOPs: 10.2B). Moreover, the per-image inference time of our \texttt{SR-GNN} is 0.8ms and 0.1ms higher than CAP and MRDMN-L \cite{xu2021multiresolution}, respectively. \texttt{SR-GNN} without refinement (4.9ms) shares the same inference time with MRDMN-L, but gains 7.5\%  higher accuracy on Dogs. A precise comparison with the existing top-10 SotA methods focusing on the inference time is given in the supplementary document, irrespective of the GPU and hardware configuration, deep learning tools (\textit{e.g.}, TensorFlow, PyTorch, MXNet, etc.) and related experimental constraints used in those works. Moreover, the Transformers are computationally more complex than \texttt{SR-GNN} as shown in Table \ref{table:ViT}.

Some works \cite{chang2020devil, ge2019weakly, ge2017borrowing} improve the accuracy by exploring secondary data. Also, such methods involve multiple steps and are resource-intensive. For example, there are three steps in \cite{ge2019weakly}: 1) object detection and segmentation using Mask R-CNN and a conditional random field (CRF); 2) complementary part mining using 512 regions; and 3) classification using context gating. Their model is trained using 4 GPUs (12GB each), per-image inference time is 27ms for Step 3 and extra 227ms in Step 2. Our model is trained on a single GPU (12GB) with per-image inference time of 5ms only. So, \texttt{SR-GNN} is faster and lighter than most of the existing methods.  

\subsection {Qualitative Analysis and Visualization} To get insight into our model's decision-making process, we visualize the feature maps at key steps. Each step provides the discriminability of our model by visualizing the class separability and compactness. To achieve this, we use t-SNE \cite{van2014accelerating}, which is shown in Fig. \ref{fig:TSNE_Aircraft}. Randomly selected 50 classes are chosen from the Aircraft test-set. The test images are processed to extract features from base CNN (Fig. \ref{fig:full_model}(a)), relation-aware transformed feature (Fig. \ref{fig:full_model}(b)), context refinement weight vector (Fig. \ref{fig:full_model}(c)) and the final refined feature descriptor. The respective visualization (unique color per class) is presented in Fig. \ref{fig:TSNE_Aircraft}. It is evident that clusters representing both relation-aware features (Fig. \ref{fig:TSNE_Aircraft}(b)) and context weight vector (Fig. \ref{fig:TSNE_Aircraft}(c)) are further apart and more compact compared to the base CNN features (Fig. \ref{fig:TSNE_Aircraft}(a)). Moreover, the clusters representing the final refined feature map (Fig. \ref{fig:TSNE_Aircraft}(d)) is further enhanced, resulting in a clearer distinction between various clusters representing different classes. In addition, the importance of the context refinement task is visualized by avoiding it from the final feature vector (Fig. \ref{fig:TSNE_Aircraft}(e)). Lastly, standalone Xception is fine-tuned by discarding our proposed modules, and its impact is shown in Fig. \ref{fig:TSNE_Aircraft}(f). Overall, these qualitative visualizations evince the essence of key components and superior performance of  \texttt{SR-GNN}.   

We have further computed the Davies-Bouldin index \cite{davies1979cluster} to quantitatively evaluate the cluster similarities using the t-SNE outputs given in Table \ref{table:davies_bouldin_score}. This index signifies the similarity between clusters, where the similarity is the ratio of within-cluster distances to between-cluster distances. As a result, a lower value implies better clustering. For the Aircraft test-set, these values are 4.00 (Fig. \ref{fig:TSNE_Aircraft}(a)), 3.34 (Fig. \ref{fig:TSNE_Aircraft}(b)), 2.65 (Fig. \ref{fig:TSNE_Aircraft}(c)), and 2.07 (Fig. \ref{fig:TSNE_Aircraft}(d)). Whereas, the value increases without feature refinement 3.42 (Fig. \ref{fig:TSNE_Aircraft}(e)), and it is very high (77.22) using Xception backbone only (Fig. \ref{fig:TSNE_Aircraft}(f)). The results on Cars and Flowers are given in the supplementary document.   

In order to understand how the relation-aware transformation is exploited (Fig. \ref{fig:full_model}(b)) in our \texttt{SR-GNN}, we also visualize the cosine similarity to measure the pairwise relationships between each pair of nodes (regions) in the graph as is shown in Fig. \ref{fig:fig_tsne}. It is not easy to visualize the graph structure associating object parts with categories. Thus, pairwise cosine similarities of nodes representing regions are explored to reflect how each object is overall represented and distinguished from each other. It is evident that the graph indeed captures the relational structure to discriminate subordinate categories. The main reason is that $\alpha$ in (\ref{eq_2}) preserves locality to avoid over-smoothing by staying close to the root node and leveraging the information from a large neighborhood.

We have also looked inside our attentional context refinement module (Fig. \ref{fig:full_model}(c)) to visualize the class-specific visual relationships jointly learned during the training. These relationships are learned as an $\mathcal{R}$$\times$$\mathcal{R}$ joint attention map and are shown in Fig. \ref{fig:fig_attn}(a) for the `A340-200', `ATR-72', `DC-10', `ERJ 135' aircraft  sub-types where $\mathcal{R}$$=$$27$. Each column represents a region conditioned on itself and other regions linking rows. \textcolor{blue}{Blue} to \textcolor{red}{red} signifies the class-specific \textit{less} to \textit{more} attention towards that region. We further link the top-2 regions (cols) to their respective top-3 joint visual attentions (rows) by exploring the attention map. These are shown in Fig. \ref{fig:fig_attn}(b) and \ref{fig:fig_attn}(c) for the respective `Boeing 737-600' and `F-16A\_B' Aircraft sub-type. These regions are drawn in the original image to show their joint relationships. From both figures, it is evident that our model learns to focus on the key context information for discriminating subtle variations. More results for visualization are given in the supplementary.  

\begin{table} 
\begin{center}
 \caption{Accuracy (\%) of Our \texttt{SR-GNN} with Varied Key Modules. 1) Uniform patch-size as an alternative to  Generate Regions 2) Performance of Various Graph Pooling Techniques in Our Feature Transform Module (Fig. \ref{fig:full_model}(b)), and 3) Effectiveness of Major Components of \texttt{SR-GNN}. }
 \label{table:Abln1}
\begin{tabular}{|l| c c c c |}
 \hline
   Key Modules  & Aircraft & Cars &  Dogs & Flowers   \\
   \hline  
 Uniform $4\times4$ grid & 94.3 & 92.9 & 96.3 & 97.0  \\
 \hline
Global average pooling & 95.0 & 95.1 & 96.2 &97.5 \\
Global max pooling &94.9 &95.5 &96.4 & 97.8 \\
Global sum pooling \cite{xu2018powerful} & 94.9 & 95.1 & 96.3 & 97.7  \\
Sort pooling \cite{zhang2018end} & 95.2 & 95.3 & 96.5 & 97.8  \\
 \hline
 W/o  GNN  & 93.5 & 94.5 & 96.0 & 94.9   \\
 W/o Refine & 93.5 & 93.7& 96.5 & 97.1\\
 W/o self-attention & 93.7 & 95.1 & 96.1 & 96.4  \\
  W/o weighted-attention & 94.4 & 95.5 & 96.0 & 95.1   \\
\hline
\textbf{\texttt{SR-GNN}} (full-model) & \textbf{95.4}  &\textbf{96.1} &\textbf{97.3} & \textbf{97.9} \\
\hline
\end{tabular}
 \end{center}
 \vspace{-0.3cm}
\end{table}
%
\begin{table}
\begin{center}
 \caption{Accuracy (\%) of \texttt{SR-GNN} with Different Numbers of Region Proposals (Fig. \ref{fig:full_model}(a)). }
 \label{table:RoI}

\begin{tabular}{|l| c c c c|}
 \hline
    Dataset & \#11 & \#19 & \#27 & \#36 \\
    \hline 
  Aircraft & 90.6 & 90.3 & \textbf{95.4} & 89.4  \\
  CUB &  86.9 & 85.8 & \textbf{91.9} & 82.8\\ 
Cars & 93.4 & 90.9 & \textbf{96.1} & 92.5  \\
  Dogs &  94.5 & 94.3 & \textbf{97.3} & 95.5\\ 
  Flowers & 95.2 & 97.5 & \textbf{97.9} & 95.8\\
\hline
\end{tabular}
 \end{center}
 \vspace{-0.5 cm}
\end{table}

\begin{table*} 
\begin{center}
 \caption{Accuracy (\%) of \texttt{SR-GNN} with 512 and 1024 Output Dimensions at Different Teleport (or Restart) Probability $\alpha \in [0.1, 0.8]$ in (\ref{eq_2}).  }
 \label{table:APPNP_alpha2}
 \begin{small}
\begin{tabular}{|l|p{4.0 mm} p{4.5 mm} p{4.5 mm} p{4.5 mm} p{4.5 mm} p{4.5 mm} p{5 mm} p{5 mm}| p{4.0 mm} p{4.5 mm} p{4.5 mm} p{4.5 mm} p{4.5 mm} p{4.5 mm} p{5 mm} p{5 mm}| }
 \hline
 \multicolumn{1}{|c|}{Dataset} &
 \multicolumn{8}{|c|}{GNN output dimension = 512}  & \multicolumn{8}{|c|}{GNN output dimension = 1024} \\
 & 0.1 & 0.2 & 0.3 & 0.4 & 0.5 & 0.6 & 0.7 & 0.8 & 0.1 & 0.2 & 0.3 & 0.4 & 0.5 & 0.6 & 0.7 & 0.8     \\ 
    \hline
  Aircraft &  91.6 &94.6 & 94.8 &94.7 &94.7 &92.1 &91.5 & 92.3 
  & 90.7 & 92.4 & \textbf{95.4} & 91.1 & 91.7 &92.1 &92.3 &90.9 \\
   Cars & 95.5 & 95.5 & 95.6 &94.9 & 93.9 & 93.9 &93.5 & 93.3
   & 95.4 & 95.7 & \textbf{96.1} & 95.8 & 96.0 &95.8 & 95.8 & 95.8\\
  Dogs & 96.7 & 96.7 & 96.9 &96.5 & 96.4 & 96.5 &96.5 & 96.1
    & 96.8 & 97.0 & \textbf{97.3} & 97.1 & 96.7 &96.7 &96.7 & 96.6\\ 
  Flowers & 97.6 & 97.8 & 97.9 & 97.8 & 97.6 &97.6 &97.7 & 97.7
  & 97.5 & 97.5 & \textbf{97.9} & 97.8 & 97.7 & 97.7 & 97.7 & 96.6\\
\hline
\end{tabular}
\end{small}
 \end{center}
 \vspace{-0.5cm}
\end{table*}
\begin{table}[h]
\begin{center}
 \caption{Accuracy (\%) with Various Propagation Steps in GNN Layers. }
\begin{tabular}{|l| p{4.5 mm} p{4.5 mm} p{4.5 mm} p{4.5 mm} p{4.5 mm} p{4.5 mm} p{4.5 mm}| }
 \hline
Iteration & \#1 &\#2 &\#3  &\#4 &\#5 &\#8 &\#10 \\
    \hline
Aircraft & \textbf{95.4} &94.7 &94.9  &94.7 &94.8 &94.8 &94.4\\
Dogs & \textbf{97.3} &97.1 &97.1 &97.0 &96.9 &97.1 &96.9  \\
Flowers & \textbf{97.9}  &97.7 &97.3 &97.8 & 97.6 &97.8 &97.6\\ 
\hline
\end{tabular}
 \label{table:iter}
 \end{center}
 \vspace{- 0.5 cm}
\end{table}

\section {Ablation study}  \label{Ablation}
The ablative study is conducted from several important aspects: suitability of using uniform-grid as an alternative to our adopted region proposals, exploring SotA graph-based pooling techniques \cite{xu2018powerful}, \cite{zhang2018end}, efficacy of each key components of \texttt{SR-GNN}, impact of the number of regions ($\mathcal{R}$) on recognition accuracy, influence of GNN layer's output dimensionality in performance, and the number of power-iterations in GNN layers. 

\subsubsection{Formation of Region Proposals and Key Modules} 
The regions with variable areas and aspect ratios that are akin to computing the HOG cells and blocks are preferred here. 
ViT uses uniform regions such as  16$\times$16 or 14$\times$14 for an image resolution of 224$\times$224. Inspired by this, our method is tested  with a uniform grid-structure as an alternative for generating region proposals. The results with 4$\times$4 grid (region-size is 16$\times$16 for up-sampled feature resolution of 64$\times$64) on four datasets are given in Table \ref{table:Abln1} (row 1), which is the best among other grid-sizes of 2$\times$2 (region-size: 32$\times$32), 3$\times$3 (region-size: 21$\times$21), and 5$\times$5 (region-size:  13$\times$13). Even though the accuracy with a regular grid of 4$\times$4 is better than many existing approaches (reported in Table \ref{table:sota_comp}), it is not a suitable choice for generating regions to learn finer details. This is evident from the accuracy gain of \texttt{SR-GNN} using regions of different aspect ratios and areas in comparison to the 4$\times$4 uniform grid. These gains are: 3.2\% over Cars, 1.1\% over Aircraft, 1.0\% over Dogs, and 0.9\% over Flowers. These results show that regular regions are not pertinent enough to capture subtle variations for spatial relation modeling among the regions. Moreover, \texttt{SR-GNN} outperforms vision Transformers (Table \ref{table:sota_comp}) that use regular regions but fail to capture the overall object structure, and hence, do not achieve superior results as ours. Also, Mask-RCNN is used for region proposals in CPM \cite{ge2019weakly} which has attained 1.5\% and 0.2\% lower accuracy than ours over the respective CUB and Dogs datasets (Table \ref{table:sota_comp}). Thus, all these results justify the benefits of our method in exploring multi-scale regions. 

We have evaluated the efficacy of the existing SotA graph-based pooling methods to compare the performance with the chosen gated attentional pooling to pool features from the nodes of the relation-aware GNN. These are the global average pooling, global max pooling, global sum pooling \cite{xu2018powerful}, and sort pooling \cite{zhang2018end}. The results are shown in Table \ref{table:Abln1}. It is evident that the gated attentional pooling performs better than these alternative methods. This is mainly because the gated attentional pooling uses element-wise sigmoid and acts as a soft attention mechanism that decides which nodes (regions) are more relevant to the current graph-level classification by selecting the most discriminative features from the regions and is thus more suitable to capture their subtle variances than the other pooling approaches. Next, the impact of varied key modules (shown in Fig. \ref{fig:full_model}) are evaluated over 4 datasets (Table \ref{table:Abln1}). It includes only self-attention to refine local features \textit{i.e.}, without (W/o) GNN, without (W/o) self-attention and self-attention without (W/o) weighted-attention. These results justify the importance of each key component in our \texttt{SR-GNN}, without which accuracy degrades significantly. 

\subsubsection{Number of region proposals}
The impact of different numbers ($\mathcal{R}$) of regions on the accuracy of our \texttt{SR-GNN} is given in Table \ref{table:RoI}. The regions are generated by varying the cell size (Section \ref{sec:roi}) as suggested in \cite{behera2020regional}. Four regions are shown in Fig. \ref{fig:regions}, and the rest are given in the supplementary document. Different regions are generated by controlling the HOG's cell size. The best accuracy is achieved for cell size of 14$\times$14 \textit{i.e.},  $\mathcal{R}$=27. Moreover, our model complexity and per-image inference time with increasing numbers of regions are presented in Table \ref{table:time}. The number of trainable parameters does not depend on the number of regions, whereas the GFLOPs and the per-image inference time increase with the number of regions, as expected.

\subsubsection{Impact of the neighborhood size of a given node on accuracy}
The neighborhood size of a given node in our \texttt{SR-GNN} (Fig. \ref{fig:full_model}(b)) is controlled by $\alpha$ in (\ref{eq_2}). In order to measure its impact on accuracy, we evaluate various values of $\alpha \in [0.1, 0.8]$ on the Aircraft, Cars, Dogs and Flowers datasets. The results for GNN output dimensions of 512 and 1024 are shown in Table \ref{table:APPNP_alpha2}. It is clear that the accuracy increases with the increasing value of $\alpha$ and reaches a maximum of around 0.3, suggesting the optimal size of the local neighborhood of a given node.  We observe a similar trend for the GNN layers with output dimensions of 512 and 1024, respectively. However, for the Dogs and Cars datasets, the accuracy is slightly higher for the latter. This is because different graphs characterize different neighborhood structures. 

\subsubsection{Number of power iteration steps in GNN}
We have assessed the performance with  various power iteration steps $K$ in (\ref{eq_2}) in the GNN layers. The iteration steps are varied from $K$$=$$1$ to $K$$=$$10$, and the results are given in Table \ref{table:iter}. For the Aircraft, the accuracy slightly decreases as $K$ increases. This could be because our \texttt{SR-GNN} advances closer to the global PageRank solution after the first iteration. However, the accuracy variations are marginal for the Dogs and Flowers datasets, and we achieve the best performance with a single propagation step in GNN for all the datasets. This is desired in real-world applications for computational efficiency without loss of accuracy. 

\vspace{-0.2 cm}
\section{Conclusion} \label {conclusion}
We have proposed a novel end-to-end deep network called \texttt{SR-GNN} to enhance the recognition accuracy of fine-grained objects and human-actions, avoiding any object-parts bounding-box annotation. The model introduces an innovative relation-aware visual feature transformation and its refinement via attentional spatial context modeling to enrich region-level description to capture  subtle variations observed and required in FGVC. The model has also proposed a gated attentional pooling to automatically aggregate the relation-aware transformed features. Ultimately, our model’s SotA quantitative and qualitative results on eight benchmark datasets and ablation study  show the efficacy of \texttt{SR-GNN}. 

In the near future, we will advance our \texttt{SR-GNN} focusing on following key aspects: 1) adapting it to a Graph Transformer Network (GTN) for generating new graph structures to learn a soft selection of connected regions and composite relations for generating useful multi-hop connections to further enhance the recognition accuracy, 2) evaluating SR-GNN on LSVC datasets consisting of distinctive
categories (\textit{e.g.}, ImageNet and COCO), and 3) optimizing and extending it to recognize fine-grained actions and activities in videos.

\section*{Acknowledgement}
This research is supported by the UKIERI-DST grant CHARM (UKIERI-2018-19-10), and Research Investment Fund at Edge Hill University. The GPU used in this research was donated by the NVIDIA. 
\ifCLASSOPTIONcaptionsoff
  \newpage
\fi
\bibliographystyle{IEEEtran}
\bibliography{Ref}
\vspace{-10mm}
\begin{IEEEbiography}[{\includegraphics[width=1in,height=1.25in,clip,keepaspectratio]{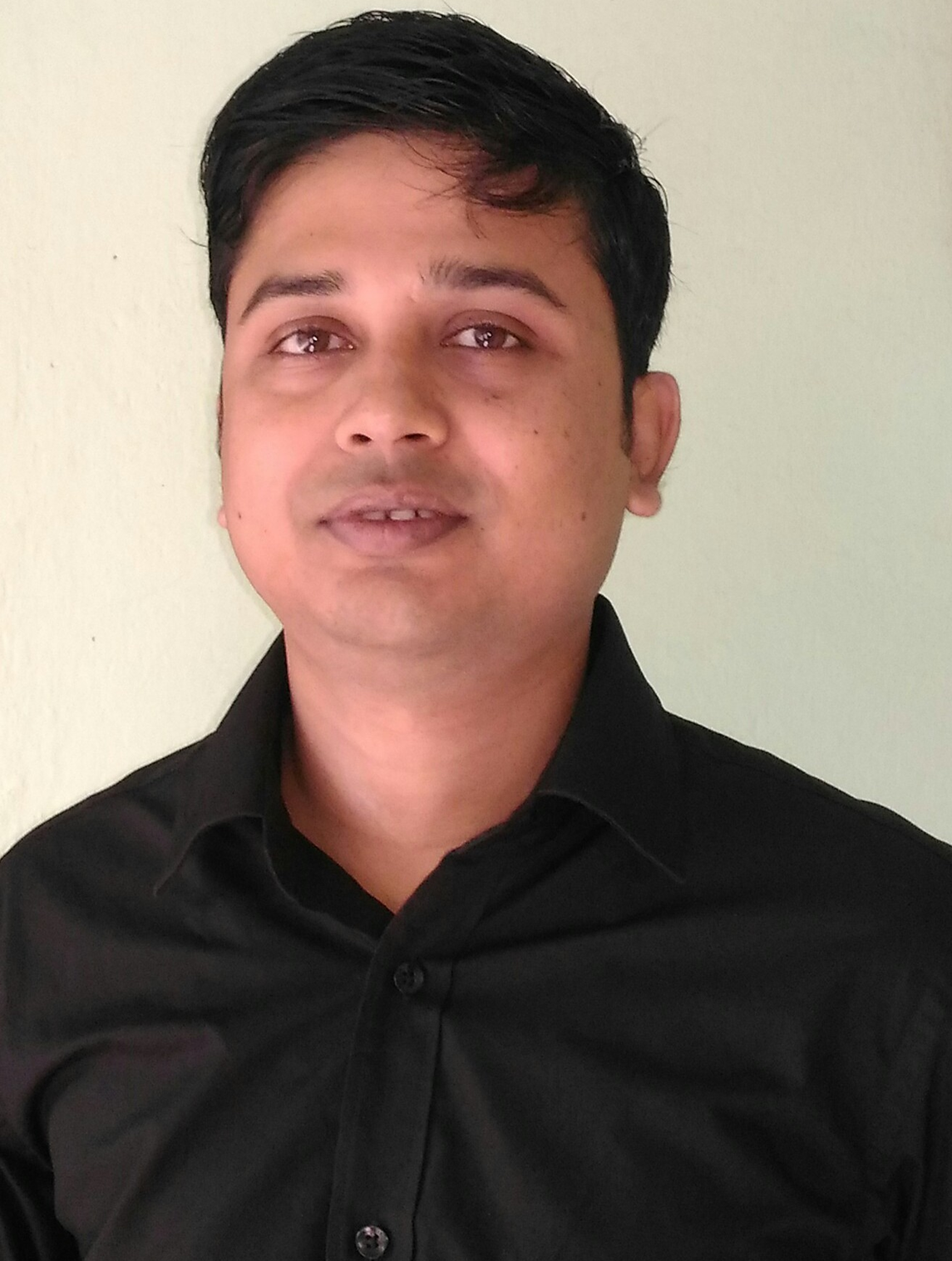}}] {Asish Bera} received the PhD degree from  the Jadavpur University, Kolkata, India,  and the M.Tech degree from the IIEST Shibpur, India. He is currently a Post-doctoral research associate at the Computer Science Department, Edge Hill University, UK. His current research interests include computer vision, deep learning, human activity recognition, and biometrics. He is a member of IEEE.
\end{IEEEbiography}
\vspace{-10mm}
\begin{IEEEbiography}[{\includegraphics[width=1in,height=1.25in,clip,keepaspectratio]{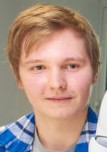}}]{Zachary Wharton} is currently an MRes student in the Department of Computer Science, Edge Hill University, UK. He obtained his Bachelor’s degree in Computing from Edge Hill University in 2019. His interests include computer vision, deep learning, human-robot interaction (HRI) and pattern recognition. 
\end{IEEEbiography}
\vspace{-10mm}
\begin{IEEEbiography}[{\includegraphics[width=1in,height=1.25in,clip,keepaspectratio]{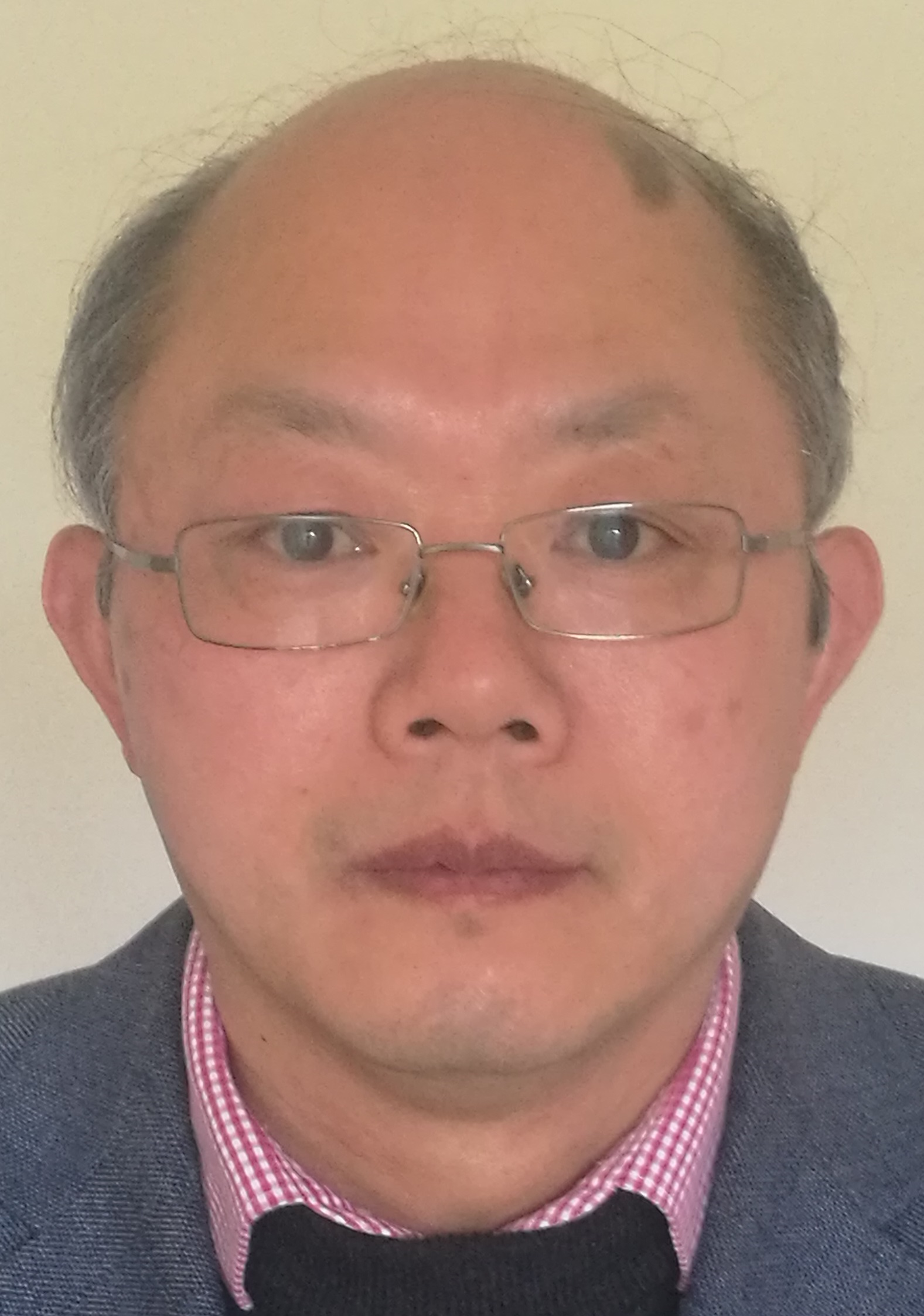}}]{Yonghuai Liu} is a professor and director of the Visual Computing Lab at Edge Hill University since 2018. He obtained his first PhD in 1997 from Northwestern Polytechnical University, P.R. China and second PhD in 2001 from The University of Hull, UK. He is an area/associate editor or editorial board member for a number of journals and conferences. He has published more than 180 papers in the top-ranked conferences and journals. His research interests lie in 3D computer vision, image processing, pattern recognition, machine learning, AI, and intelligent systems. He is a senior member of IEEE, Fellow of BCS, and Fellow of HEA of the UK.
\end{IEEEbiography}
\vspace{-10mm}
\begin{IEEEbiography}[{\includegraphics[width=1in,height=1.25in,clip,keepaspectratio]{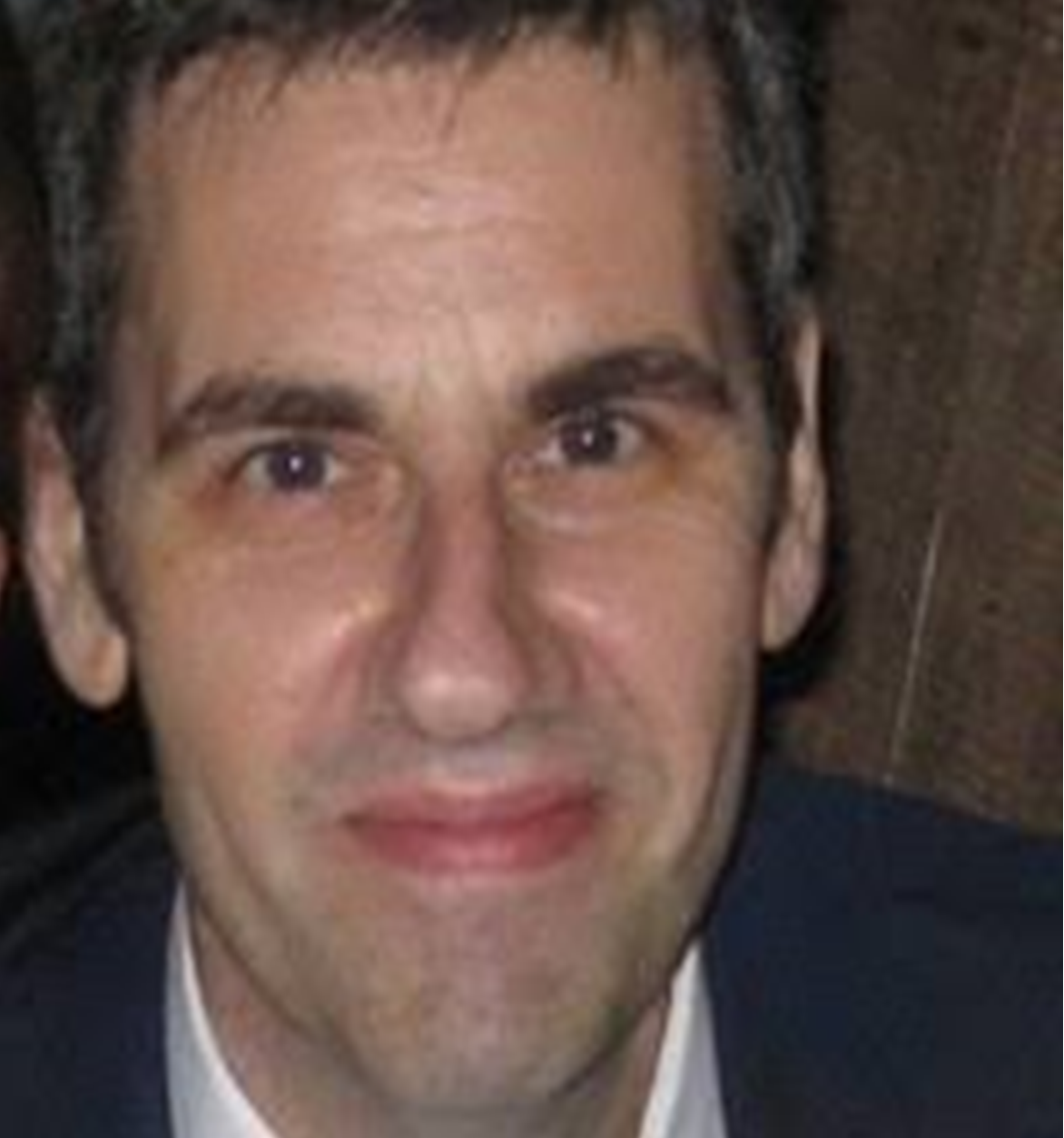}}]{Nik Bessis} received his BA from the T.E.I. Athens and his MA and PhD degrees from De Montfort University, UK. He is a full Professor (2010) and since 2015, the Head (Chair) of the Department of Computer Science at Edge Hill University, UK. He is a FHEA, FBCS and a senior member of IEEE. His research is on social graphs for network and big data analytics as well as developing data push and resource provisioning services in IoT, FI and inter-clouds. He is involved in a number of funded research and commercial projects in these areas. Prof Bessis has published over 300 works and won 4 best papers awards.
\end{IEEEbiography}
\vspace{-10mm}
\begin{IEEEbiography}[{\includegraphics[width=1in,height=1.25in,clip,keepaspectratio]{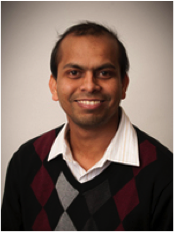}}]{Ardhendu Behera} received the PhD degree in Computer Science from the University of Fribourg, Switzerland and MEng degree in System Science and Automation from the Indian Institute Science (IISc) Bangalore, India. He is currently a Reader in Computer Vision \& AI in the Department of Computer Science, Edge Hill University, UK. He has worked as a Research Fellow and Senior Research Fellow in Computer Vision Group at the University of Leeds. He is a Fellow of HEA and member of {IEEE}, British Machine Vision Association, Applied Vision Association, British Computing Society, affiliated member of IAPR and ECAI. His main interests include computer vision, deep learning, human-robot social interaction, activity analysis and recognition.
\end{IEEEbiography}

\clearpage
\begin{center}
\Large
\textbf{ Supplementary Document}
\large
\end{center}
\vspace{ 0.5 cm}
The accuracy of our \texttt{SR-GNN} is higher than the state-of-the-art on diverse datasets. To justify the benefits of our model, a precise comparison with the existing top-10 methods focused on the inference time is given in Table \ref{table:InfT}, irrespective of the GPU/hardware configuration, deep learning tools (\textit{e.g.}, TensorFlow, PyTorch, MXNet, etc.) and related experimental constraints used in those works. Table \ref{table:InfT} is incorporated in Table \ref{table:Complx2} with the model parameters and GFLOPs by comparing with the top-5 SotA approaches. Some approaches in Table \ref{table:InfT} do not provide these two metrics. We have also provided the accuracy comparison with those works to reflect a trade-off between the accuracy (\%) and inference time in milliseconds (ms). For this purpose, we have specified the best performance of those referred works on a FGVC dataset and our performance and accuracy gain (in parenthesis) on the same dataset.

\begin{table*} [!h]
\begin{center}

 \caption{Performance comparison based on inference time with the SotA methods}
 \label{table:InfT}
\begin{tabular}{|p{5.5 mm}| p{30 mm} p{14 mm} p{15 mm}p{19 mm} p{25 mm} p{34 mm} |}
 \hline
Sl. No & Method & Param (M) & GFLOPs (B) &
 Infer. time per img. (ms) & Accuracy (\%) (dataset) & Our accuracy and (+gain) in \% \\
\hline
1 &	CAP \cite{behera2021context} & 34.2 & 10.2 &	4.2 &	94.9 (Aircraft) &	95.4 (+ 0.5) \\  \hline
2a &	MRDMN-L \cite{xu2021multiresolution} & 51.2 &	14.0 &	4.9 &	89.0 (Dogs) &	96.5 (+7.5), W/o refine  97.3 (+8.3), \texttt{\texttt{SR-GNN}} \\  \hline
2b & \textbf{\texttt{SR-GNN} (W/o Refine)}& 24.4 &	9.3	 &4.9 &	 Paper Table II &	Section IV-D \\  
3 &	\textbf{\texttt{SR-GNN}  (Full Model)}  & 30.9 &	9.8 &	5.0	 &	 &	 \\  \hline
4 &	AG-Net \cite{bera2021attend} & 54.8 &10.4
 &	5.2 &	97.8 (Stanf.40) &	98.8 (+1.0) \\  \hline
5 &	TASN \cite{zheng2019looking} & 37.3 &	21.9 &	7.5	 &87.9 (CUB-200) &	91.9 (+4.0) \\  \hline
6 &	WARN \cite{lopez2020pay} & - &-  &	11.3 &	85.6 (CUB-200) &	91.9 (+6.3) \\  \hline
7&	RG \cite{huang2020interpretable} & - &-	 &23.8 &	87.3 (CUB-200) &	91.9 (+4.6) \\  \hline
8 &	SCAPNet \cite{liu2021learning} & - &- &	24.4 &	93.6 (Aircraft) &	95.4 (+1.8) \\  \hline
9 &	ME-ASN \cite{zhang2021enhancing} & - &- &	33.9 &	89.5 (CUB-200) &	91.9 (+2.4) \\  \hline
10 &	NTS-Net \cite{yang2018learning} & - &- &	35.0 &	93.9 (Cars) &	96.1 (+2.2) \\  \hline
11 &	RA-CNN \cite{fu2017look} & - &- &	36.9 &	87.3 (Dogs)	 &97.3 (+10.0) \\  \hline
\end{tabular}
 \end{center}
\end{table*}
\begin{figure*}[h]
\renewcommand*\thesubfigure{\arabic{subfigure}}
    \centering
    \subfloat[]{\includegraphics[width=0.11\textwidth] {Figures/r0.jpg}}\hfill
        \subfloat[]{\includegraphics[width=0.11\textwidth] {Figures/r1.jpg}}\hfill
            \subfloat[]{\includegraphics[width=0.11\textwidth] {Figures/r2.jpg}}\hfill
                \subfloat[]{\includegraphics[width=0.11\textwidth] {Figures/r3.jpg}}\hfill
                    \subfloat[]{\includegraphics[width=0.11\textwidth] {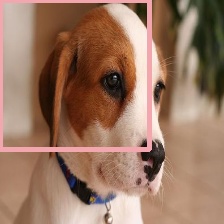}}\hfill
                        \subfloat[]{\includegraphics[width=0.11\textwidth] {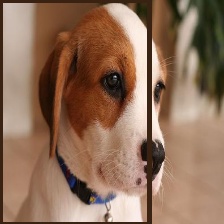}}\hfill
                            \subfloat[]{\includegraphics[width=0.11\textwidth] {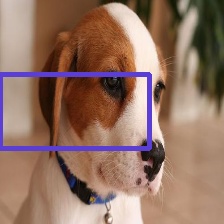}}\hfill
                                \subfloat[]{\includegraphics[width=0.11\textwidth] {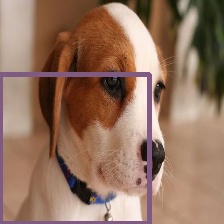}}\hfill
                                    \subfloat[]{\includegraphics[width=0.11\textwidth] {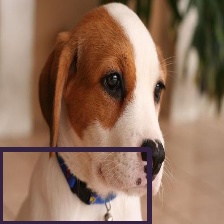}}\\
    \subfloat[]{\includegraphics[width=0.11\textwidth] {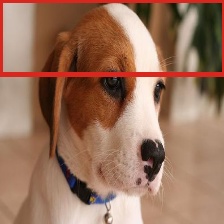}}\hfill
        \subfloat[]{\includegraphics[width=0.11\textwidth] {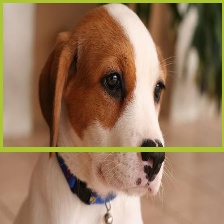}}\hfill
            \subfloat[]{\includegraphics[width=0.11\textwidth] {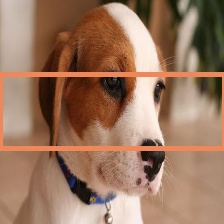}}\hfill
                \subfloat[]{\includegraphics[width=0.11\textwidth] {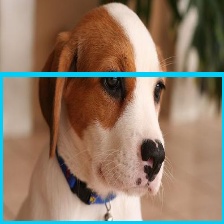}}\hfill
                    \subfloat[]{\includegraphics[width=0.11\textwidth] {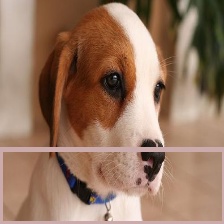}}\hfill
                        \subfloat[]{\includegraphics[width=0.11\textwidth] {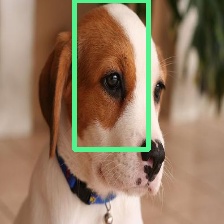}}\hfill
                            \subfloat[]{\includegraphics[width=0.11\textwidth] {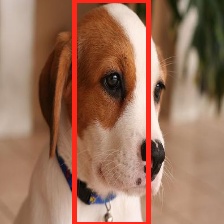}}\hfill
                                \subfloat[]{\includegraphics[width=0.11\textwidth] {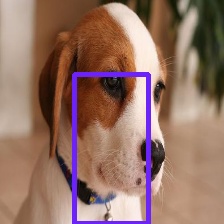}}\hfill
                                    \subfloat[]{\includegraphics[width=0.11\textwidth] {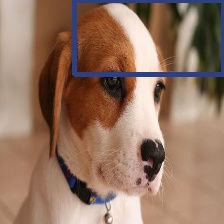}}\\
    \subfloat[]{\includegraphics[width=0.11\textwidth] {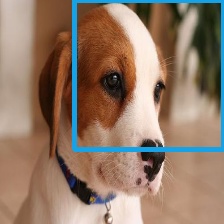}}\hfill
        \subfloat[]{\includegraphics[width=0.11\textwidth] {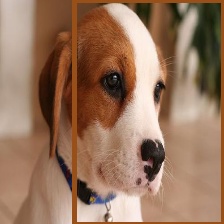}}\hfill
            \subfloat[]{\includegraphics[width=0.11\textwidth] {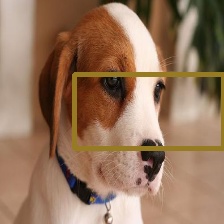}}\hfill
                \subfloat[]{\includegraphics[width=0.11\textwidth] {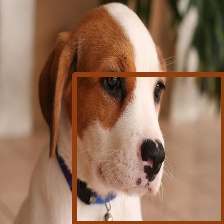}}\hfill
                    \subfloat[]{\includegraphics[width=0.11\textwidth] {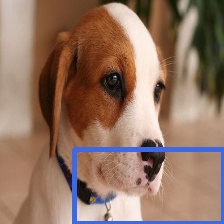}}\hfill
                        \subfloat[]{\includegraphics[width=0.11\textwidth] {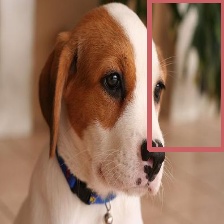}}\hfill
                            \subfloat[]{\includegraphics[width=0.11\textwidth] {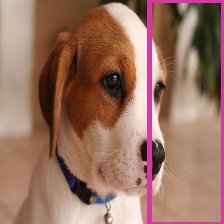}}\hfill
                                \subfloat[]{\includegraphics[width=0.11\textwidth] {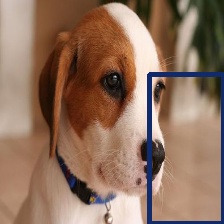}}\hfill
                                    \subfloat[]{\includegraphics[width=0.11\textwidth] {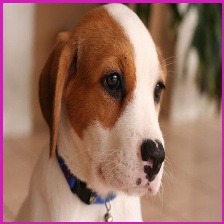}}\hfill
        
\caption{Bounding box displaying the optimal number ($\mathcal{R}=27$) of patches/regions in a given input image. These regions are used for bilinear pooling from the upsampled CNN features in Fig. 2(a) (Section III.B). The last region (\#27) is the whole image.}
    \label{fig:regions}
\end{figure*} 

It shows that our \texttt{SR-GNN} (full-model) stands in the third position and outperforms other eight SotA approaches based on the inference time. From this study, it is evident that \texttt{SR-GNN} requires very competitive inference time with 0.8 ms more than CAP \cite{behera2021context}. It is noted that our \texttt{SR-GNN} without Refine module (4.9 ms) shares the second position with MRDMN-L \cite{xu2021multiresolution} and achieves 7.5\% accuracy gain on the Dogs dataset over this approach. On the contrary, \texttt{SR-GNN} computationally lighter and requires lesser parameters and GFLOPs than these two methods, mentioned in Table \ref{table:Complx2}  of revised manuscript. Also, the accuracy gain of \texttt{SR-GNN} is the highest on various FGVC datasets compared to these works. In this context, it can be noted that \texttt{SR-GNN} offers an excellent balance to maintain the trade-off between the accuracy, model complexity, and inference time over a diverse category of recent approaches. Therefore, \texttt{SR-GNN} performs the best considering all the aspects of experimental analysis over the existing SotA methods. Particularly, it stands as the second (W/o Refine) and third (full-model) regarding the inference time in comparison with the top-10 SotA methods.   

We have included additional visualizations related to our manuscript. 
(A) Fig. \ref{fig:regions} shows all the region proposals ($\mathcal{R}=27$) and it is related to Fig. \ref{fig:regions} in the main paper.

\begin{figure*} [h]
    \centering
    \subfloat[Base CNN within \texttt{SR-GNN}]{\includegraphics[width=0.3\textwidth] {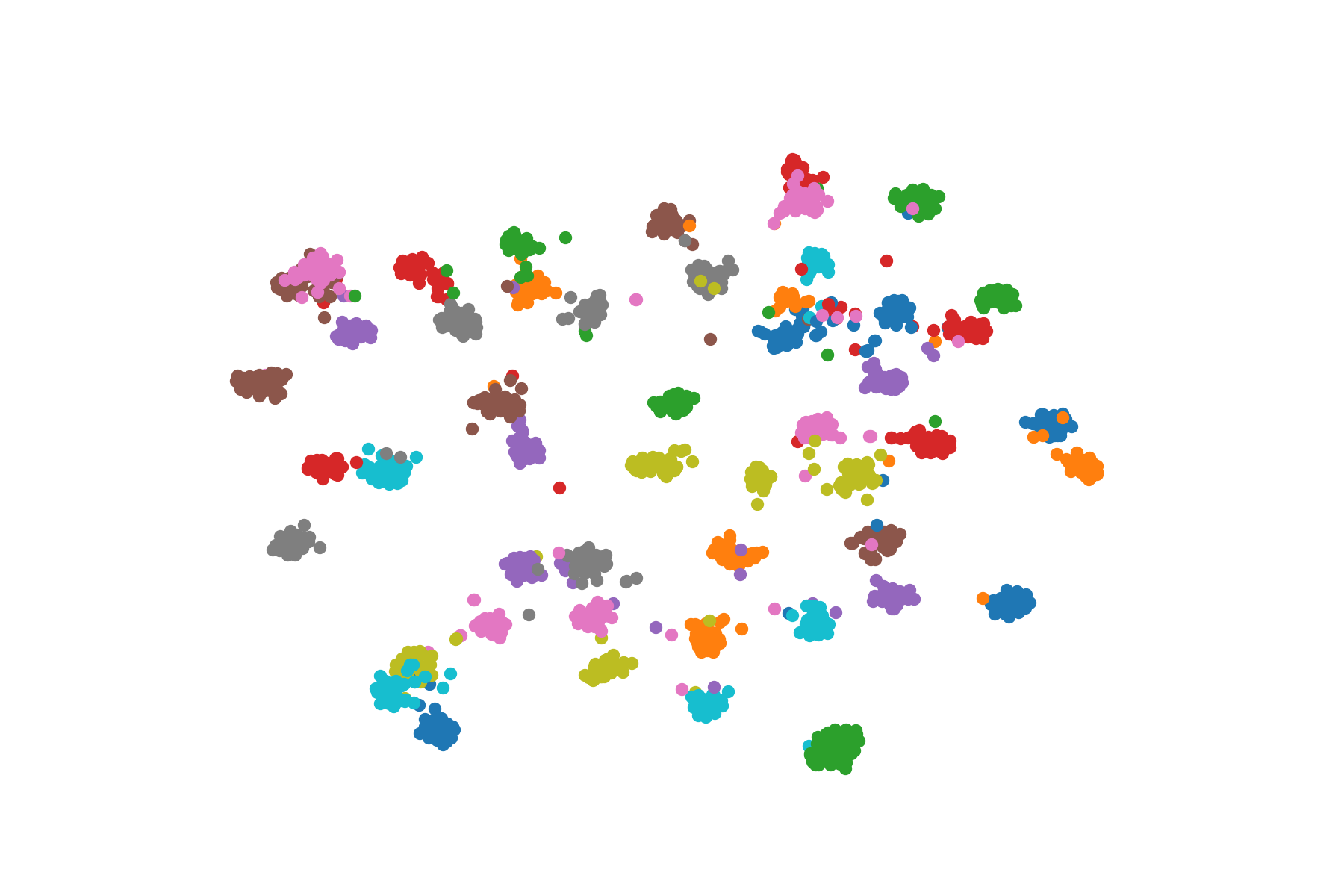}}\hfill
    \subfloat[Transformed Feature $f_t$]{\includegraphics[width=0.3\textwidth] {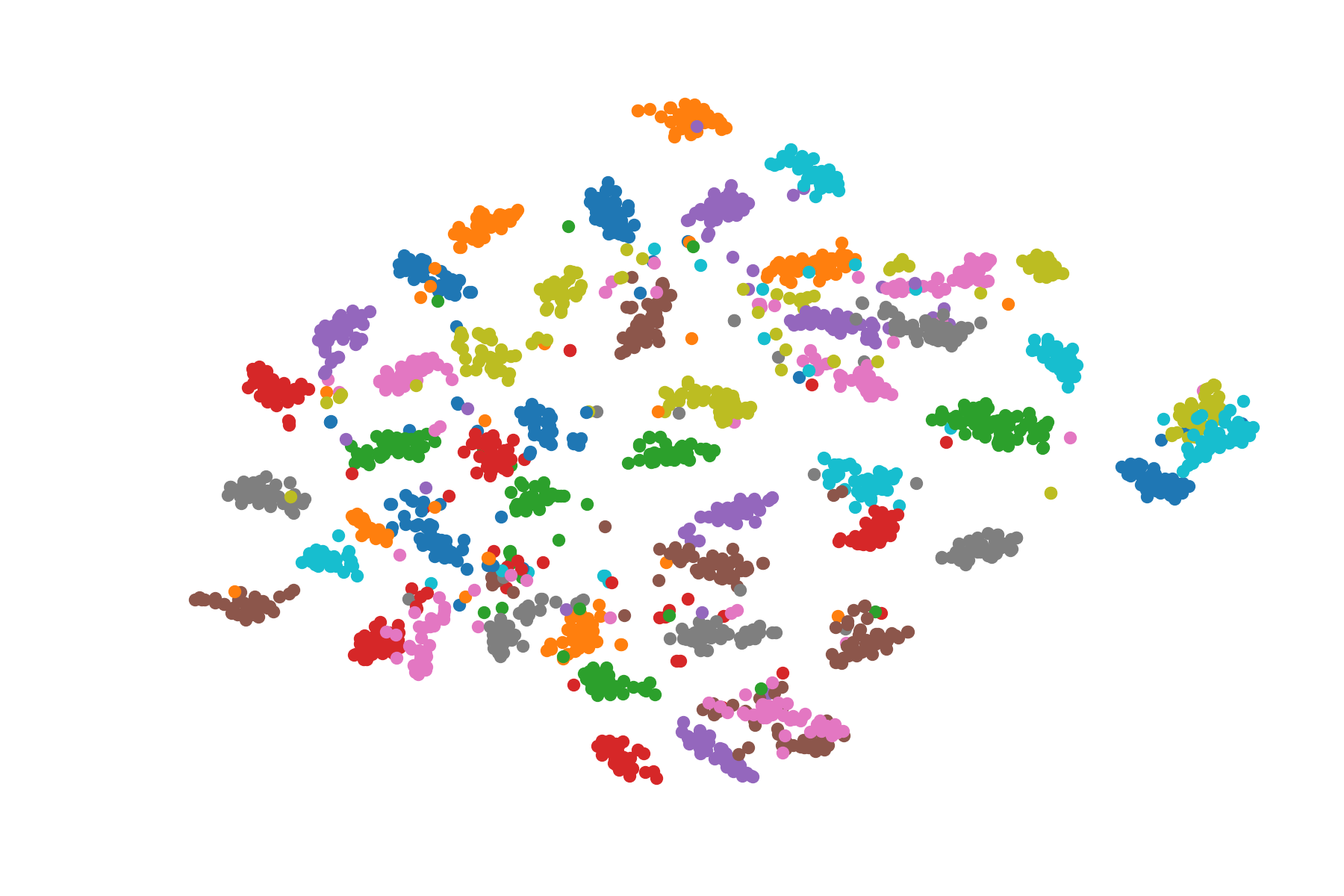}}\hfill
    \subfloat[Context Refinement ($\mathbf{v}$)]{\includegraphics[width=0.3\textwidth] {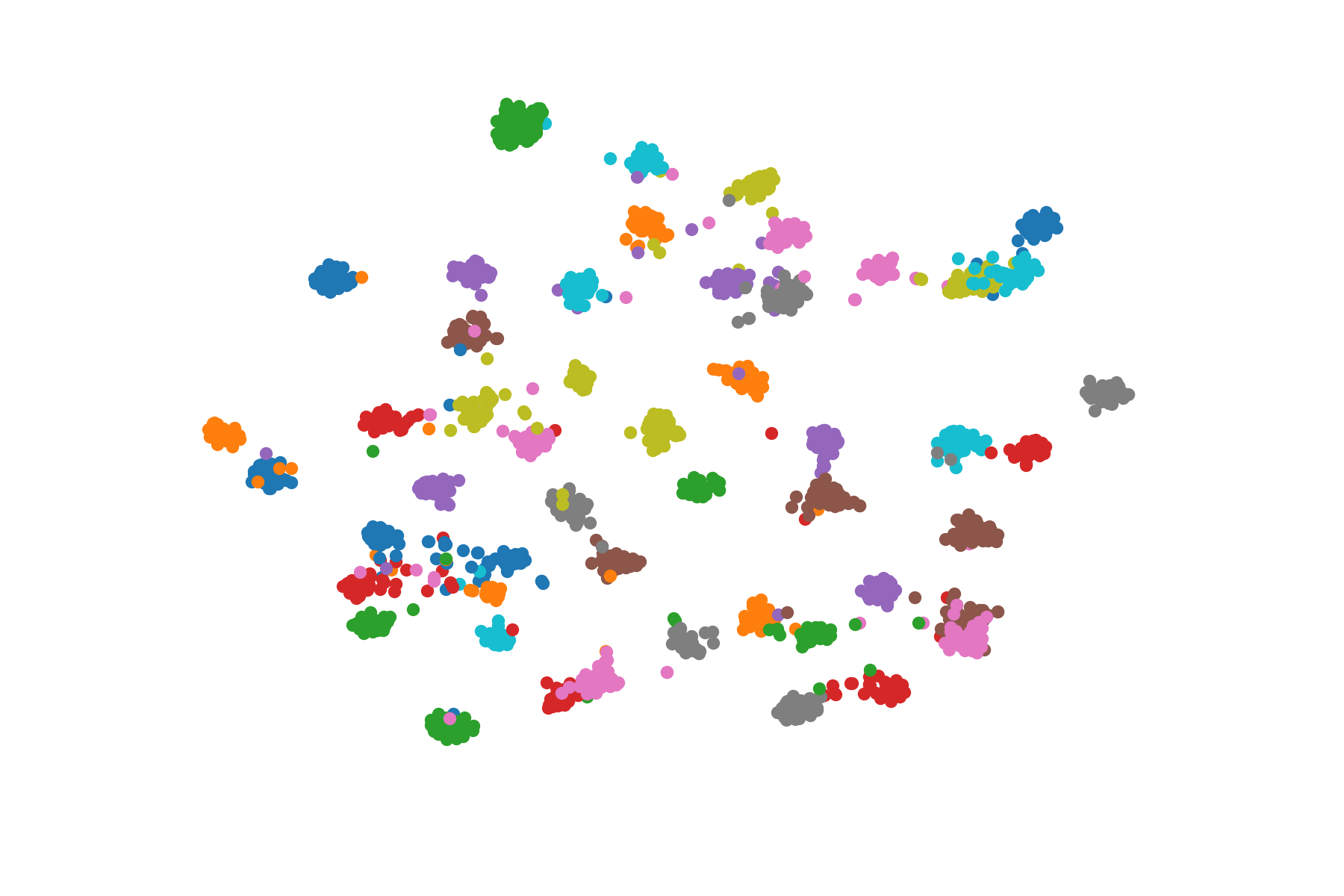}} \\ 
    \subfloat[Final feature  $\bar{f}_t = f_t + f_t \otimes \mathbf{v}$]{\includegraphics[width=0.3\textwidth] {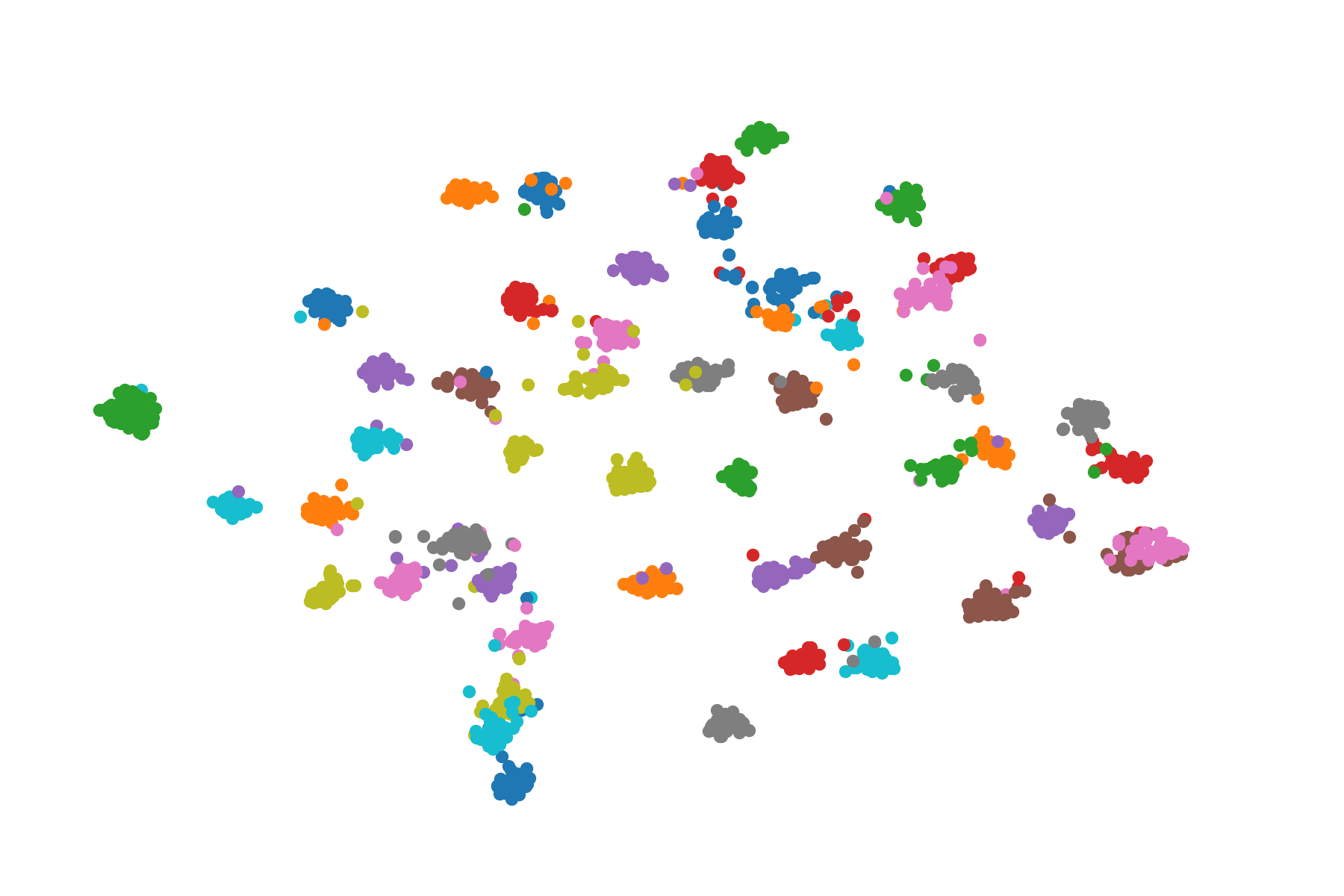}} \hfill
         \subfloat[Final feature $\bar{f}_t = f_t$ \textbf{without} the context refinement (weight $\mathbf{v}$) module]{\includegraphics[width=0.3\textwidth] {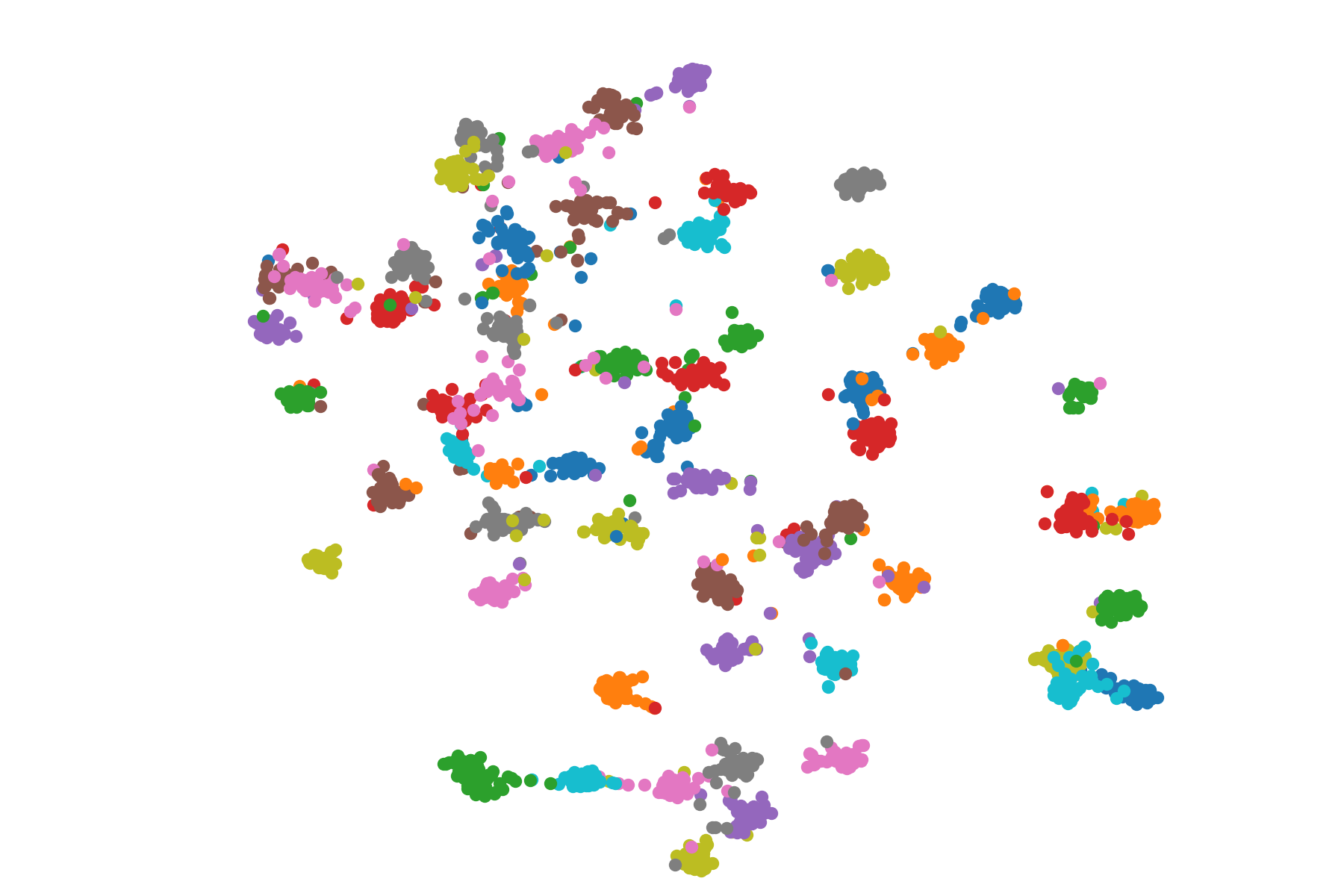}}
     \subfloat[Standalone Base CNN]{\includegraphics[width=0.3\textwidth] {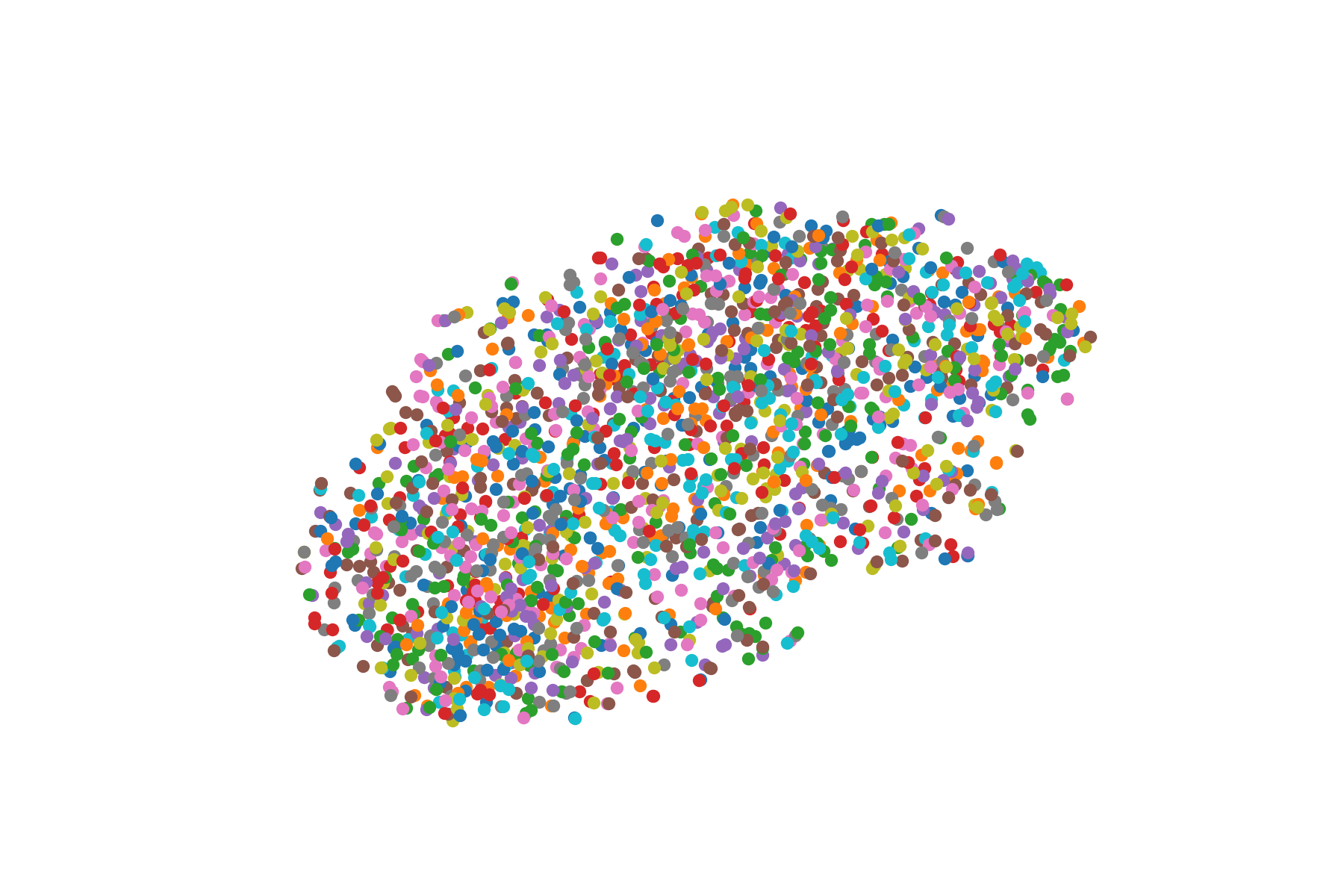}}\hspace{1cm}
\caption{\texttt{SR-GNN}'s discriminability using t-SNE to visualize class separability and compactness using features from a) base CNN (Xception, Fig. 2(a)) within our model, b) relation-aware transformed feature using GCN (Fig. 2(b)), c) attentional context refinement weight-vector $\mathbf{v}$ (Fig. 2(c)), and d) the final image-level feature map $\bar{f}_t$ for classification (Fig. 2(c)). Each color represents a particular class. There are 50 classes chosen randomly from the \textbf{Car's} test set. e) \texttt{SR-GNN} \textbf{without} the context refinement module, and f) Standalone Xception base CNN without our modules (re-trained on the Cars dataset). }
    \label{fig:TSNE_Car}
\end{figure*}

(B)  t-SNE plots related to Table \ref{table:davies_bouldin_score} (in the main paper) for Cars (Fig. \ref{fig:TSNE_Car}) and Flowers (Fig. \ref{fig:fig_attn_car}) datasets.

(C) Joint attention maps are shown on Cars (Fig. \ref{fig:fig_attn_car}) and Flowers (Fig. \ref{fig:fig_attn_flower}) datasets, related to Fig.   \ref{fig:fig_attn} in the main manuscript. \\

\begin{figure*}
    \centering
    \subfloat[Base CNN within \texttt{SR-GNN}]{\includegraphics[width=0.3\textwidth] {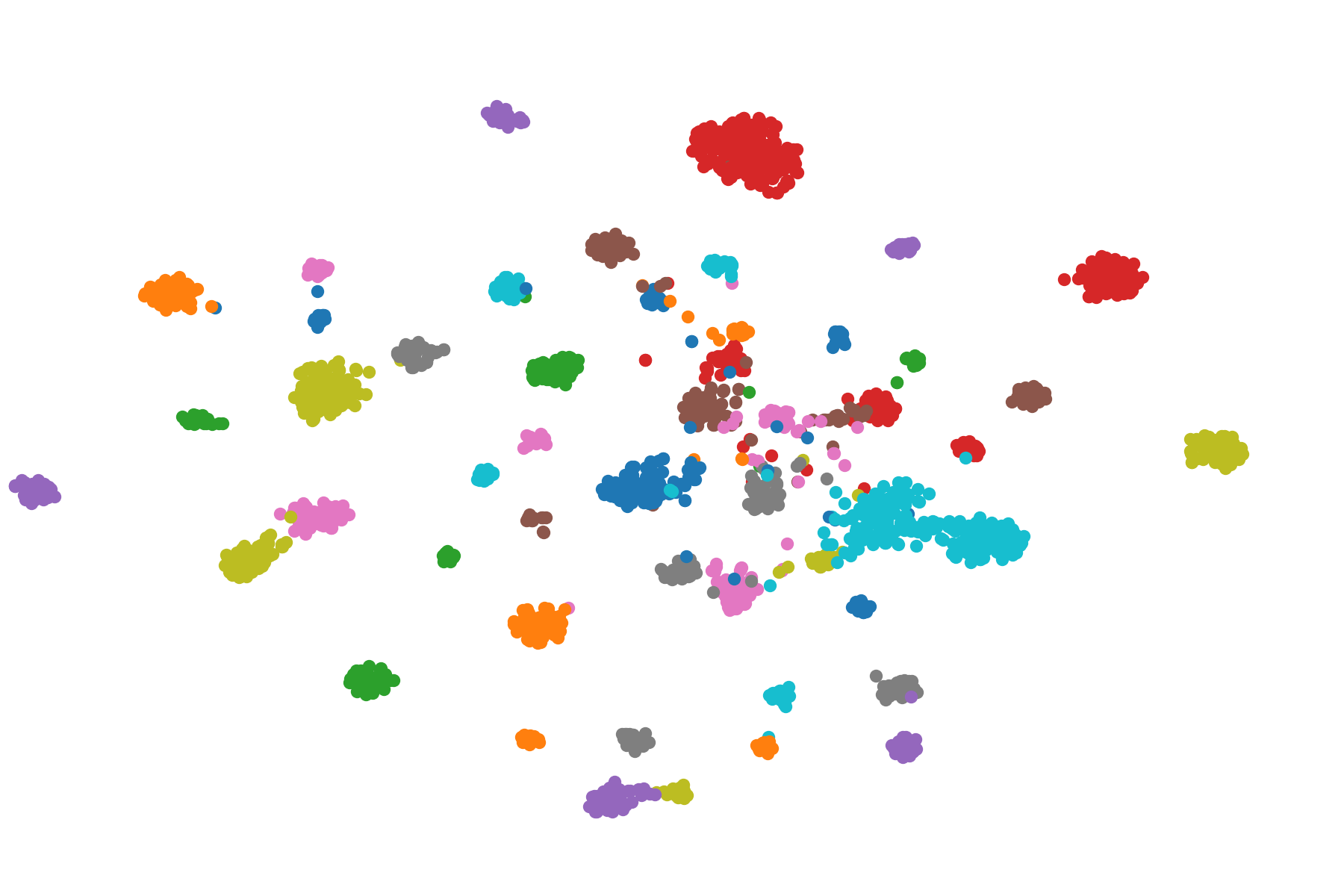}}\hfill
    \subfloat[Transformed Feature $f_t$]{\includegraphics[width=0.3\textwidth] {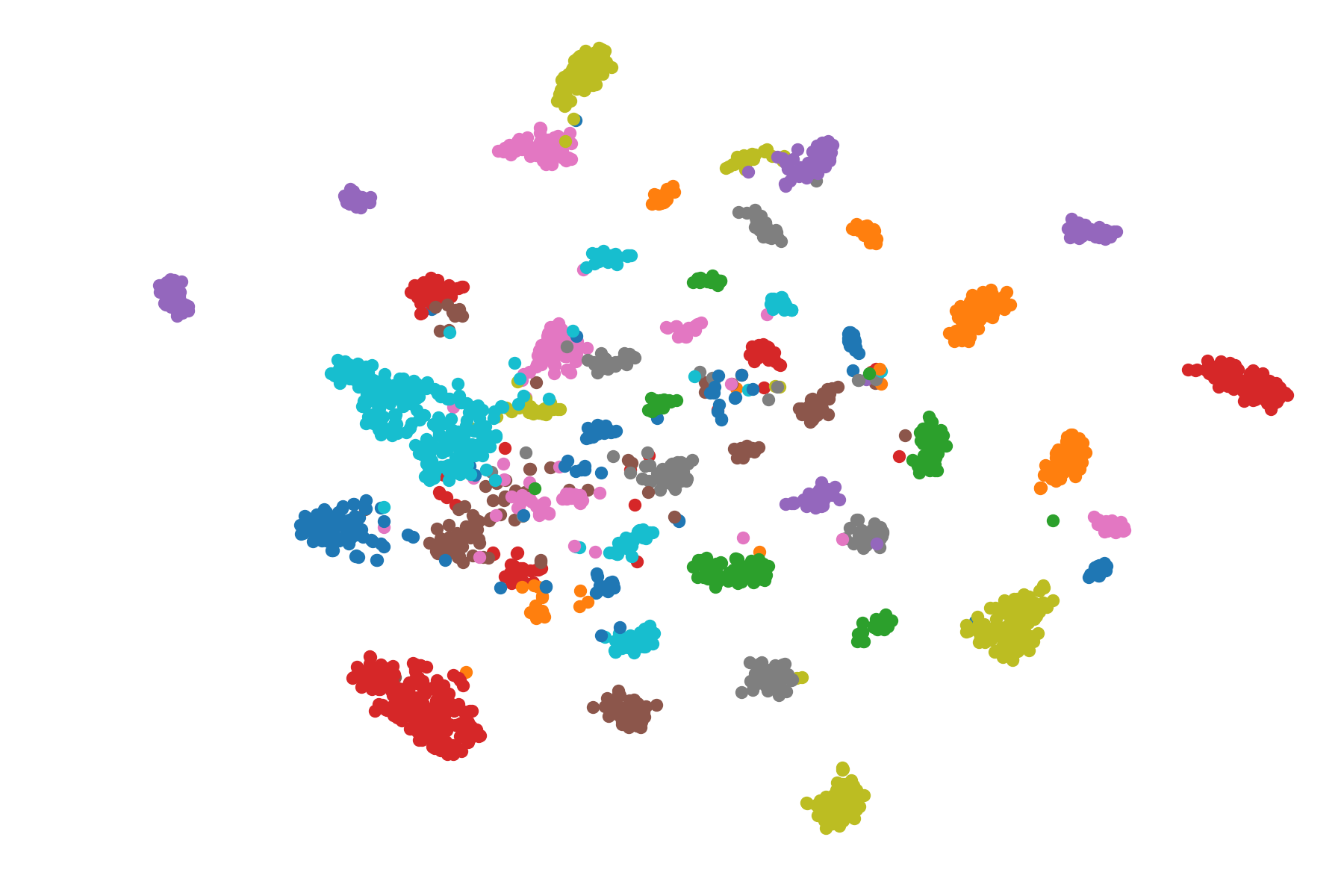}}\hfill
    \subfloat[Context Refinement ($\mathbf{v}$)]{\includegraphics[width=0.3\textwidth] {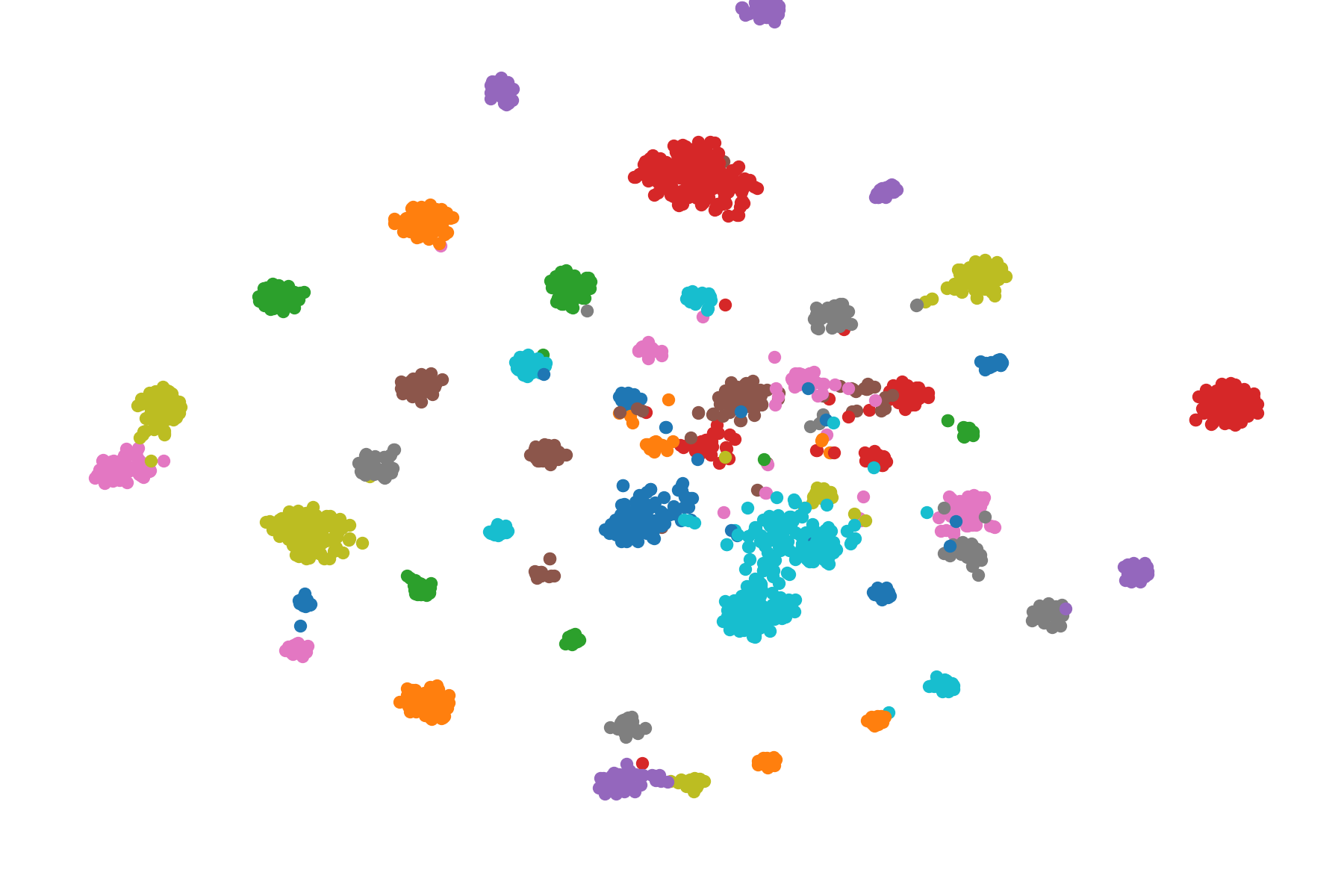}}\hfill \\
    \subfloat[Final feature  $\bar{f}_t = f_t + f_t \otimes \mathbf{v}$]{\includegraphics[width=0.3\textwidth] {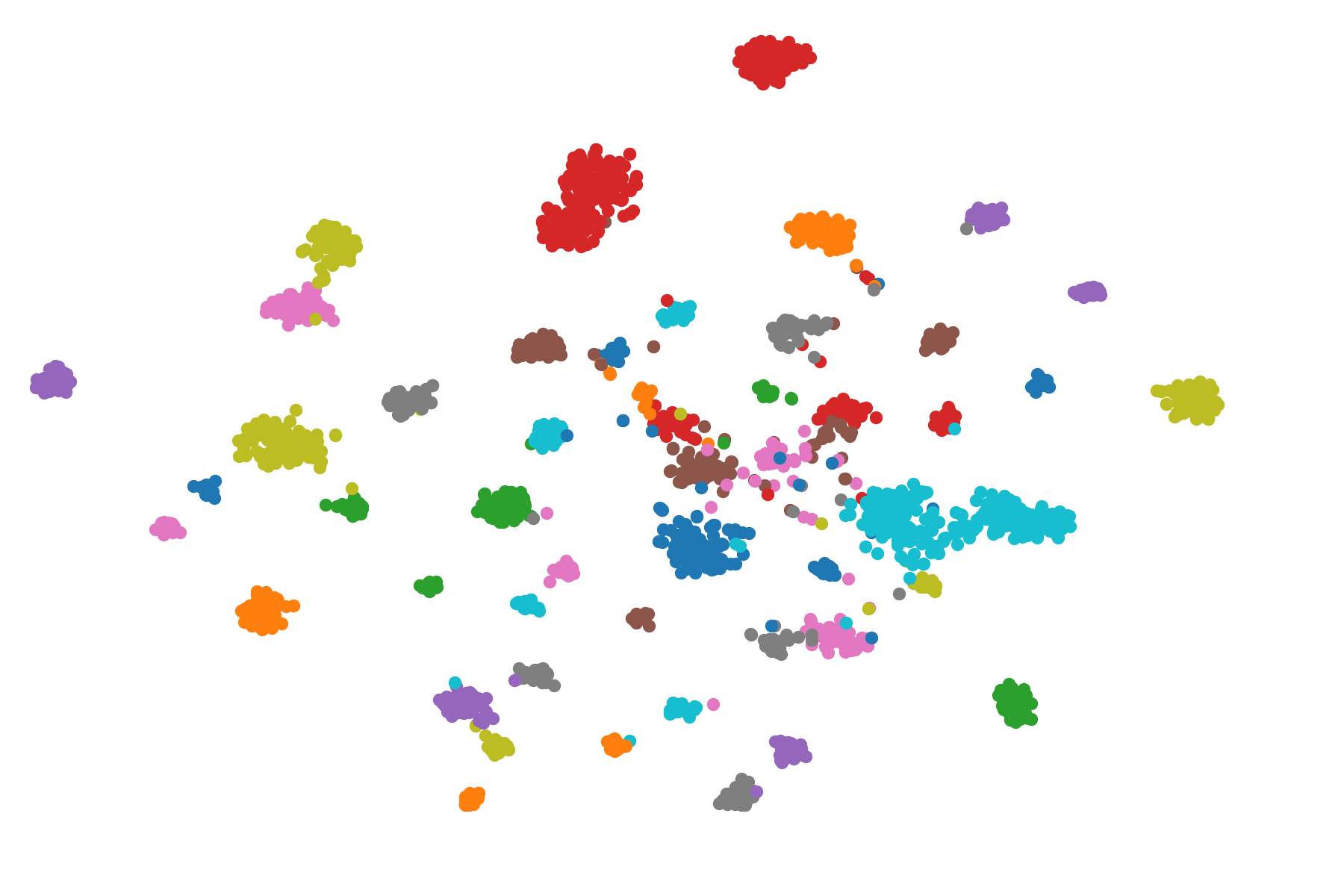}} \hfill
         \subfloat[Final feature $\bar{f}_t = f_t$ \textbf{without} the context refinement (weight $\mathbf{v}$) module]{\includegraphics[width=0.3\textwidth] {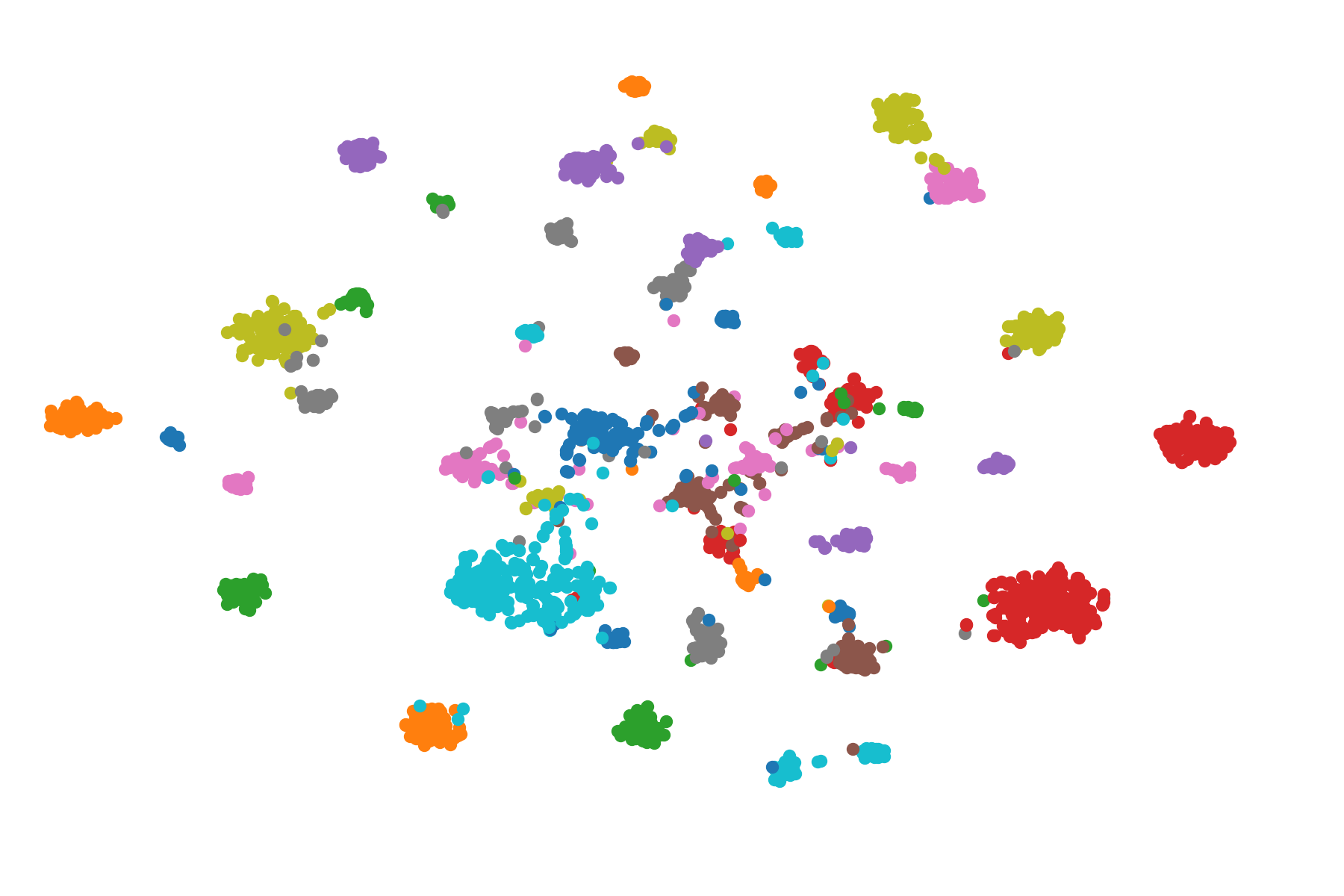}} \hfill
     \subfloat[Standalone Base CNN]{\includegraphics[width=0.3\textwidth] {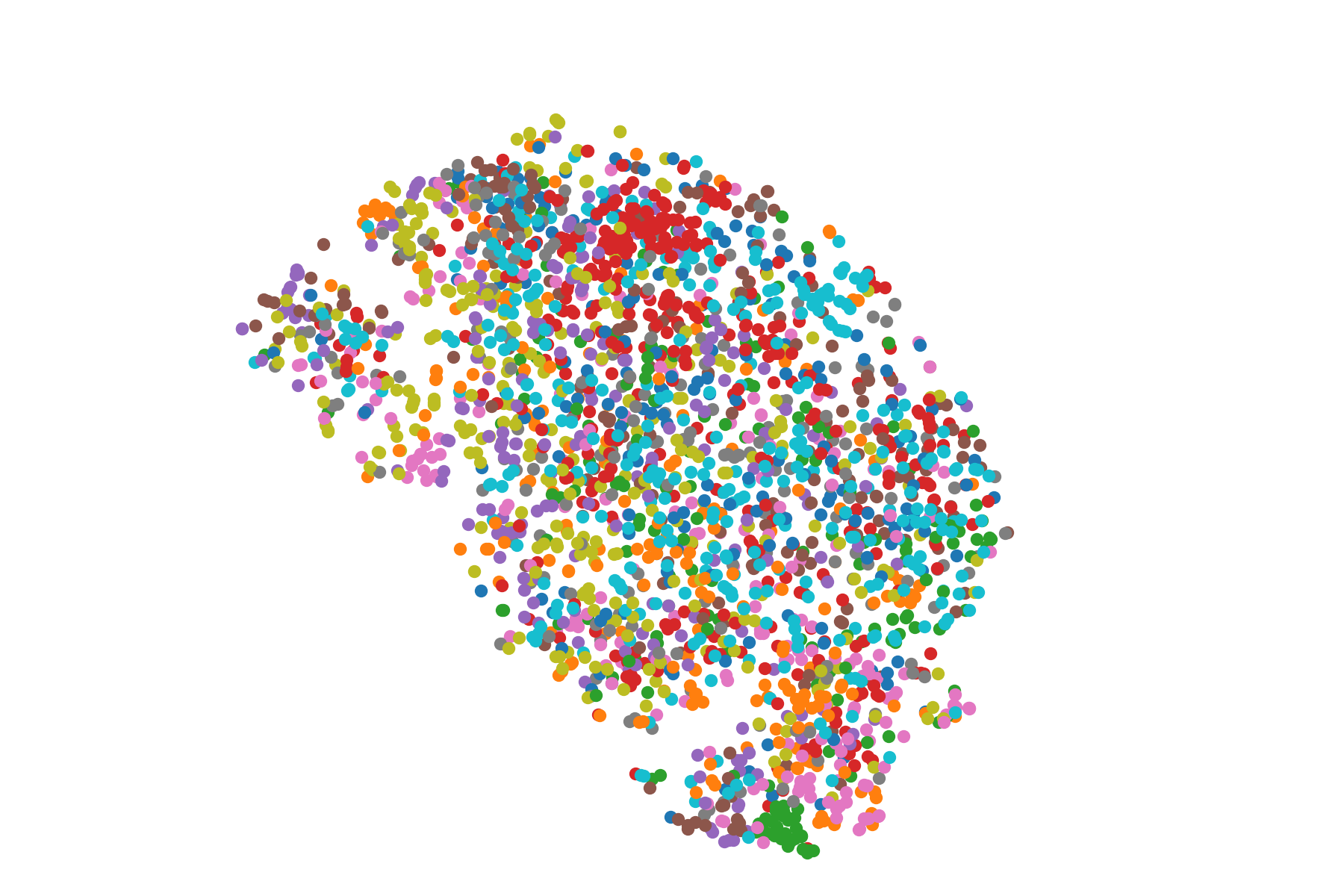}}\hspace{1cm}
    \caption{\texttt{SR-GNN}'s discriminability using t-SNE to visualize class separability and compactness using features from a) base CNN (Xception, Fig. 2(a)) within our model, b) relation-aware transformed feature using GNN (Fig. 2(b)), c) attentional context refinement weight-vector $\mathbf{v}$ (Fig. 2(c)), and d) the final image-level feature map $\bar{f}_t$ for classification (Fig. 2(c)). Each color represents a particular class. There are 50 classes chosen randomly from the \textbf{Flower's} test set. e)  \texttt{SR-GNN} \textbf{without} the context refinement module, and f) Standalone Xception base CNN without our modules (re-trained on the Flowers dataset). }
    \label{fig:TSNE_flower}
    \vspace{ -0.5 cm}
\end{figure*}
\begin{figure*}[h]
    \centering
     \subfloat[AM General Hummer SUV 2000: top-2 regions (9 \& 22) and their top-3 joint attentions (Fig. 2c)]{\includegraphics[width=0.9\textwidth] {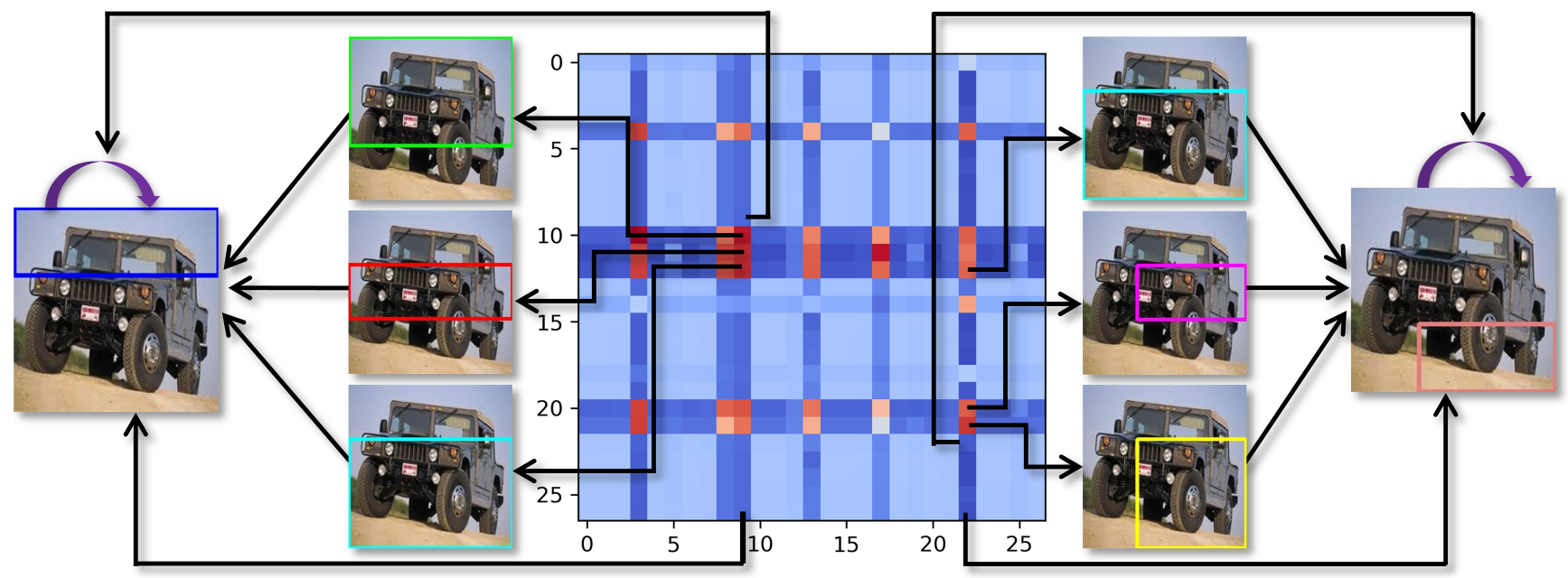}}\hfill
    \subfloat[Ford GT Coupe 2006: top-2 regions (3 \& 17) and their top-3 joint attentions (Fig. 2c)]{\includegraphics[width=0.9\textwidth] {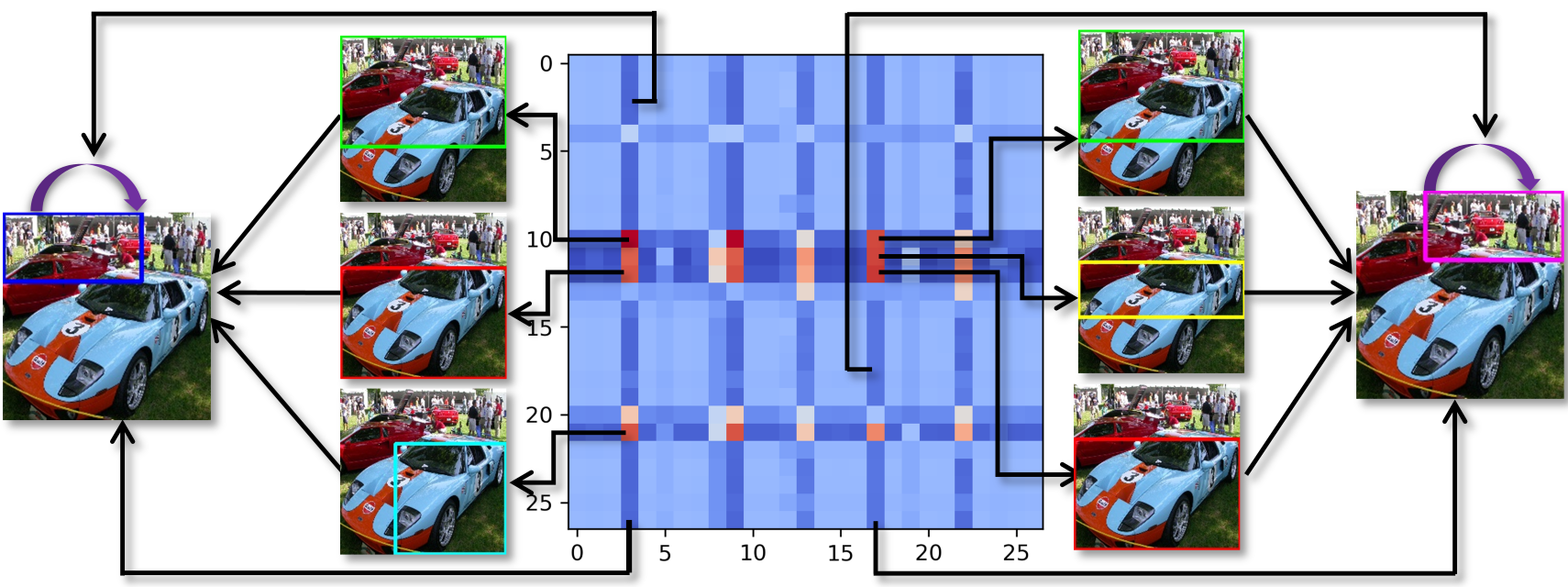}}
    \caption{Visualization within attentional context refinement (Fig. 2(c)): a) Top-2 regions (cols 9 \& 22) contributing towards sub-type `AM General Hummer SUV 2000' conditioned on the respective other top-3 regions (rows) in joint decision-making. The self-attention (self-loop) is also shown in the top-2 regions. b) Similarly, top-2 regions (cols 3 \& 17) contributing towards sub-type `Ford GT Coupe 2006' conditioned on the respective other top-3 regions (rows). Region proposals are shown in the respective original images.}
    \label{fig:fig_attn_car}
\end{figure*}
\begin{figure*}
    \centering
     \subfloat[Flower class 1: top-2 regions (1 \& 5) and their top-3 joint attentions (Fig. 2c)]{\includegraphics[width=1.0\textwidth] {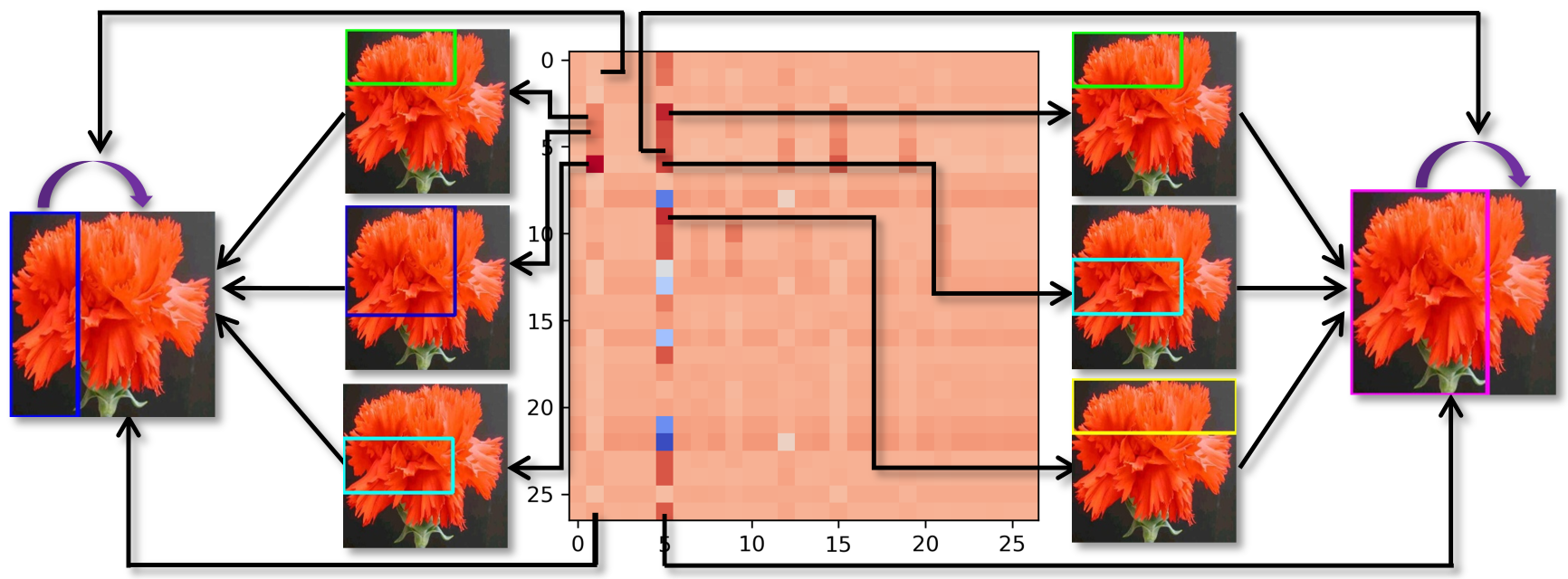}}\hfill
    \subfloat[Flower class 61: top-2 regions (5 \& 12) and their top-3 joint attentions (Fig. 2c)]{\includegraphics[width=1.0\textwidth] {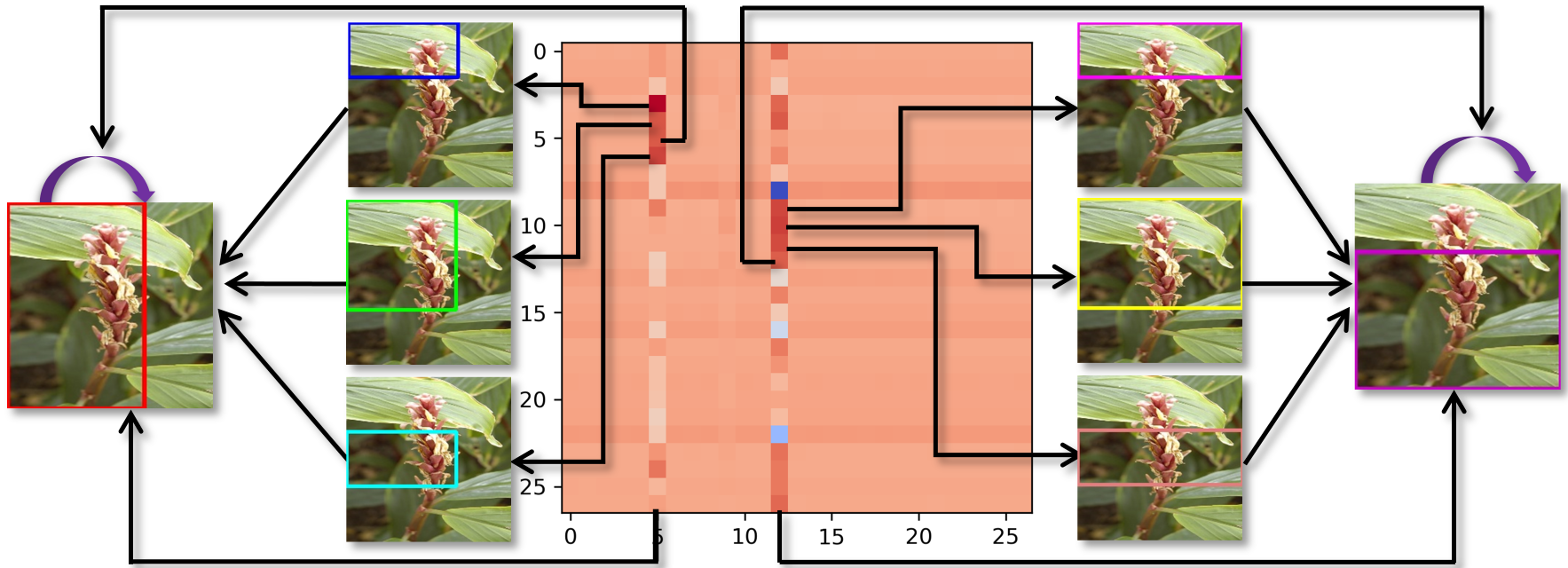}}
    \caption{Visualization within attentional context refinement (Fig. 2(c)): a) Top-2 regions (cols 1 \& 5) contributing towards sub-type `Flower class 1' conditioned on the respective other top-3 regions (rows) in joint decision-making. The self-attention (self-loop) is also shown in the top-2 regions. b) Similarly, top-2 regions (cols 5 \& 12) contributing towards sub-type `Flower class 61' conditioned on the respective other top-3 regions (rows). Region proposals are shown in the respective original images.}
    \label{fig:fig_attn_flower}
\end{figure*}

\end{document}